\documentclass[]{bytedance_seed}

\usepackage[toc,page,header]{appendix}
\usepackage{amsmath,amsfonts,amssymb}
\usepackage{pifont}
\DeclareRobustCommand{\circlednum}[1]{\raisebox{-0.15ex}{\scalebox{1.25}{\ding{#1}}}}
\usepackage{graphicx}
\usepackage{eso-pic}
\usepackage{booktabs}
\usepackage{array}
\usepackage{tabularx}
\usepackage{longtable}

\usepackage{multirow}
\usepackage{makecell}
\usepackage{enumitem}
\usepackage{xcolor}
\usepackage{colortbl}
\usepackage{tikz}
\usetikzlibrary{arrows.meta,positioning,calc}
\usepackage{placeins}
\usepackage[type1]{droidserif}

\definecolor{seedblue}{HTML}{2E5AA8}

\usepackage{fontawesome5}
\definecolor{tableInk}{HTML}{263D55}
\definecolor{tableHeader}{HTML}{EEF3F8}
\definecolor{tableRule}{HTML}{A8B7C6}
\definecolor{TableEntryRule}{HTML}{B6C5D4}

\definecolor{tableMuted}{HTML}{A4AEB8}
\definecolor{resourceInk}{HTML}{8A5A2B}
\definecolor{citationInk}{HTML}{315F9B}
\colorlet{resourcePaper}{resourceInk}
\colorlet{resourceCode}{resourceInk}
\colorlet{resourceData}{resourceInk}
\colorlet{resourceBench}{resourceInk}
\hypersetup{citecolor=citationInk,urlcolor=citationInk}
\newcommand{\ResourceGlyph}[1]{\tikz[x=0.85pt,y=0.85pt,baseline=0.6pt,line width=0.42pt,line cap=butt,line join=miter]{\path[use as bounding box] (0,0) rectangle (7,7);#1}}
\newcommand{\PaperResourceIcon}{\ResourceGlyph{\draw[draw=.] (1,0.5)--(1,6.5)--(4,6.5)--(6,4.5)--(6,0.5)--cycle;\draw[draw=.] (4,6.5)--(4,4.5)--(6,4.5) (2,3.2)--(5,3.2) (2,2)--(4.5,2);}}
\newcommand{\CodeResourceIcon}{\raisebox{-0.6pt}{\resizebox{6pt}{!}{\faGithub}}}
\newcommand{\DataResourceIcon}{\ResourceGlyph{\draw[draw=.] (3.5,5.5) ellipse (2.6 and 0.9);\draw[draw=.] (0.9,5.5)--(0.9,1.5) .. controls (0.9,0.3) and (6.1,0.3) .. (6.1,1.5)--(6.1,5.5);\draw[draw=.] (0.9,3.5) .. controls (0.9,2.3) and (6.1,2.3) .. (6.1,3.5);}}
\newcommand{\BenchResourceIcon}{\ResourceGlyph{\draw[draw=.] (3,6)--(1,6)--(1,1)--(6,1)--(6,3);\draw[draw=.] (3.1,3.1)--(6,6) (4,6)--(6,6)--(6,4);}}
\newcommand{\ResourceFont}{\normalfont\fontfamily{phv}\mdseries\fontsize{7}{8.4}\selectfont}
\newcommand{\DeclareTableResource}[6]{%
  \expandafter\gdef\csname tablelink@#1@#2\endcsname{\href{#3}{\textcolor{#6}{\hbox{#4\kern0.22em#5}}}}%
  \expandafter\gdef\csname tableicon@#1@#2\endcsname{\href{#3}{\textcolor{#6}{#4}}}%
}
\newcommand{\TableLink}[2]{{\ResourceFont\ifcsname tablelink@#1@#2\endcsname\csname tablelink@#1@#2\endcsname\else\textcolor{tableMuted}{\textendash}\fi}}
\newcommand{\ResourceInline}[1]{{\ResourceFont%
  \ifcsname tablelink@paper@#1\endcsname\csname tablelink@paper@#1\endcsname\hspace{0.5em}\fi%
  \ifcsname tablelink@code@#1\endcsname\csname tablelink@code@#1\endcsname\hspace{0.5em}\fi%
  \ifcsname tablelink@data@#1\endcsname\csname tablelink@data@#1\endcsname\hspace{0.5em}\fi%
  \ifcsname tablelink@bench@#1\endcsname\csname tablelink@bench@#1\endcsname\fi}}
\newcommand{\ResourceIcons}[1]{{\ResourceFont%
  \ifcsname tableicon@paper@#1\endcsname\csname tableicon@paper@#1\endcsname\,\fi%
  \ifcsname tableicon@code@#1\endcsname\csname tableicon@code@#1\endcsname\,\fi%
  \ifcsname tableicon@data@#1\endcsname\csname tableicon@data@#1\endcsname\,\fi%
  \ifcsname tableicon@bench@#1\endcsname\csname tableicon@bench@#1\endcsname\fi}}

\newcommand{\MethodName}[1]{\textcolor{tableInk}{#1}}
\newcommand{\MethodEntryRes}[3]{\MethodName{#1}~\cite{#2}\par\nobreak\ResourceInline{#3}\par\vspace{2pt}}
\newcommand{\MethodEntry}[2]{\MethodEntryRes{#1}{#2}{#2}}

\DeclareTableResource{paper}{admire}{https://arxiv.org/abs/2602.11524}{\PaperResourceIcon}{Paper}{resourcePaper}
\DeclareTableResource{paper}{aeslides}{https://arxiv.org/abs/2604.22840}{\PaperResourceIcon}{Paper}{resourcePaper}
\DeclareTableResource{code}{aeslides}{https://github.com/ympan0508/aeslides}{\CodeResourceIcon}{Code}{resourceCode}
\DeclareTableResource{data}{aeslides}{https://huggingface.co/datasets/ympan/aeslides-reward-bench}{\DataResourceIcon}{Data}{resourceData}
\DeclareTableResource{paper}{agentcompany}{https://arxiv.org/abs/2412.14161}{\PaperResourceIcon}{Paper}{resourcePaper}
\DeclareTableResource{code}{agentcompany}{https://github.com/TheAgentCompany/TheAgentCompany}{\CodeResourceIcon}{Code}{resourceCode}
\DeclareTableResource{bench}{agentcompany}{https://the-agent-company.com/}{\BenchResourceIcon}{Website}{resourceBench}
\DeclareTableResource{paper}{agentifoneday}{https://arxiv.org/abs/2601.20613}{\PaperResourceIcon}{Paper}{resourcePaper}
\DeclareTableResource{data}{agentifoneday}{https://huggingface.co/datasets/xbench/AgentIF-OneDay}{\DataResourceIcon}{Data}{resourceData}
\DeclareTableResource{bench}{agentifoneday}{https://xbench.org/}{\BenchResourceIcon}{Website}{resourceBench}
\DeclareTableResource{paper}{agentjudge}{https://arxiv.org/abs/2410.10934}{\PaperResourceIcon}{Paper}{resourcePaper}
\DeclareTableResource{code}{agentjudge}{https://github.com/metauto-ai/agent-as-a-judge}{\CodeResourceIcon}{Code}{resourceCode}
\DeclareTableResource{data}{agentjudge}{https://huggingface.co/datasets/DEVAI-benchmark/DEVAI}{\DataResourceIcon}{Data}{resourceData}
\DeclareTableResource{paper}{agentmercury}{https://arxiv.org/abs/2608.20634}{\PaperResourceIcon}{Paper}{resourcePaper}
\DeclareTableResource{data}{agentmercury}{https://huggingface.co/datasets/Minbyul/AgentMercury-corpus-sample}{\DataResourceIcon}{Data}{resourceData}
\DeclareTableResource{bench}{agentmercury}{https://minstar.github.io/AgentMercury/}{\BenchResourceIcon}{Website}{resourceBench}
\DeclareTableResource{paper}{agentrealwork}{https://arxiv.org/abs/2603.01203}{\PaperResourceIcon}{Paper}{resourcePaper}
\DeclareTableResource{paper}{agentworldmodel}{https://arxiv.org/abs/2602.10090}{\PaperResourceIcon}{Paper}{resourcePaper}
\DeclareTableResource{code}{agentworldmodel}{https://github.com/Snowflake-Labs/agent-world-model}{\CodeResourceIcon}{Code}{resourceCode}
\DeclareTableResource{data}{agentworldmodel}{https://huggingface.co/datasets/Snowflake/AgentWorldModel-1K}{\DataResourceIcon}{Data}{resourceData}
\DeclareTableResource{paper}{anchor}{https://arxiv.org/abs/2605.26321}{\PaperResourceIcon}{Paper}{resourcePaper}
\DeclareTableResource{code}{anchor}{https://github.com/agentic-labs/erp-bench}{\CodeResourceIcon}{Code}{resourceCode}
\DeclareTableResource{bench}{anchor}{https://erpbench.ai/}{\BenchResourceIcon}{Website}{resourceBench}
\DeclareTableResource{paper}{androidworld}{https://arxiv.org/abs/2405.14573}{\PaperResourceIcon}{Paper}{resourcePaper}
\DeclareTableResource{code}{androidworld}{https://github.com/google-research/android_world}{\CodeResourceIcon}{Code}{resourceCode}
\DeclareTableResource{data}{androidworld}{https://github.com/google-research/android_world/tree/main/android_world/task_evals}{\DataResourceIcon}{Data}{resourceData}
\DeclareTableResource{bench}{androidworld}{https://google-research.github.io/android_world/}{\BenchResourceIcon}{Website}{resourceBench}
\DeclareTableResource{paper}{apex}{https://arxiv.org/abs/2601.14242}{\PaperResourceIcon}{Paper}{resourcePaper}
\DeclareTableResource{code}{apex}{https://github.com/Mercor-Intelligence/archipelago}{\CodeResourceIcon}{Code}{resourceCode}
\DeclareTableResource{data}{apex}{https://huggingface.co/datasets/mercor/apex-agents}{\DataResourceIcon}{Data}{resourceData}
\DeclareTableResource{paper}{apibank}{https://arxiv.org/abs/2304.08244}{\PaperResourceIcon}{Paper}{resourcePaper}
\DeclareTableResource{code}{apibank}{https://github.com/AlibabaResearch/DAMO-ConvAI/tree/main/api-bank}{\CodeResourceIcon}{Code}{resourceCode}
\DeclareTableResource{data}{apibank}{https://huggingface.co/datasets/liminghao1630/API-Bank}{\DataResourceIcon}{Data}{resourceData}
\DeclareTableResource{paper}{appworld}{https://arxiv.org/abs/2407.18901}{\PaperResourceIcon}{Paper}{resourcePaper}
\DeclareTableResource{code}{appworld}{https://github.com/stonybrooknlp/appworld}{\CodeResourceIcon}{Code}{resourceCode}
\DeclareTableResource{bench}{appworld}{https://appworld.dev/}{\BenchResourceIcon}{Website}{resourceBench}
\DeclareTableResource{paper}{autodesign}{https://arxiv.org/abs/2608.13560}{\PaperResourceIcon}{Paper}{resourcePaper}
\DeclareTableResource{code}{autodesign}{https://github.com/Yaxin9Luo/AutoDesign}{\CodeResourceIcon}{Code}{resourceCode}
\DeclareTableResource{data}{autodesign}{https://huggingface.co/datasets/YaxinLuo/PosterBench}{\DataResourceIcon}{Data}{resourceData}
\DeclareTableResource{bench}{autodesign}{https://autodesign.designanything.ai/}{\BenchResourceIcon}{Website}{resourceBench}
\DeclareTableResource{paper}{automationbench}{https://arxiv.org/abs/2604.18934}{\PaperResourceIcon}{Paper}{resourcePaper}
\DeclareTableResource{code}{automationbench}{https://github.com/zapier/AutomationBench}{\CodeResourceIcon}{Code}{resourceCode}
\DeclareTableResource{bench}{automationbench}{https://zapier.com/benchmarks}{\BenchResourceIcon}{Website}{resourceBench}
\DeclareTableResource{paper}{bankertoolbench}{https://arxiv.org/abs/2604.11304}{\PaperResourceIcon}{Paper}{resourcePaper}
\DeclareTableResource{code}{bankertoolbench}{https://github.com/Handshake-AI-Research/bankertoolbench}{\CodeResourceIcon}{Code}{resourceCode}
\DeclareTableResource{data}{bankertoolbench}{https://huggingface.co/datasets/handshake-ai-research/bankertoolbench}{\DataResourceIcon}{Data}{resourceData}
\DeclareTableResource{paper}{benchcad}{https://arxiv.org/abs/2605.10865}{\PaperResourceIcon}{Paper}{resourcePaper}
\DeclareTableResource{code}{benchcad}{https://github.com/BenchCAD/BenchCAD-main}{\CodeResourceIcon}{Code}{resourceCode}
\DeclareTableResource{data}{benchcad}{https://huggingface.co/datasets/BenchCAD/BenchCAD}{\DataResourceIcon}{Data}{resourceData}
\DeclareTableResource{bench}{benchcad}{https://benchcad.github.io/BenchCAD_webpage/}{\BenchResourceIcon}{Website}{resourceBench}
\DeclareTableResource{paper}{betterharnesses}{https://arxiv.org/abs/2607.08938}{\PaperResourceIcon}{Paper}{resourcePaper}
\DeclareTableResource{paper}{browsergym}{https://arxiv.org/abs/2412.05467}{\PaperResourceIcon}{Paper}{resourcePaper}
\DeclareTableResource{code}{browsergym}{https://github.com/ServiceNow/BrowserGym}{\CodeResourceIcon}{Code}{resourceCode}
\DeclareTableResource{bench}{browsergym}{https://huggingface.co/spaces/ServiceNow/browsergym-leaderboard}{\BenchResourceIcon}{Website}{resourceBench}
\DeclareTableResource{paper}{centaurbench}{https://arxiv.org/abs/2608.18554}{\PaperResourceIcon}{Paper}{resourcePaper}
\DeclareTableResource{code}{centaurbench}{https://github.com/kennywong524/best-player-not-best-coach}{\CodeResourceIcon}{Code}{resourceCode}
\DeclareTableResource{data}{centaurbench}{https://github.com/kennywong524/best-player-not-best-coach/tree/main/tasks}{\DataResourceIcon}{Data}{resourceData}
\DeclareTableResource{bench}{centaurbench}{https://kennywong524.github.io/centaur-benchmark/}{\BenchResourceIcon}{Website}{resourceBench}
\DeclareTableResource{paper}{ciwork}{https://arxiv.org/abs/2604.21308}{\PaperResourceIcon}{Paper}{resourcePaper}
\DeclareTableResource{code}{ciwork}{https://github.com/microsoft/ACV}{\CodeResourceIcon}{Code}{resourceCode}
\DeclareTableResource{paper}{claudedocs}{https://code.claude.com/docs/en/overview}{\PaperResourceIcon}{Docs}{resourcePaper}
\DeclareTableResource{code}{claudedocs}{https://github.com/anthropics/claude-code}{\CodeResourceIcon}{Repo}{resourceCode}
\DeclareTableResource{paper}{clbench}{https://arxiv.org/abs/2602.03587}{\PaperResourceIcon}{Paper}{resourcePaper}
\DeclareTableResource{code}{clbench}{https://github.com/Tencent-Hunyuan/CL-bench}{\CodeResourceIcon}{Code}{resourceCode}
\DeclareTableResource{data}{clbench}{https://huggingface.co/datasets/tencent/CL-bench}{\DataResourceIcon}{Data}{resourceData}
\DeclareTableResource{bench}{clbench}{https://www.clbench.com/}{\BenchResourceIcon}{Website}{resourceBench}
\DeclareTableResource{paper}{codeskill}{https://arxiv.org/abs/2605.25430}{\PaperResourceIcon}{Paper}{resourcePaper}
\DeclareTableResource{paper}{codexdocs}{https://developers.openai.com/codex/codex-manual.md}{\PaperResourceIcon}{Docs}{resourcePaper}
\DeclareTableResource{code}{codexdocs}{https://github.com/openai/codex}{\CodeResourceIcon}{Repo}{resourceCode}
\DeclareTableResource{paper}{cogevol}{https://arxiv.org/abs/2608.30968}{\PaperResourceIcon}{Paper}{resourcePaper}
\DeclareTableResource{code}{cogevol}{https://github.com/CogEvol/CogEvol-4B}{\CodeResourceIcon}{Code}{resourceCode}
\DeclareTableResource{paper}{convfinqa}{https://arxiv.org/abs/2210.03849}{\PaperResourceIcon}{Paper}{resourcePaper}
\DeclareTableResource{code}{convfinqa}{https://github.com/czyssrs/ConvFinQA}{\CodeResourceIcon}{Code}{resourceCode}
\DeclareTableResource{data}{convfinqa}{https://github.com/czyssrs/ConvFinQA/blob/main/data.zip}{\DataResourceIcon}{Data}{resourceData}
\DeclareTableResource{paper}{crmapro}{https://arxiv.org/abs/2505.18878}{\PaperResourceIcon}{Paper}{resourcePaper}
\DeclareTableResource{code}{crmapro}{https://github.com/SalesforceAIResearch/CRMArena}{\CodeResourceIcon}{Code}{resourceCode}
\DeclareTableResource{data}{crmapro}{https://huggingface.co/datasets/Salesforce/CRMArenaPro}{\DataResourceIcon}{Data}{resourceData}
\DeclareTableResource{paper}{crmarena}{https://arxiv.org/abs/2411.02305}{\PaperResourceIcon}{Paper}{resourcePaper}
\DeclareTableResource{code}{crmarena}{https://github.com/SalesforceAIResearch/CRMArena}{\CodeResourceIcon}{Code}{resourceCode}
\DeclareTableResource{data}{crmarena}{https://huggingface.co/datasets/Salesforce/CRMArena}{\DataResourceIcon}{Data}{resourceData}
\DeclareTableResource{paper}{curateevo}{https://arxiv.org/abs/2607.06140}{\PaperResourceIcon}{Paper}{resourcePaper}
\DeclareTableResource{paper}{davincidev}{https://arxiv.org/abs/2601.18418}{\PaperResourceIcon}{Paper}{resourcePaper}
\DeclareTableResource{code}{davincidev}{https://github.com/GAIR-NLP/daVinci-Dev}{\CodeResourceIcon}{Code}{resourceCode}
\DeclareTableResource{data}{davincidev}{https://huggingface.co/datasets/GAIR/daVinci-Dev}{\DataResourceIcon}{Data}{resourceData}
\DeclareTableResource{paper}{discoverybench}{https://arxiv.org/abs/2407.01725}{\PaperResourceIcon}{Paper}{resourcePaper}
\DeclareTableResource{code}{discoverybench}{https://github.com/allenai/discoverybench}{\CodeResourceIcon}{Code}{resourceCode}
\DeclareTableResource{data}{discoverybench}{https://huggingface.co/datasets/allenai/discoverybench}{\DataResourceIcon}{Data}{resourceData}
\DeclareTableResource{paper}{dmrl}{https://arxiv.org/abs/2609.02170}{\PaperResourceIcon}{Paper}{resourcePaper}
\DeclareTableResource{paper}{docvqa}{https://arxiv.org/abs/2007.00398}{\PaperResourceIcon}{Paper}{resourcePaper}
\DeclareTableResource{data}{docvqa}{https://www.docvqa.org/datasets/docvqa}{\DataResourceIcon}{Data}{resourceData}
\DeclareTableResource{bench}{docvqa}{https://www.docvqa.org/}{\BenchResourceIcon}{Website}{resourceBench}
\DeclareTableResource{paper}{doris2026cadbench}{https://arxiv.org/abs/2605.10873}{\PaperResourceIcon}{Paper}{resourcePaper}
\DeclareTableResource{code}{doris2026cadbench}{https://github.com/anniedoris/CADBench}{\CodeResourceIcon}{Code}{resourceCode}
\DeclareTableResource{data}{doris2026cadbench}{https://huggingface.co/datasets/DeCoDELab/CADBench}{\DataResourceIcon}{Data}{resourceData}
\DeclareTableResource{bench}{doris2026cadbench}{https://anniedoris.github.io/CADBench/}{\BenchResourceIcon}{Website}{resourceBench}
\DeclareTableResource{paper}{dsbench}{https://arxiv.org/abs/2409.07703}{\PaperResourceIcon}{Paper}{resourcePaper}
\DeclareTableResource{code}{dsbench}{https://github.com/LiqiangJing/DSBench}{\CodeResourceIcon}{Code}{resourceCode}
\DeclareTableResource{data}{dsbench}{https://huggingface.co/datasets/liqiang888/DSBench}{\DataResourceIcon}{Data}{resourceData}
\DeclareTableResource{bench}{dsbench}{https://liqiangjing.github.io/dsbench.github.io/}{\BenchResourceIcon}{Website}{resourceBench}
\DeclareTableResource{paper}{du2024blenderllmtraininglargelanguage}{https://arxiv.org/abs/2412.14203}{\PaperResourceIcon}{Paper}{resourcePaper}
\DeclareTableResource{code}{du2024blenderllmtraininglargelanguage}{https://github.com/FreedomIntelligence/BlenderLLM}{\CodeResourceIcon}{Code}{resourceCode}
\DeclareTableResource{data}{du2024blenderllmtraininglargelanguage}{https://huggingface.co/datasets/FreedomIntelligence/CADBench}{\DataResourceIcon}{Data}{resourceData}
\DeclareTableResource{paper}{ecommercebench}{https://arxiv.org/abs/2608.30730}{\PaperResourceIcon}{Paper}{resourcePaper}
\DeclareTableResource{code}{ecommercebench}{https://github.com/QwenLM/E-CommerceBench}{\CodeResourceIcon}{Code}{resourceCode}
\DeclareTableResource{data}{ecommercebench}{https://github.com/QwenLM/E-CommerceBench/tree/main/data}{\DataResourceIcon}{Data}{resourceData}
\DeclareTableResource{bench}{ecommercebench}{https://ecbench.github.io/}{\BenchResourceIcon}{Website}{resourceBench}
\DeclareTableResource{paper}{enterprisebench}{https://arxiv.org/abs/2510.27287}{\PaperResourceIcon}{Paper}{resourcePaper}
\DeclareTableResource{code}{enterprisebench}{https://github.com/ast-fri/EnterpriseBench}{\CodeResourceIcon}{Code}{resourceCode}
\DeclareTableResource{data}{enterprisebench}{https://huggingface.co/datasets/AST-FRI/EnterpriseBench}{\DataResourceIcon}{Data}{resourceData}
\DeclareTableResource{bench}{enterprisebench}{https://ast-fri.github.io/EnterpriseBench/}{\BenchResourceIcon}{Website}{resourceBench}
\DeclareTableResource{paper}{enterpriselab}{https://arxiv.org/abs/2603.21630}{\PaperResourceIcon}{Paper}{resourcePaper}
\DeclareTableResource{code}{enterpriselab}{https://github.com/ast-fri/EnterpriseLab}{\CodeResourceIcon}{Code}{resourceCode}
\DeclareTableResource{bench}{enterpriselab}{https://ast-fri.github.io/EnterpriseLab/}{\BenchResourceIcon}{Website}{resourceBench}
\DeclareTableResource{paper}{enterpriseops}{https://arxiv.org/abs/2603.13594}{\PaperResourceIcon}{Paper}{resourcePaper}
\DeclareTableResource{code}{enterpriseops}{https://github.com/ServiceNow/EnterpriseOps-Gym}{\CodeResourceIcon}{Code}{resourceCode}
\DeclareTableResource{data}{enterpriseops}{https://huggingface.co/datasets/ServiceNow-AI/EnterpriseOps-Gym}{\DataResourceIcon}{Data}{resourceData}
\DeclareTableResource{bench}{enterpriseops}{https://enterpriseops-gym.github.io/}{\BenchResourceIcon}{Website}{resourceBench}
\DeclareTableResource{paper}{envcraft}{https://arxiv.org/abs/2609.05576}{\PaperResourceIcon}{Paper}{resourcePaper}
\DeclareTableResource{code}{envcraft}{https://github.com/zeng-yirong/EnvCraft}{\CodeResourceIcon}{Code}{resourceCode}
\DeclareTableResource{paper}{fastrl}{https://arxiv.org/abs/2605.12484}{\PaperResourceIcon}{Paper}{resourcePaper}
\DeclareTableResource{bench}{fastrl}{https://gepa-ai.github.io/gepa/blog/2026/05/11/learning-fast-and-slow/}{\BenchResourceIcon}{Website}{resourceBench}
\DeclareTableResource{paper}{fieldworkarena}{https://arxiv.org/abs/2505.19662}{\PaperResourceIcon}{Paper}{resourcePaper}
\DeclareTableResource{code}{fieldworkarena}{https://github.com/FujitsuResearch/FieldWorkArena}{\CodeResourceIcon}{Code}{resourceCode}
\DeclareTableResource{data}{fieldworkarena}{https://github.com/FujitsuResearch/FieldWorkArena/tree/main/data/image}{\DataResourceIcon}{Data}{resourceData}
\DeclareTableResource{paper}{financeagentbenchmark}{https://arxiv.org/abs/2508.00828}{\PaperResourceIcon}{Paper}{resourcePaper}
\DeclareTableResource{code}{financeagentbenchmark}{https://github.com/vals-ai/finance-agent}{\CodeResourceIcon}{Code}{resourceCode}
\DeclareTableResource{bench}{financeagentbenchmark}{https://www.vals.ai/benchmarks/finance_agent}{\BenchResourceIcon}{Website}{resourceBench}
\DeclareTableResource{code}{financeagentv2}{https://github.com/vals-ai/finance-agent-v2}{\CodeResourceIcon}{Code}{resourceCode}
\DeclareTableResource{bench}{financeagentv2}{https://www.vals.ai/benchmarks/fabv2}{\BenchResourceIcon}{Website}{resourceBench}
\DeclareTableResource{paper}{financebench}{https://arxiv.org/abs/2311.11944}{\PaperResourceIcon}{Paper}{resourcePaper}
\DeclareTableResource{code}{financebench}{https://github.com/patronus-ai/financebench}{\CodeResourceIcon}{Code}{resourceCode}
\DeclareTableResource{data}{financebench}{https://huggingface.co/datasets/PatronusAI/financebench}{\DataResourceIcon}{Data}{resourceData}
\DeclareTableResource{paper}{finqa}{https://arxiv.org/abs/2109.00122}{\PaperResourceIcon}{Paper}{resourcePaper}
\DeclareTableResource{code}{finqa}{https://github.com/czyssrs/FinQA}{\CodeResourceIcon}{Code}{resourceCode}
\DeclareTableResource{data}{finqa}{https://github.com/czyssrs/FinQA/tree/main/dataset}{\DataResourceIcon}{Data}{resourceData}
\DeclareTableResource{paper}{gaia}{https://arxiv.org/abs/2311.12983}{\PaperResourceIcon}{Paper}{resourcePaper}
\DeclareTableResource{data}{gaia}{https://huggingface.co/datasets/gaia-benchmark/GAIA}{\DataResourceIcon}{Data}{resourceData}
\DeclareTableResource{paper}{gdppdf}{https://arxiv.org/abs/2607.11192}{\PaperResourceIcon}{Paper}{resourcePaper}
\DeclareTableResource{code}{gdppdf}{https://github.com/surge-ai/gdp-pdf}{\CodeResourceIcon}{Code}{resourceCode}
\DeclareTableResource{data}{gdppdf}{https://huggingface.co/datasets/surgeai/GDP.pdf}{\DataResourceIcon}{Data}{resourceData}
\DeclareTableResource{bench}{gdppdf}{https://artificialanalysis.ai/evaluations/gdp-pdf}{\BenchResourceIcon}{Website}{resourceBench}
\DeclareTableResource{paper}{gdpval}{https://cdn.openai.com/pdf/d5eb7428-c4e9-4a33-bd86-86dd4bcf12ce/GDPval.pdf}{\PaperResourceIcon}{Paper}{resourcePaper}
\DeclareTableResource{data}{gdpval}{https://huggingface.co/datasets/openai/gdpval}{\DataResourceIcon}{Data}{resourceData}
\DeclareTableResource{bench}{gdpval}{https://openai.com/index/gdpval/}{\BenchResourceIcon}{Website}{resourceBench}
\DeclareTableResource{paper}{gepa}{https://arxiv.org/abs/2507.19457}{\PaperResourceIcon}{Paper}{resourcePaper}
\DeclareTableResource{code}{gepa}{https://github.com/gepa-ai/gepa}{\CodeResourceIcon}{Code}{resourceCode}
\DeclareTableResource{bench}{gepa}{https://gepa-ai.github.io/gepa/}{\BenchResourceIcon}{Website}{resourceBench}
\DeclareTableResource{paper}{ghd}{https://arxiv.org/abs/2608.06065}{\PaperResourceIcon}{Paper}{resourcePaper}
\DeclareTableResource{paper}{graphforge}{https://arxiv.org/abs/2609.38923}{\PaperResourceIcon}{Paper}{resourcePaper}
\DeclareTableResource{data}{graphforge}{https://huggingface.co/datasets/groundhogLLM/GraphForge-SFT-2169-clean}{\DataResourceIcon}{Data}{resourceData}
\DeclareTableResource{code}{harbor}{https://github.com/harbor-framework/harbor}{\CodeResourceIcon}{Code}{resourceCode}
\DeclareTableResource{bench}{harbor}{https://harborframework.com/}{\BenchResourceIcon}{Website}{resourceBench}
\DeclareTableResource{paper}{harnessbench}{https://arxiv.org/abs/2605.27922}{\PaperResourceIcon}{Paper}{resourcePaper}
\DeclareTableResource{code}{harveylab}{https://github.com/harveyai/harvey-labs/tree/v1.0}{\CodeResourceIcon}{Code}{resourceCode}
\DeclareTableResource{data}{harveylab}{https://github.com/harveyai/harvey-labs/tree/v1.0/tasks}{\DataResourceIcon}{Data}{resourceData}
\DeclareTableResource{paper}{healthagentbench}{https://arxiv.org/abs/2606.31179}{\PaperResourceIcon}{Paper}{resourcePaper}
\DeclareTableResource{code}{healthagentbench}{https://github.com/microsoft/HealthAgentBench}{\CodeResourceIcon}{Code}{resourceCode}
\DeclareTableResource{data}{healthagentbench}{https://github.com/microsoft/HealthAgentBench/tree/main/tasks}{\DataResourceIcon}{Data}{resourceData}
\DeclareTableResource{bench}{healthagentbench}{https://microsoft.github.io/HealthAgentBench/}{\BenchResourceIcon}{Website}{resourceBench}
\DeclareTableResource{paper}{hiper}{https://arxiv.org/abs/2602.16165}{\PaperResourceIcon}{Paper}{resourcePaper}
\DeclareTableResource{code}{hiper}{https://github.com/JonP07/HiPER-agent}{\CodeResourceIcon}{Code}{resourceCode}
\DeclareTableResource{bench}{hiper}{https://jonp07.github.io/HiPER-agent/}{\BenchResourceIcon}{Website}{resourceBench}
\DeclareTableResource{code}{inspectai}{https://github.com/UKGovernmentBEIS/inspect_ai}{\CodeResourceIcon}{Code}{resourceCode}
\DeclareTableResource{bench}{inspectai}{https://inspect.aisi.org.uk/}{\BenchResourceIcon}{Website}{resourceBench}
\DeclareTableResource{paper}{jitagent}{https://arxiv.org/abs/2608.25593}{\PaperResourceIcon}{Paper}{resourcePaper}
\DeclareTableResource{code}{jitagent}{https://github.com/bingreeky/JIT}{\CodeResourceIcon}{Code}{resourceCode}
\DeclareTableResource{data}{jitagent}{https://github.com/bingreeky/JIT/tree/main/dataset}{\DataResourceIcon}{Data}{resourceData}
\DeclareTableResource{bench}{jitagent}{https://guibinz.top/JIT-site/}{\BenchResourceIcon}{Website}{resourceBench}
\DeclareTableResource{data}{jobbench}{https://huggingface.co/datasets/JobBench/job-bench}{\DataResourceIcon}{Data}{resourceData}
\DeclareTableResource{paper}{jtpro}{https://arxiv.org/abs/2604.19821}{\PaperResourceIcon}{Paper}{resourcePaper}
\DeclareTableResource{paper}{lawbench}{https://arxiv.org/abs/2309.16289}{\PaperResourceIcon}{Paper}{resourcePaper}
\DeclareTableResource{code}{lawbench}{https://github.com/open-compass/LawBench}{\CodeResourceIcon}{Code}{resourceCode}
\DeclareTableResource{data}{lawbench}{https://github.com/open-compass/LawBench/tree/main/data}{\DataResourceIcon}{Data}{resourceData}
\DeclareTableResource{paper}{legalbench}{https://arxiv.org/abs/2308.11462}{\PaperResourceIcon}{Paper}{resourcePaper}
\DeclareTableResource{code}{legalbench}{https://github.com/HazyResearch/legalbench}{\CodeResourceIcon}{Code}{resourceCode}
\DeclareTableResource{data}{legalbench}{https://huggingface.co/datasets/nguha/legalbench}{\DataResourceIcon}{Data}{resourceData}
\DeclareTableResource{paper}{litegui}{https://arxiv.org/abs/2605.07505}{\PaperResourceIcon}{Paper}{resourcePaper}
\DeclareTableResource{paper}{m365docs}{https://learn.microsoft.com/en-us/copilot/microsoft-365/microsoft-365-copilot-architecture}{\PaperResourceIcon}{Docs}{resourcePaper}
\DeclareTableResource{paper}{mcpatlas}{https://arxiv.org/abs/2602.00933}{\PaperResourceIcon}{Paper}{resourcePaper}
\DeclareTableResource{code}{mcpatlas}{https://github.com/scaleapi/mcp-atlas}{\CodeResourceIcon}{Code}{resourceCode}
\DeclareTableResource{data}{mcpatlas}{https://huggingface.co/datasets/ScaleAI/MCP-Atlas}{\DataResourceIcon}{Data}{resourceData}
\DeclareTableResource{paper}{medagentbench}{https://arxiv.org/abs/2501.14654}{\PaperResourceIcon}{Paper}{resourcePaper}
\DeclareTableResource{code}{medagentbench}{https://github.com/stanfordmlgroup/MedAgentBench}{\CodeResourceIcon}{Code}{resourceCode}
\DeclareTableResource{bench}{medcode}{https://www.vals.ai/benchmarks/medcode}{\BenchResourceIcon}{Website}{resourceBench}
\DeclareTableResource{bench}{medscribe}{https://www.vals.ai/benchmarks/medscribe}{\BenchResourceIcon}{Website}{resourceBench}
\DeclareTableResource{paper}{metaharness}{https://arxiv.org/abs/2603.28052}{\PaperResourceIcon}{Paper}{resourcePaper}
\DeclareTableResource{code}{metaharness}{https://github.com/stanford-iris-lab/meta-harness-tbench2-artifact}{\CodeResourceIcon}{Code}{resourceCode}
\DeclareTableResource{bench}{metaharness}{https://yoonholee.com/meta-harness/}{\BenchResourceIcon}{Website}{resourceBench}
\DeclareTableResource{code}{mlcraa}{https://github.com/Wisedocs-AI/medical-long-context-reasoning}{\CodeResourceIcon}{Code}{resourceCode}
\DeclareTableResource{data}{mlcraa}{https://huggingface.co/datasets/Wisedocs/mlcr-dataset}{\DataResourceIcon}{Data}{resourceData}
\DeclareTableResource{bench}{mlcraa}{https://artificialanalysis.ai/evaluations/mlcr-aa}{\BenchResourceIcon}{Website}{resourceBench}
\DeclareTableResource{paper}{mmlongbenchdoc}{https://arxiv.org/abs/2407.01523}{\PaperResourceIcon}{Paper}{resourcePaper}
\DeclareTableResource{code}{mmlongbenchdoc}{https://github.com/mayubo2333/MMLongBench-Doc}{\CodeResourceIcon}{Code}{resourceCode}
\DeclareTableResource{data}{mmlongbenchdoc}{https://github.com/mayubo2333/MMLongBench-Doc/tree/main/data}{\DataResourceIcon}{Data}{resourceData}
\DeclareTableResource{bench}{mmlongbenchdoc}{https://mayubo2333.github.io/MMLongBench-Doc/}{\BenchResourceIcon}{Website}{resourceBench}
\DeclareTableResource{paper}{mmtb}{https://arxiv.org/abs/2504.02623}{\PaperResourceIcon}{Paper}{resourcePaper}
\DeclareTableResource{code}{mmtb}{https://github.com/yupeijei1997/MMTB}{\CodeResourceIcon}{Code}{resourceCode}
\DeclareTableResource{data}{mmtb}{https://huggingface.co/datasets/jpy/MMTB}{\DataResourceIcon}{Data}{resourceData}
\DeclareTableResource{bench}{mmtb}{https://harrywgcn.github.io/mmtb-leaderboard/}{\BenchResourceIcon}{Website}{resourceBench}
\DeclareTableResource{paper}{mtbench}{https://arxiv.org/abs/2306.05685}{\PaperResourceIcon}{Paper}{resourcePaper}
\DeclareTableResource{code}{mtbench}{https://github.com/lm-sys/FastChat/tree/main/fastchat/llm_judge}{\CodeResourceIcon}{Code}{resourceCode}
\DeclareTableResource{data}{mtbench}{https://huggingface.co/datasets/lmsys/mt_bench_human_judgments}{\DataResourceIcon}{Data}{resourceData}
\DeclareTableResource{paper}{multiagentbench}{https://arxiv.org/abs/2503.01935}{\PaperResourceIcon}{Paper}{resourcePaper}
\DeclareTableResource{code}{multiagentbench}{https://github.com/MultiagentBench/MARBLE}{\CodeResourceIcon}{Code}{resourceCode}
\DeclareTableResource{data}{multiagentbench}{https://github.com/MultiagentBench/MARBLE/tree/main/data}{\DataResourceIcon}{Data}{resourceData}
\DeclareTableResource{paper}{nexforge}{https://arxiv.org/abs/2607.14186}{\PaperResourceIcon}{Paper}{resourcePaper}
\DeclareTableResource{bench}{nexforge}{https://nex.sii.edu.cn/}{\BenchResourceIcon}{Website}{resourceBench}
\DeclareTableResource{paper}{odysseys}{https://arxiv.org/abs/2604.24964}{\PaperResourceIcon}{Paper}{resourcePaper}
\DeclareTableResource{bench}{odysseys}{https://odysseysbench.com/}{\BenchResourceIcon}{Website}{resourceBench}
\DeclareTableResource{paper}{officebench}{https://arxiv.org/abs/2407.19056}{\PaperResourceIcon}{Paper}{resourcePaper}
\DeclareTableResource{code}{officebench}{https://github.com/zlwang-cs/OfficeBench}{\CodeResourceIcon}{Code}{resourceCode}
\DeclareTableResource{data}{officebench}{https://github.com/zlwang-cs/OfficeBench/tree/main/tasks}{\DataResourceIcon}{Data}{resourceData}
\DeclareTableResource{paper}{officeqapro}{https://arxiv.org/abs/2603.08655}{\PaperResourceIcon}{Paper}{resourcePaper}
\DeclareTableResource{code}{officeqapro}{https://github.com/databricks/officeqa}{\CodeResourceIcon}{Code}{resourceCode}
\DeclareTableResource{data}{officeqapro}{https://huggingface.co/datasets/databricks/officeqa}{\DataResourceIcon}{Data}{resourceData}
\DeclareTableResource{bench}{officeqapro}{https://www.databricks.com/blog/pdfs-production-announcing-state-art-document-intelligence-databricks}{\BenchResourceIcon}{Website}{resourceBench}
\DeclareTableResource{paper}{odysseybench}{https://arxiv.org/abs/2508.09124}{\PaperResourceIcon}{Paper}{resourcePaper}
\DeclareTableResource{data}{omegauseofficeval}{https://huggingface.co/datasets/baidu-frontier-research/OmegaUse-OfficeVal}{\DataResourceIcon}{Data}{resourceData}
\DeclareTableResource{paper}{omniabench}{https://arxiv.org/abs/2607.14989}{\PaperResourceIcon}{Paper}{resourcePaper}
\DeclareTableResource{code}{omniabench}{https://github.com/scuuy/OmniaBench}{\CodeResourceIcon}{Code}{resourceCode}
\DeclareTableResource{data}{omniabench}{https://huggingface.co/datasets/scuuy666/OmniaBench}{\DataResourceIcon}{Data}{resourceData}
\DeclareTableResource{bench}{omniabench}{https://scuuy.github.io/OmniaBench/}{\BenchResourceIcon}{Website}{resourceBench}
\DeclareTableResource{paper}{onemillionbench}{https://arxiv.org/abs/2603.07980}{\PaperResourceIcon}{Paper}{resourcePaper}
\DeclareTableResource{code}{onemillionbench}{https://github.com/humanlaya/OneMillion-Bench}{\CodeResourceIcon}{Code}{resourceCode}
\DeclareTableResource{data}{onemillionbench}{https://huggingface.co/datasets/humanlaya-data-lab/OneMillion-Bench}{\DataResourceIcon}{Data}{resourceData}
\DeclareTableResource{paper}{openhands}{https://arxiv.org/abs/2407.16741}{\PaperResourceIcon}{Paper}{resourcePaper}
\DeclareTableResource{code}{openhands}{https://github.com/OpenHands/OpenHands}{\CodeResourceIcon}{Code}{resourceCode}
\DeclareTableResource{paper}{openthoughtsagent}{https://arxiv.org/abs/2606.24855}{\PaperResourceIcon}{Paper}{resourcePaper}
\DeclareTableResource{code}{openthoughtsagent}{https://github.com/open-thoughts/OpenThoughts-Agent}{\CodeResourceIcon}{Code}{resourceCode}
\DeclareTableResource{data}{openthoughtsagent}{https://github.com/open-thoughts/OpenThoughts-Agent/tree/main/data}{\DataResourceIcon}{Data}{resourceData}
\DeclareTableResource{paper}{osworld}{https://arxiv.org/abs/2404.07972}{\PaperResourceIcon}{Paper}{resourcePaper}
\DeclareTableResource{code}{osworld}{https://github.com/xlang-ai/OSWorld}{\CodeResourceIcon}{Code}{resourceCode}
\DeclareTableResource{data}{osworld}{https://github.com/xlang-ai/OSWorld/tree/main/evaluation_examples}{\DataResourceIcon}{Data}{resourceData}
\DeclareTableResource{bench}{osworld}{https://os-world.github.io/}{\BenchResourceIcon}{Website}{resourceBench}
\DeclareTableResource{paper}{paperbench}{https://arxiv.org/abs/2504.01848}{\PaperResourceIcon}{Paper}{resourcePaper}
\DeclareTableResource{code}{paperbench}{https://github.com/openai/frontier-evals/tree/main/project/paperbench}{\CodeResourceIcon}{Code}{resourceCode}
\DeclareTableResource{data}{paperbench}{https://github.com/openai/frontier-evals/tree/main/project/paperbench/data}{\DataResourceIcon}{Data}{resourceData}
\DeclareTableResource{code}{piagent}{https://github.com/earendil-works/pi}{\CodeResourceIcon}{Code}{resourceCode}
\DeclareTableResource{paper}{pptagent}{https://arxiv.org/abs/2501.03936}{\PaperResourceIcon}{Paper}{resourcePaper}
\DeclareTableResource{code}{pptagent}{https://github.com/icip-cas/PPTAgent}{\CodeResourceIcon}{Code}{resourceCode}
\DeclareTableResource{paper}{pptc}{https://arxiv.org/abs/2311.01767}{\PaperResourceIcon}{Paper}{resourcePaper}
\DeclareTableResource{code}{pptc}{https://github.com/gydpku/PPTC}{\CodeResourceIcon}{Code}{resourceCode}
\DeclareTableResource{data}{pptc}{https://github.com/gydpku/PPTC/tree/main/test}{\DataResourceIcon}{Data}{resourceData}
\DeclareTableResource{paper}{prbench}{https://arxiv.org/abs/2511.11562}{\PaperResourceIcon}{Paper}{resourcePaper}
\DeclareTableResource{code}{prbench}{https://github.com/scaleapi/PRBench}{\CodeResourceIcon}{Code}{resourceCode}
\DeclareTableResource{data}{prbench}{https://huggingface.co/datasets/ScaleAI/PRBench}{\DataResourceIcon}{Data}{resourceData}
\DeclareTableResource{bench}{prbench}{https://labs.scale.com/papers/prbench}{\BenchResourceIcon}{Website}{resourceBench}
\DeclareTableResource{paper}{presentbench}{https://arxiv.org/abs/2603.07244}{\PaperResourceIcon}{Paper}{resourcePaper}
\DeclareTableResource{code}{presentbench}{https://github.com/PresentBench/PresentBench}{\CodeResourceIcon}{Code}{resourceCode}
\DeclareTableResource{data}{presentbench}{https://huggingface.co/datasets/PresentBench/PresentBench}{\DataResourceIcon}{Data}{resourceData}
\DeclareTableResource{bench}{presentbench}{https://presentbench.github.io/}{\BenchResourceIcon}{Website}{resourceBench}
\DeclareTableResource{paper}{prism}{https://arxiv.org/abs/2609.05736}{\PaperResourceIcon}{Paper}{resourcePaper}
\DeclareTableResource{code}{prism}{https://github.com/airbnb/agent-harness-optimizer}{\CodeResourceIcon}{Code}{resourceCode}
\DeclareTableResource{data}{prism}{https://github.com/airbnb/agent-harness-optimizer/tree/main/data/bfcl}{\DataResourceIcon}{Data}{resourceData}
\DeclareTableResource{paper}{psebench}{https://arxiv.org/abs/2606.05463}{\PaperResourceIcon}{Paper}{resourcePaper}
\DeclareTableResource{paper}{ready}{https://arxiv.org/abs/2609.02095}{\PaperResourceIcon}{Paper}{resourcePaper}
\DeclareTableResource{bench}{ready}{https://labs.scale.com/ready-benchmarks}{\BenchResourceIcon}{Website}{resourceBench}
\DeclareTableResource{paper}{reasoningbank}{https://arxiv.org/abs/2509.25140}{\PaperResourceIcon}{Paper}{resourcePaper}
\DeclareTableResource{code}{reasoningbank}{https://github.com/google-research/reasoning-bank}{\CodeResourceIcon}{Code}{resourceCode}
\DeclareTableResource{paper}{repro}{https://arxiv.org/abs/2606.14302}{\PaperResourceIcon}{Paper}{resourcePaper}
\DeclareTableResource{paper}{rli}{https://arxiv.org/abs/2510.26787}{\PaperResourceIcon}{Paper}{resourcePaper}
\DeclareTableResource{code}{rli}{https://github.com/centerforaisafety/rli_evaluation_platform}{\CodeResourceIcon}{Code}{resourceCode}
\DeclareTableResource{data}{rli}{https://huggingface.co/datasets/cais/rli-public-set}{\DataResourceIcon}{Data}{resourceData}
\DeclareTableResource{bench}{rli}{https://www.remotelabor.ai/}{\BenchResourceIcon}{Website}{resourceBench}
\DeclareTableResource{paper}{rlsd}{https://arxiv.org/abs/2601.20802}{\PaperResourceIcon}{Paper}{resourcePaper}
\DeclareTableResource{code}{rlsd}{https://github.com/lasgroup/SDPO}{\CodeResourceIcon}{Code}{resourceCode}
\DeclareTableResource{data}{rlsd}{https://github.com/lasgroup/SDPO/tree/main/datasets}{\DataResourceIcon}{Data}{resourceData}
\DeclareTableResource{bench}{rlsd}{https://self-distillation.github.io/SDPO}{\BenchResourceIcon}{Website}{resourceBench}
\DeclareTableResource{paper}{scienceagentbench}{https://arxiv.org/abs/2410.05080}{\PaperResourceIcon}{Paper}{resourcePaper}
\DeclareTableResource{code}{scienceagentbench}{https://github.com/osu-nlp-group/ScienceAgentBench}{\CodeResourceIcon}{Code}{resourceCode}
\DeclareTableResource{data}{scienceagentbench}{https://huggingface.co/datasets/osunlp/ScienceAgentBench}{\DataResourceIcon}{Data}{resourceData}
\DeclareTableResource{bench}{scienceagentbench}{https://osu-nlp-group.github.io/ScienceAgentBench/}{\BenchResourceIcon}{Website}{resourceBench}
\DeclareTableResource{paper}{see}{https://arxiv.org/abs/2607.18046}{\PaperResourceIcon}{Paper}{resourcePaper}
\DeclareTableResource{code}{see}{https://github.com/Zhh-Fan/SEE-Structure-aware-Exploring-Exploiting-for-Long-horizon-GUI-Agent-Trajectory-Synthesis}{\CodeResourceIcon}{Code}{resourceCode}
\DeclareTableResource{paper}{sia}{https://arxiv.org/abs/2605.27276}{\PaperResourceIcon}{Paper}{resourcePaper}
\DeclareTableResource{code}{sia}{https://github.com/hexo-ai/sia}{\CodeResourceIcon}{Code}{resourceCode}
\DeclareTableResource{paper}{skill2env}{https://arxiv.org/abs/2609.33772}{\PaperResourceIcon}{Paper}{resourcePaper}
\DeclareTableResource{code}{skill2env}{https://github.com/AllSpark-Research/AgentEnv}{\CodeResourceIcon}{Code}{resourceCode}
\DeclareTableResource{data}{skill2env}{https://github.com/AllSpark-Research/AgentEnv/tree/main/Skill2Env/tasks}{\DataResourceIcon}{Data}{resourceData}
\DeclareTableResource{paper}{skillopt}{https://arxiv.org/abs/2605.23904}{\PaperResourceIcon}{Paper}{resourcePaper}
\DeclareTableResource{code}{skillopt}{https://github.com/microsoft/SkillOpt}{\CodeResourceIcon}{Code}{resourceCode}
\DeclareTableResource{bench}{skillopt}{https://microsoft.github.io/SkillOpt/}{\BenchResourceIcon}{Website}{resourceBench}
\DeclareTableResource{paper}{skillrl}{https://arxiv.org/abs/2602.08234}{\PaperResourceIcon}{Paper}{resourcePaper}
\DeclareTableResource{code}{skillrl}{https://github.com/aiming-lab/SkillRL}{\CodeResourceIcon}{Code}{resourceCode}
\DeclareTableResource{data}{skillrl}{https://huggingface.co/datasets/Jianwen/SkillRL-SFT-Data}{\DataResourceIcon}{Data}{resourceData}
\DeclareTableResource{paper}{slim}{https://arxiv.org/abs/2605.10923}{\PaperResourceIcon}{Paper}{resourcePaper}
\DeclareTableResource{code}{slim}{https://github.com/ejhshen/SLIM}{\CodeResourceIcon}{Code}{resourceCode}
\DeclareTableResource{paper}{skillsmith}{https://arxiv.org/abs/2606.01314}{\PaperResourceIcon}{Paper}{resourcePaper}
\DeclareTableResource{paper}{sliderl}{https://arxiv.org/abs/2603.16839}{\PaperResourceIcon}{Paper}{resourcePaper}
\DeclareTableResource{code}{sliderl}{https://github.com/pushing-the-frontier/slide-forge-llm}{\CodeResourceIcon}{Code}{resourceCode}
\DeclareTableResource{data}{sliderl}{https://huggingface.co/datasets/KarthikRagunathAnandaKumar/sliderl-multi-turn-rollouts}{\DataResourceIcon}{Data}{resourceData}
\DeclareTableResource{paper}{socraticswe}{https://arxiv.org/abs/2606.07412}{\PaperResourceIcon}{Paper}{resourcePaper}
\DeclareTableResource{paper}{spider2}{https://arxiv.org/abs/2411.07763}{\PaperResourceIcon}{Paper}{resourcePaper}
\DeclareTableResource{code}{spider2}{https://github.com/xlang-ai/Spider2}{\CodeResourceIcon}{Code}{resourceCode}
\DeclareTableResource{bench}{spider2}{https://spider2-sql.github.io/}{\BenchResourceIcon}{Website}{resourceBench}
\DeclareTableResource{paper}{spreadsheetbench}{https://arxiv.org/abs/2406.14991}{\PaperResourceIcon}{Paper}{resourcePaper}
\DeclareTableResource{code}{spreadsheetbench}{https://github.com/RUCKBReasoning/SpreadsheetBench}{\CodeResourceIcon}{Code}{resourceCode}
\DeclareTableResource{data}{spreadsheetbench}{https://huggingface.co/datasets/KAKA22/SpreadsheetBench}{\DataResourceIcon}{Data}{resourceData}
\DeclareTableResource{bench}{spreadsheetbench}{https://spreadsheetbench.github.io/}{\BenchResourceIcon}{Website}{resourceBench}
\DeclareTableResource{paper}{spreadsheetrl}{https://arxiv.org/abs/2605.22642}{\PaperResourceIcon}{Paper}{resourcePaper}
\DeclareTableResource{code}{spreadsheetrl}{https://github.com/Spreadsheet-RL/Spreadsheet-RL}{\CodeResourceIcon}{Code}{resourceCode}
\DeclareTableResource{data}{spreadsheetrl}{https://huggingface.co/datasets/Spreadsheet-RL/Spreadsheet-RL}{\DataResourceIcon}{Data}{resourceData}
\DeclareTableResource{bench}{spreadsheetrl}{https://spreadsheet-rl.github.io/}{\BenchResourceIcon}{Website}{resourceBench}
\DeclareTableResource{paper}{startupbench}{https://arxiv.org/abs/2608.17800}{\PaperResourceIcon}{Paper}{resourcePaper}
\DeclareTableResource{bench}{startupbench}{https://startupbench.github.io/}{\BenchResourceIcon}{Website}{resourceBench}
\DeclareTableResource{paper}{sun2026agentslastexam}{https://arxiv.org/abs/2606.05405}{\PaperResourceIcon}{Paper}{resourcePaper}
\DeclareTableResource{data}{sun2026agentslastexam}{https://huggingface.co/datasets/agents-last-exam/agents-last-exam}{\DataResourceIcon}{Data}{resourceData}
\DeclareTableResource{paper}{sweagent}{https://arxiv.org/abs/2405.15793}{\PaperResourceIcon}{Paper}{resourcePaper}
\DeclareTableResource{code}{sweagent}{https://github.com/SWE-agent/SWE-agent}{\CodeResourceIcon}{Code}{resourceCode}
\DeclareTableResource{bench}{sweagent}{https://swe-agent.com/}{\BenchResourceIcon}{Website}{resourceBench}
\DeclareTableResource{paper}{swebench}{https://arxiv.org/abs/2310.06770}{\PaperResourceIcon}{Paper}{resourcePaper}
\DeclareTableResource{code}{swebench}{https://github.com/SWE-bench/SWE-bench}{\CodeResourceIcon}{Code}{resourceCode}
\DeclareTableResource{data}{swebench}{https://huggingface.co/datasets/SWE-bench/SWE-bench}{\DataResourceIcon}{Data}{resourceData}
\DeclareTableResource{bench}{swebench}{https://www.swebench.com/}{\BenchResourceIcon}{Website}{resourceBench}
\DeclareTableResource{paper}{taco}{https://arxiv.org/abs/2604.19572}{\PaperResourceIcon}{Paper}{resourcePaper}
\DeclareTableResource{code}{taco}{https://github.com/multimodal-art-projection/TACO}{\CodeResourceIcon}{Code}{resourceCode}
\DeclareTableResource{paper}{tatqa}{https://arxiv.org/abs/2105.07624}{\PaperResourceIcon}{Paper}{resourcePaper}
\DeclareTableResource{code}{tatqa}{https://github.com/NExTplusplus/TAT-QA}{\CodeResourceIcon}{Code}{resourceCode}
\DeclareTableResource{data}{tatqa}{https://github.com/NExTplusplus/TAT-QA/tree/master/dataset_raw}{\DataResourceIcon}{Data}{resourceData}
\DeclareTableResource{bench}{tatqa}{https://nextplusplus.github.io/TAT-QA/}{\BenchResourceIcon}{Website}{resourceBench}
\DeclareTableResource{paper}{tau2bench}{https://arxiv.org/abs/2506.07982}{\PaperResourceIcon}{Paper}{resourcePaper}
\DeclareTableResource{code}{tau2bench}{https://github.com/sierra-research/tau2-bench}{\CodeResourceIcon}{Code}{resourceCode}
\DeclareTableResource{data}{tau2bench}{https://github.com/sierra-research/tau2-bench/tree/main/data/tau2}{\DataResourceIcon}{Data}{resourceData}
\DeclareTableResource{paper}{taubench}{https://arxiv.org/abs/2406.12045}{\PaperResourceIcon}{Paper}{resourcePaper}
\DeclareTableResource{code}{taubench}{https://github.com/sierra-research/tau-bench}{\CodeResourceIcon}{Code}{resourceCode}
\DeclareTableResource{paper}{tcod}{https://arxiv.org/abs/2604.24005}{\PaperResourceIcon}{Paper}{resourcePaper}
\DeclareTableResource{code}{tcod}{https://github.com/kokolerk/TCOD}{\CodeResourceIcon}{Code}{resourceCode}
\DeclareTableResource{paper}{terminalbench}{https://arxiv.org/abs/2601.11868}{\PaperResourceIcon}{Paper}{resourcePaper}
\DeclareTableResource{code}{terminalbench}{https://github.com/harbor-framework/terminal-bench}{\CodeResourceIcon}{Code}{resourceCode}
\DeclareTableResource{paper}{terminalbench2}{https://arxiv.org/abs/2601.11868}{\PaperResourceIcon}{Paper}{resourcePaper}
\DeclareTableResource{code}{terminalbench2}{https://github.com/harbor-framework/terminal-bench-2}{\CodeResourceIcon}{Code}{resourceCode}
\DeclareTableResource{bench}{terminalbench2}{https://www.tbench.ai/}{\BenchResourceIcon}{Website}{resourceBench}
\DeclareTableResource{bench}{terminalbench}{https://www.tbench.ai/}{\BenchResourceIcon}{Website}{resourceBench}
\DeclareTableResource{paper}{toolbench}{https://arxiv.org/abs/2307.16789}{\PaperResourceIcon}{Paper}{resourcePaper}
\DeclareTableResource{code}{toolbench}{https://github.com/OpenBMB/ToolBench}{\CodeResourceIcon}{Code}{resourceCode}
\DeclareTableResource{paper}{toolsandbox}{https://arxiv.org/abs/2408.04682}{\PaperResourceIcon}{Paper}{resourcePaper}
\DeclareTableResource{code}{toolsandbox}{https://github.com/apple-aiml-research/ToolSandbox}{\CodeResourceIcon}{Code}{resourceCode}
\DeclareTableResource{paper}{trajlong}{https://arxiv.org/abs/2610.04973}{\PaperResourceIcon}{Paper}{resourcePaper}
\DeclareTableResource{paper}{uicopilot}{https://arxiv.org/abs/2604.13822}{\PaperResourceIcon}{Paper}{resourcePaper}
\DeclareTableResource{code}{uicopilot}{https://github.com/ZJU-REAL/UI-Copilot}{\CodeResourceIcon}{Code}{resourceCode}
\DeclareTableResource{paper}{uimem}{https://arxiv.org/abs/2602.05832}{\PaperResourceIcon}{Paper}{resourcePaper}
\DeclareTableResource{bench}{uimem}{https://ui-mem.github.io/}{\BenchResourceIcon}{Website}{resourceBench}
\DeclareTableResource{code}{valslegalresearch}{https://github.com/vals-ai/legal-research-bench}{\CodeResourceIcon}{Code}{resourceCode}
\DeclareTableResource{bench}{valslegalresearch}{https://www.vals.ai/benchmarks/legal_research}{\BenchResourceIcon}{Website}{resourceBench}
\DeclareTableResource{paper}{vera}{https://arxiv.org/abs/2610.05923}{\PaperResourceIcon}{Paper}{resourcePaper}
\DeclareTableResource{paper}{visualwebarena}{https://arxiv.org/abs/2401.13649}{\PaperResourceIcon}{Paper}{resourcePaper}
\DeclareTableResource{code}{visualwebarena}{https://github.com/web-arena-x/visualwebarena}{\CodeResourceIcon}{Code}{resourceCode}
\DeclareTableResource{data}{visualwebarena}{https://github.com/web-arena-x/visualwebarena/tree/main/config_files}{\DataResourceIcon}{Data}{resourceData}
\DeclareTableResource{bench}{visualwebarena}{https://jykoh.com/vwa}{\BenchResourceIcon}{Website}{resourceBench}
\DeclareTableResource{paper}{webarena}{https://arxiv.org/abs/2307.13854}{\PaperResourceIcon}{Paper}{resourcePaper}
\DeclareTableResource{code}{webarena}{https://github.com/web-arena-x/webarena}{\CodeResourceIcon}{Code}{resourceCode}
\DeclareTableResource{data}{webarena}{https://github.com/web-arena-x/webarena/tree/main/config_files}{\DataResourceIcon}{Data}{resourceData}
\DeclareTableResource{bench}{webarena}{https://webarena.dev/}{\BenchResourceIcon}{Website}{resourceBench}
\DeclareTableResource{paper}{wildclawbench}{https://arxiv.org/abs/2605.10912}{\PaperResourceIcon}{Paper}{resourcePaper}
\DeclareTableResource{code}{wildclawbench}{https://github.com/internlm/WildClawBench}{\CodeResourceIcon}{Code}{resourceCode}
\DeclareTableResource{data}{wildclawbench}{https://huggingface.co/datasets/internlm/WildClawBench}{\DataResourceIcon}{Data}{resourceData}
\DeclareTableResource{bench}{wildclawbench}{https://internlm.github.io/WildClawBench/}{\BenchResourceIcon}{Website}{resourceBench}
\DeclareTableResource{paper}{windowsarena}{https://arxiv.org/abs/2409.08264}{\PaperResourceIcon}{Paper}{resourcePaper}
\DeclareTableResource{code}{windowsarena}{https://github.com/microsoft/WindowsAgentArena}{\CodeResourceIcon}{Code}{resourceCode}
\DeclareTableResource{bench}{windowsarena}{https://microsoft.github.io/WindowsAgentArena/}{\BenchResourceIcon}{Website}{resourceBench}
\DeclareTableResource{paper}{workarena}{https://arxiv.org/abs/2403.07718}{\PaperResourceIcon}{Paper}{resourcePaper}
\DeclareTableResource{code}{workarena}{https://github.com/ServiceNow/WorkArena}{\CodeResourceIcon}{Code}{resourceCode}
\DeclareTableResource{data}{workarena}{https://huggingface.co/datasets/ServiceNow/WorkArena-Instances}{\DataResourceIcon}{Data}{resourceData}
\DeclareTableResource{bench}{workarena}{https://servicenow.github.io/WorkArena/}{\BenchResourceIcon}{Website}{resourceBench}
\DeclareTableResource{paper}{workarenapp}{https://arxiv.org/abs/2407.05291}{\PaperResourceIcon}{Paper}{resourcePaper}
\DeclareTableResource{code}{workarenapp}{https://github.com/ServiceNow/WorkArena}{\CodeResourceIcon}{Code}{resourceCode}
\DeclareTableResource{data}{workarenapp}{https://huggingface.co/datasets/ServiceNow/WorkArena-Instances}{\DataResourceIcon}{Data}{resourceData}
\DeclareTableResource{bench}{workarenapp}{https://servicenow.github.io/WorkArena/}{\BenchResourceIcon}{Website}{resourceBench}
\DeclareTableResource{paper}{workbench}{https://arxiv.org/abs/2405.00823}{\PaperResourceIcon}{Paper}{resourcePaper}
\DeclareTableResource{code}{workbench}{https://github.com/olly-styles/WorkBench}{\CodeResourceIcon}{Code}{resourceCode}
\DeclareTableResource{data}{workbench}{https://github.com/olly-styles/WorkBench/tree/main/data}{\DataResourceIcon}{Data}{resourceData}
\DeclareTableResource{paper}{workgenesis}{https://arxiv.org/abs/2609.39325}{\PaperResourceIcon}{Paper}{resourcePaper}
\DeclareTableResource{paper}{workspacebench}{https://arxiv.org/abs/2605.03596}{\PaperResourceIcon}{Paper}{resourcePaper}
\DeclareTableResource{code}{workspacebench}{https://github.com/OpenDataBox/Workspace-Bench}{\CodeResourceIcon}{Code}{resourceCode}
\DeclareTableResource{data}{workspacebench}{https://huggingface.co/datasets/Workspace-Bench/Workspace-Bench}{\DataResourceIcon}{Data}{resourceData}
\DeclareTableResource{bench}{workspacebench}{https://workspace-bench.github.io/}{\BenchResourceIcon}{Website}{resourceBench}
\DeclareTableResource{paper}{yin2026visionasinversegraphics}{https://arxiv.org/abs/2601.11109}{\PaperResourceIcon}{Paper}{resourcePaper}
\DeclareTableResource{code}{yin2026visionasinversegraphics}{https://github.com/Fugtemypt123/VIGA}{\CodeResourceIcon}{Code}{resourceCode}
\DeclareTableResource{data}{yin2026visionasinversegraphics}{https://huggingface.co/datasets/DietCoke4671/blenderbench}{\DataResourceIcon}{Data}{resourceData}
\DeclareTableResource{bench}{yin2026visionasinversegraphics}{https://fugtemypt123.github.io/VIGA-website/}{\BenchResourceIcon}{Website}{resourceBench}
\DeclareTableResource{paper}{yuan2026osworld2}{https://arxiv.org/abs/2606.29537}{\PaperResourceIcon}{Paper}{resourcePaper}
\DeclareTableResource{code}{yuan2026osworld2}{https://github.com/xlang-ai/OSWorld-V2}{\CodeResourceIcon}{Code}{resourceCode}
\DeclareTableResource{data}{yuan2026osworld2}{https://huggingface.co/datasets/xlangai/osworld_v2_tasks}{\DataResourceIcon}{Data}{resourceData}
\DeclareTableResource{bench}{yuan2026osworld2}{https://osworld-v2.xlang.ai/}{\BenchResourceIcon}{Website}{resourceBench}
\DeclareTableResource{paper}{zeng2026loca}{https://arxiv.org/abs/2602.07962}{\PaperResourceIcon}{Paper}{resourcePaper}
\DeclareTableResource{code}{zeng2026loca}{https://github.com/hkust-nlp/LOCA-bench}{\CodeResourceIcon}{Code}{resourceCode}
\DeclareTableResource{bench}{zeng2026loca}{https://loca-bench.github.io/}{\BenchResourceIcon}{Website}{resourceBench}
\DeclareTableResource{paper}{zhang2026frontierfinance}{https://arxiv.org/abs/2608.11683}{\PaperResourceIcon}{Paper}{resourcePaper}
\DeclareTableResource{code}{zhang2026frontierfinance}{https://github.com/samaya-ai/frontier-finance}{\CodeResourceIcon}{Code}{resourceCode}
\DeclareTableResource{data}{zhang2026frontierfinance}{https://huggingface.co/datasets/samaya-ai/FrontierFinance}{\DataResourceIcon}{Data}{resourceData}
\DeclareTableResource{bench}{zhang2026frontierfinance}{https://research.samaya.ai/benchmarks/frontier-finance}{\BenchResourceIcon}{Website}{resourceBench}
\definecolor{TableYesInk}{HTML}{79AC85}
\definecolor{TableNoInk}{HTML}{D68A8A}

\title{{\fontsize{26}{30}\selectfont\color{seedblue}Work Agents:}\\[1.5mm]
Systems, Capabilities, and Optimization}

\author[1,*]{Zishan~Xu}
\author[2,3,*]{Enjun~Du}
\author[4,*]{Mingqi~Gao}
\author[3]{Cheng~Yang}
\author[5]{Junhao~Shen}
\author[4]{Zirong~Chen}
\author[6]{Zhiyuan~Yao}
\author[7]{Yifu~Guo}
\author[7]{Yuquan~Lu}
\author[8]{Jinyang~Huang}
\author[6]{Zhengxi~Lu}
\author[3]{Siyi~Liu}
\author[2]{Ziyu~Zheng}
\author[3,\dagger]{Yongqi~Zhang}
\author[2,\dagger]{Difan~Zou}
\author[1,\dagger]{Weiwen~Liu}

\renewcommand{\affiliationlist}{%
  \begin{tabular}{@{}c@{}}
    \affiliationformat[1]{Shanghai Jiao Tong University}\quad
    \affiliationformat[2]{The University of Hong Kong}\\
    \affiliationformat[3]{The Hong Kong University of Science and Technology (Guangzhou)}\quad
    \affiliationformat[4]{Tsinghua University}\\
    \affiliationformat[5]{The Chinese University of Hong Kong}\hspace{4pt}%
    \affiliationformat[6]{Zhejiang University}\hspace{4pt}%
    \affiliationformat[7]{Sun Yat-sen University}\hspace{4pt}%
    \affiliationformat[8]{Central South University}
  \end{tabular}%
}

\contribution[*]{Equal contribution}
\contribution[\dagger]{Corresponding authors}
\gappto{\contributionlist}{%
  \par\vspace{0.7mm}%
  {\contributionfont
    \href{mailto:yongqizhang@hkust-gz.edu.cn}{\nolinkurl{yongqizhang@hkust-gz.edu.cn}}\quad
    \href{mailto:dzou@cs.hku.hk}{\nolinkurl{dzou@cs.hku.hk}}\quad
    \href{mailto:wwliu@sjtu.edu.cn}{\nolinkurl{wwliu@sjtu.edu.cn}}}}
\renewcommand{\projectlinklist}{%
  {\normalfont\fontfamily{droidserif}\fontsize{9}{11}\selectfont
    \href{https://github.com/EnjunDu/awesome-working-agent}{\faGithub\enspace Awesome Work Agents}%
    \quad\textcolor{tableMuted}{\textbar}\quad
    \href{https://enjundu.com/blog/working_agents}{\faGlobe\enspace Project Homepage}}%
}

\usepackage{calc}
\usepackage{wasysym}
\usepackage{etoolbox}
\usepackage{xurl}
\usepackage{needspace}

\titleformat*{\section}{\sectionfont\sffamily\bfseries\color{seedblue}}
\titleformat*{\subsection}{\subsectionfont\sffamily\bfseries\color{seedblue}}
\definecolor{level0}{HTML}{FFF5F0}
\definecolor{level1}{HTML}{FEE0D2}
\definecolor{level2}{HTML}{FCBBA1}
\definecolor{level3}{HTML}{FC9272}
\definecolor{level4}{HTML}{FB6A4A}
\definecolor{level5}{HTML}{EF3B2C}
\definecolor{level6}{HTML}{CB181D}
\renewcommand{\textbf}[1]{{\bfseries #1}}

\abstract{As foundation models evolve from information generation to digital work execution, agents are increasingly capable of completing complex tasks across tools, applications, and workflows. However, existing research lacks a unified perspective on how agents reliably accomplish delegated work. To address this gap, we present a work-centric, system-level perspective that views work agents as systems combining foundation models and runtime harnesses to execute tasks in stateful workspaces and deliver verifiable outcomes. We introduce two complementary dimensions, work persistence and work openness, to characterize working conditions, and classify digital work into five forms according to task scope and organization. Building on this framework, we identify five core capability families required for reliable work execution and examine approaches to model optimization, harness optimization, and their joint optimization. We further analyze existing evaluation methods and metrics from the perspective of work outcomes. Finally, we discuss future research directions, including sustained work delegation, adaptation to open environments, and the interaction between work execution and capability improvement. This paper provides a unified framework for understanding, optimizing, and evaluating work agents, supporting the development of systems capable of delivering reliable, verifiable, and practically valuable work outcomes.

{\fontsize{8.8}{10.2}\selectfont\textbf{Keywords:} Work Agents; Work Persistence and Openness; Agent Capabilities; Agent Optimization}}

\newtcolorbox{elementbox}{
  colback=tableHeader!50!white,colframe=tableRule,boxrule=0.5pt,
  grow to left by=6pt,grow to right by=6pt,
  arc=2pt,left=5.5pt,right=5.5pt,top=3.5pt,bottom=3.5pt,breakable}

\providecommand{\CoverStyle}{paper-journey-v6-reference-pins}
\def\CoverReferencePinsStyle{paper-journey-v6-reference-pins}
\def\CoverReferenceCurvesStyle{paper-journey-v6-reference-curves}
\def\CoverSixBezierStyle{paper-journey-v6-bezier}
\def\CoverJourneyStyle{paper-journey}
\def\CoverRisingStyle{paper-rising}
\def\CoverEvolutionStyle{cosmic-evolution}
\def\CoverWorkingStyle{cosmic-working}
\def\CoverRefinedStyle{cosmic-refined}
\newcommand{\CoverBackground}{figures/cover-variants/cover-\CoverStyle.png}
\newcommand{\CoverJourneyArt}{figures/cover-variants/cover-paper-journey-v6-clean.png}
\definecolor{coverInk}{HTML}{263D55}
\definecolor{coverMuted}{HTML}{53657B}
\definecolor{coverCleanBlue}{HTML}{527CB4}
\definecolor{coverCleanGold}{HTML}{B79A65}
\definecolor{coverCleanTeal}{HTML}{76A1AA}
\definecolor{coverCleanMist}{HTML}{CBE0EA}
\definecolor{coverCleanWarm}{HTML}{E1E1D8}
\newcommand{\CoverJourneyCleanFrame}{%
  \begin{scope}[line cap=round,line join=round]
  \clip (0,0) rectangle (1,1);
  \path[fill=coverCleanMist,fill opacity=0.25,draw=none]
    (0.14,1.04) .. controls (-0.07,0.87) and (-0.10,0.67) .. (0.045,0.50)
    .. controls (0.175,0.348) and (0.37,0.26) .. (0.58,0.18)
    .. controls (0.79,0.10) and (0.93,0.07) .. (1.04,0.07)
    -- (1.04,0.04)
    .. controls (0.93,0.04) and (0.77,0.075) .. (0.56,0.155)
    .. controls (0.35,0.235) and (0.15,0.315) .. (0.02,0.47)
    .. controls (-0.11,0.625) and (-0.10,0.86) .. (0.20,1.04) -- cycle;
  \path[fill=coverCleanMist,fill opacity=0.20,draw=none]
    (1.04,0.67) .. controls (0.91,0.57) and (0.95,0.46) .. (1.005,0.34)
    .. controls (1.06,0.20) and (1.00,0.07) .. (0.89,-0.04)
    -- (1.04,-0.04)
    .. controls (1.13,0.14) and (1.04,0.34) .. (1.00,0.46)
    .. controls (0.96,0.57) and (1.01,0.63) .. (1.04,0.67) -- cycle;
  \path[fill=coverCleanMist,fill opacity=0.28,draw=none]
    (-0.04,-0.015) .. controls (0.17,0.02) and (0.32,0.085) .. (0.53,0.15)
    .. controls (0.74,0.215) and (0.90,0.285) .. (1.04,0.31)
    -- (1.04,0.285)
    .. controls (0.90,0.260) and (0.74,0.190) .. (0.53,0.125)
    .. controls (0.32,0.060) and (0.17,-0.005) .. (-0.04,-0.04) -- cycle;
  \path[fill=coverCleanWarm,fill opacity=0.25,draw=none]
    (-0.04,0.163) .. controls (0.13,0.213) and (0.31,0.138) .. (0.49,0.103)
    .. controls (0.67,0.068) and (0.83,0.098) .. (1.04,0.223)
    -- (1.04,0.199)
    .. controls (0.83,0.074) and (0.67,0.044) .. (0.49,0.079)
    .. controls (0.31,0.114) and (0.13,0.189) .. (-0.04,0.139) -- cycle;
  \draw[coverCleanBlue!85,line width=0.46pt]
    (0.24,1.04) .. controls (-0.03,0.89) and (-0.13,0.70) .. (0.03,0.50)
    .. controls (0.15,0.35) and (0.37,0.27) .. (0.58,0.18)
    .. controls (0.79,0.09) and (0.96,0.07) .. (1.04,0.075);
  \draw[coverCleanGold!75,line width=0.40pt]
    (0.31,1.04) .. controls (0.02,0.91) and (-0.13,0.69) .. (0.05,0.49)
    .. controls (0.23,0.29) and (0.45,0.255) .. (0.67,0.16)
    .. controls (0.89,0.065) and (0.99,0.10) .. (1.04,0.15);
  \draw[coverCleanBlue!70,line width=0.42pt]
    (1.04,0.82) .. controls (0.90,0.68) and (0.91,0.56) .. (0.98,0.43)
    .. controls (1.05,0.30) and (1.04,0.19) .. (0.84,-0.04);
  \draw[coverCleanGold!60,line width=0.36pt]
    (1.04,0.73) .. controls (0.92,0.60) and (0.90,0.48) .. (0.99,0.35)
    .. controls (1.08,0.22) and (1.02,0.105) .. (0.93,-0.04);
  \draw[coverCleanBlue!85,line width=0.43pt]
    (-0.04,-0.015) .. controls (0.17,0.02) and (0.32,0.085) .. (0.53,0.15)
    .. controls (0.74,0.215) and (0.90,0.285) .. (1.04,0.31);
  \draw[coverCleanGold!80,line width=0.38pt]
    (-0.04,-0.04) .. controls (0.17,-0.005) and (0.32,0.060) .. (0.53,0.125)
    .. controls (0.74,0.190) and (0.90,0.260) .. (1.04,0.285);
  \draw[coverCleanTeal!80,line width=0.38pt]
    (-0.04,0.155) .. controls (0.13,0.205) and (0.31,0.13) .. (0.49,0.095)
    .. controls (0.67,0.06) and (0.83,0.09) .. (1.04,0.215);
  \draw[coverCleanBlue!55,line width=0.34pt]
    (-0.04,0.08) .. controls (0.14,0.02) and (0.30,0.06) .. (0.47,0.14)
    .. controls (0.64,0.22) and (0.82,0.245) .. (1.04,0.175);
  \end{scope}%
}

\definecolor{coverV6Blue}{HTML}{658BB3}
\definecolor{coverV6Teal}{HTML}{82A7AE}
\definecolor{coverV6Gold}{HTML}{C1A36F}
\definecolor{coverV6Mist}{HTML}{DCEAF0}
\definecolor{coverV6Warm}{HTML}{E6E3D7}
\newcommand{\CoverSixOuterCurves}{%
  \begin{scope}[line cap=round,line join=round]
  \clip (0,0) rectangle (1,1);
  \path[fill=coverV6Mist,fill opacity=0.36,draw=none]
    (0.17,1.04) .. controls (-0.10,0.86) and (-0.11,0.67) .. (0.035,0.495)
    .. controls (0.11,0.408) and (0.08,0.365) .. (-0.04,0.33)
    -- (-0.04,0.35)
    .. controls (0.06,0.38) and (0.085,0.415) .. (0.018,0.48)
    .. controls (-0.13,0.655) and (-0.10,0.86) .. (0.20,1.04) -- cycle;
  \draw[coverV6Blue!85,line width=0.43pt]
    (0.245,1.04) .. controls (-0.055,0.875) and (-0.13,0.685) .. (0.035,0.495)
    .. controls (0.11,0.408) and (0.08,0.365) .. (-0.04,0.33);
  \draw[coverV6Blue!48,line width=0.27pt]
    (0.225,1.04) .. controls (-0.075,0.87) and (-0.14,0.68) .. (0.027,0.488)
    .. controls (0.103,0.401) and (0.074,0.36) .. (-0.04,0.325);
  \draw[coverV6Gold!78,line width=0.36pt]
    (0.31,1.04) .. controls (-0.005,0.89) and (-0.12,0.665) .. (0.052,0.48)
    .. controls (0.13,0.396) and (0.10,0.37) .. (-0.04,0.35);
  \draw[coverV6Teal!55,line width=0.28pt]
    (0.175,1.04) .. controls (-0.125,0.86) and (-0.115,0.675) .. (0.018,0.515)
    .. controls (0.09,0.428) and (0.065,0.39) .. (-0.04,0.365);
  \draw[coverV6Gold!35,line width=0.23pt]
    (0.345,1.04) .. controls (0.03,0.895) and (-0.10,0.67) .. (0.065,0.485)
    .. controls (0.143,0.398) and (0.10,0.362) .. (-0.04,0.345);
  \draw[coverV6Blue!75,line width=0.37pt]
    (1.04,0.875) .. controls (0.895,0.72) and (0.91,0.565) .. (0.985,0.435)
    .. controls (1.06,0.305) and (1.01,0.13) .. (0.88,-0.04);
  \draw[coverV6Blue!42,line width=0.25pt]
    (1.04,0.845) .. controls (0.915,0.705) and (0.927,0.565) .. (0.995,0.44)
    .. controls (1.063,0.315) and (1.035,0.145) .. (0.91,-0.04);
  \draw[coverV6Gold!60,line width=0.31pt]
    (1.04,0.79) .. controls (0.925,0.665) and (0.93,0.545) .. (1.00,0.42)
    .. controls (1.07,0.295) and (1.045,0.14) .. (0.95,-0.04);
  \draw[coverV6Teal!42,line width=0.24pt]
    (1.04,0.925) .. controls (0.90,0.76) and (0.90,0.59) .. (0.975,0.46)
    .. controls (1.05,0.33) and (1.01,0.15) .. (0.86,-0.04);
  \end{scope}%
}
\newcommand{\CoverSixFlowCurves}{%
  \begin{scope}[line cap=round,line join=round]
  \clip (0,0) rectangle (1,1);
  \draw[coverV6Blue!65,line width=0.31pt]
    (-0.04,0.135) .. controls (0.11,0.15) and (0.29,0.155) .. (0.47,0.20)
    .. controls (0.65,0.245) and (0.70,0.29) .. (0.82,0.300)
    .. controls (0.94,0.31) and (1.00,0.30) .. (1.04,0.295);
  \draw[coverV6Teal!50,line width=0.24pt]
    (-0.04,0.126) .. controls (0.11,0.141) and (0.29,0.15) .. (0.47,0.195)
    .. controls (0.65,0.24) and (0.70,0.283) .. (0.82,0.293)
    .. controls (0.94,0.303) and (1.00,0.294) .. (1.04,0.29);
  \draw[coverV6Gold!45,line width=0.24pt]
    (-0.04,0.147) .. controls (0.11,0.162) and (0.29,0.164) .. (0.47,0.209)
    .. controls (0.65,0.254) and (0.70,0.296) .. (0.82,0.306)
    .. controls (0.94,0.316) and (1.00,0.305) .. (1.04,0.300);
  \path[fill=coverV6Mist,fill opacity=0.32,draw=none]
    (0.32,-0.04) .. controls (0.36,0.035) and (0.45,0.085) .. (0.59,0.095)
    .. controls (0.73,0.105) and (0.91,0.13) .. (1.04,0.09)
    -- (1.04,0.080)
    .. controls (0.91,0.12) and (0.73,0.095) .. (0.59,0.085)
    .. controls (0.45,0.075) and (0.36,0.025) .. (0.30,-0.04) -- cycle;
  \draw[coverV6Blue!67,line width=0.34pt]
    (0.32,-0.04) .. controls (0.36,0.035) and (0.45,0.085) .. (0.59,0.095)
    .. controls (0.73,0.105) and (0.91,0.13) .. (1.04,0.09);
  \draw[coverV6Blue!35,line width=0.24pt]
    (0.34,-0.04) .. controls (0.38,0.03) and (0.46,0.079) .. (0.60,0.089)
    .. controls (0.74,0.099) and (0.91,0.12) .. (1.04,0.08);
  \draw[coverV6Gold!58,line width=0.29pt]
    (0.29,-0.04) .. controls (0.35,0.025) and (0.44,0.073) .. (0.58,0.083)
    .. controls (0.72,0.093) and (0.91,0.11) .. (1.04,0.07);
  \draw[coverV6Teal!60,line width=0.30pt]
    (1.04,-0.015) .. controls (0.85,0.035) and (0.75,0.08) .. (0.58,0.085)
    .. controls (0.41,0.09) and (0.31,0.07) .. (0.27,-0.04);
  \draw[coverV6Gold!32,line width=0.23pt]
    (1.04,-0.029) .. controls (0.85,0.021) and (0.75,0.069) .. (0.58,0.074)
    .. controls (0.41,0.079) and (0.32,0.055) .. (0.29,-0.04);
  \end{scope}%
}
\newcommand{\CoverSixAnnotation}[6]{%
  \node (coverV6Stage) [anchor=north,align=left,text width=35mm,
    inner xsep=0pt,inner ysep=2pt,fill=none,draw=none]
    at (#1,#2) {%
      \hyphenpenalty=10000\exhyphenpenalty=10000\relax
      {\fontfamily{droidserif}\fontsize{10.2}{12.2}\selectfont\bfseries
        \textcolor{coverInk}{#5}}\\[2.5pt]
      {\fontfamily{droidserif}\fontsize{8.2}{10.3}\selectfont
        \textcolor{coverMuted}{#6}}};%
  \draw[draw=coverMuted!65,line width=0.4pt]
    (coverV6Stage.south) .. controls (#1,{#2-0.065})
    and (#3,{#4+0.016}) .. (#3,#4);%
  \fill[coverInk!65] (#3,#4) circle (0.85pt);%
}
\newcommand{\CoverSixBezierCover}{%
  \AtPageLowerLeft{%
    \begin{tikzpicture}[overlay,x=\paperwidth,y=\paperheight]
      \fill[white] (0,0) rectangle (1,1);
      \CoverSixOuterCurves
    \end{tikzpicture}%
  }%
  \AtPageLowerLeft{%
    \includegraphics[width=\paperwidth]{figures/cover-variants/cover-paper-journey-v6.png}%
  }%
  \AtPageLowerLeft{%
    \begin{tikzpicture}[overlay,x=\paperwidth,y=\paperheight]
      \CoverSixFlowCurves
      \CoverSixAnnotation{0.120}{0.220}{0.090}{0.110}
        {Conversation-level}{Interpret evidence\\Produce answers}
      \CoverSixAnnotation{0.300}{0.245}{0.290}{0.130}
        {Atomic-action}{Edit a local object\\Verify the change}
      \CoverSixAnnotation{0.480}{0.285}{0.425}{0.160}
        {Multi-step}{Manage dependencies\\Complete objectives}
      \CoverSixAnnotation{0.660}{0.320}{0.590}{0.205}
        {Workflow}{Coordinate stages\\and handoffs}
      \CoverSixAnnotation{0.850}{0.370}{0.780}{0.263}
        {Workspace}{Find related objects\\Follow dependencies}
    \end{tikzpicture}%
  }%
}

\newcommand{\CoverReferenceCurvesCover}{%
  \AtPageLowerLeft{%
    \makebox[\paperwidth][c]{%
      \includegraphics[width=\paperwidth,height=\paperheight]
        {figures/cover-variants/cover-bezier-reference-background-fit.png}}%
  }%
  \AtPageLowerLeft{%
    \includegraphics[width=\paperwidth]
      {figures/cover-variants/cover-paper-journey-v6-clean.png}%
  }%
  \AtPageLowerLeft{%
    \begin{tikzpicture}[overlay,x=\paperwidth,y=\paperheight]
      \CoverSixAnnotation{0.120}{0.220}{0.090}{0.110}
        {Conversation-level}{Interpret evidence\\Produce answers}
      \CoverSixAnnotation{0.300}{0.245}{0.290}{0.130}
        {Atomic-action}{Edit a local object\\Verify the change}
      \CoverSixAnnotation{0.480}{0.285}{0.425}{0.160}
        {Multi-step}{Manage dependencies\\Complete objectives}
      \CoverSixAnnotation{0.660}{0.320}{0.590}{0.205}
        {Workflow}{Coordinate stages\\and handoffs}
      \CoverSixAnnotation{0.850}{0.370}{0.780}{0.263}
        {Workspace}{Find related objects\\Follow dependencies}
    \end{tikzpicture}%
  }%
}

\input{figures/cover-variants/cover-reference-pins.tex}
\newcommand{\CoverLabelInk}[2]{\textcolor{#1}{#2}}
\newcommand{\CoverTypography}{%
  \renewcommand{\rmdefault}{droidserif}%
  \renewcommand{\sfdefault}{droidserif}%
  \normalfont
  \renewcommand{\affiliationformat}[2][]{{%
    \affiliationfont\fontfamily{droidserif}\selectfont $^{##1}$##2}}%
}
\newcommand{\CoverWorkLabel}[4][0.202]{%
  \node[anchor=north,align=center,text width=35mm,
    inner xsep=3pt,inner ysep=2pt,fill=none,draw=none]
    at (#2,#1) {%
      {\fontfamily{droidserif}\fontsize{10.2}{12.2}\selectfont\bfseries
        \CoverLabelInk{coverInk}{#3}}\\[2.2pt]
      {\fontfamily{droidserif}\fontsize{8.4}{10.2}\selectfont
        \CoverLabelInk{coverMuted}{#4}}};%
}
\newcommand{\CoverStageLeader}[3]{%
  \draw[draw=coverMuted!45,line width=0.35pt] (#1,#2)--(#1,#3);%
}
\newcommand{\CoverJourneyAnnotation}[4]{%
  \node [anchor=north,align=left,text width=37mm,
    inner xsep=0pt,inner ysep=2pt,fill=none,draw=none]
    at (#1,#2) {%
      \hyphenpenalty=10000\exhyphenpenalty=10000\relax
      {\fontfamily{droidserif}\fontsize{12.0}{14.2}\selectfont\bfseries
        \textcolor{coverInk}{#3}}\\[2.5pt]
      {\fontfamily{droidserif}\fontsize{9.3}{11.3}\selectfont
        \textcolor{coverMuted}{#4}}};%
}
\newcommand{\CoverJourneyBackdrop}{%
  \AtPageLowerLeft{%
    \begin{tikzpicture}[overlay,x=\paperwidth,y=\paperheight]
      \fill[white] (0,0) rectangle (1,1);
      \CoverJourneyCleanFrame
    \end{tikzpicture}%
  }%
  \AtPageLowerLeft{%
    \includegraphics[width=\paperwidth]{\CoverJourneyArt}%
  }%
}
\newcommand{\WorkingAgentCover}{%
  \edef\CoverSelectedStyle{\CoverStyle}%
  \edef\CoverLabelStyle{\CoverStyle}%
  \ifx\CoverSelectedStyle\CoverReferencePinsStyle
    \CoverReferencePinsCover
  \else
  \ifx\CoverSelectedStyle\CoverReferenceCurvesStyle
    \CoverReferenceCurvesCover
  \else
  \ifx\CoverSelectedStyle\CoverSixBezierStyle
    \CoverSixBezierCover
  \else
  \ifx\CoverSelectedStyle\CoverRefinedStyle
    \let\CoverSelectedStyle\CoverEvolutionStyle
  \fi
  \ifx\CoverSelectedStyle\CoverWorkingStyle
    \let\CoverSelectedStyle\CoverEvolutionStyle
  \fi
  \ifx\CoverSelectedStyle\CoverJourneyStyle
    \CoverJourneyBackdrop
  \else
  \AtPageLowerLeft{%
    \makebox[\paperwidth][c]{%
      \ifx\CoverSelectedStyle\CoverEvolutionStyle
      \raisebox{-0.04\paperheight}{%
        \includegraphics[height=1.04\paperheight]{\CoverBackground}}%
      \else
      \includegraphics[width=\paperwidth,height=\paperheight,keepaspectratio]{\CoverBackground}%
      \fi}%
  }%
  \fi
  \AtPageLowerLeft{%
    \begin{tikzpicture}[overlay,x=\paperwidth,y=\paperheight]
      \ifx\CoverLabelStyle\CoverJourneyStyle
      \CoverJourneyAnnotation{0.120}{0.238}
        {Conversation-level}{Interpret evidence\\Produce answers}
      \CoverJourneyAnnotation{0.300}{0.267}
        {Atomic-action}{Edit a local object\\Verify the change}
      \CoverJourneyAnnotation{0.480}{0.337}
        {Multi-step}{Manage dependencies\\Complete objectives}
      \CoverJourneyAnnotation{0.660}{0.360}
        {Workflow}{Coordinate stages\\and handoffs}
      \CoverJourneyAnnotation{0.860}{0.380}
        {Workspace}{Find related objects\\Follow dependencies}
      \else
      \ifx\CoverLabelStyle\CoverRisingStyle
      \CoverWorkLabel[0.070]{0.120}{Conversation-level}{Answers and analysis}
      \CoverWorkLabel[0.100]{0.312}{Atomic-action}{Local changes}
      \CoverWorkLabel[0.110]{0.500}{Multi-step}{Dependent actions}
      \CoverWorkLabel[0.115]{0.650}{Workflow}{Stages and handoffs}
      \CoverWorkLabel[0.150]{0.903}{Workspace}{Connected work objects}
      \else
      \ifx\CoverSelectedStyle\CoverEvolutionStyle
      \ifx\CoverLabelStyle\CoverRefinedStyle
      \CoverStageLeader{0.120}{0.065}{0.063}
      \CoverStageLeader{0.300}{0.122}{0.105}
      \CoverStageLeader{0.490}{0.135}{0.125}
      \CoverStageLeader{0.680}{0.170}{0.150}
      \CoverStageLeader{0.895}{0.195}{0.160}
      \CoverWorkLabel[0.060]{0.120}{Conversation-level}{Answers and analysis}
      \CoverWorkLabel[0.100]{0.300}{Atomic-action}{Local changes}
      \CoverWorkLabel[0.120]{0.490}{Multi-step}{Dependent actions}
      \CoverWorkLabel[0.145]{0.680}{Workflow}{Stages and handoffs}
      \CoverWorkLabel[0.155]{0.895}{Workspace}{Connected work objects}
      \else
      \CoverStageLeader{0.120}{0.083}{0.075}
      \CoverStageLeader{0.300}{0.142}{0.120}
      \CoverStageLeader{0.490}{0.161}{0.150}
      \CoverStageLeader{0.680}{0.194}{0.180}
      \CoverStageLeader{0.870}{0.234}{0.215}
      \CoverWorkLabel[0.070]{0.120}{Conversation-level}{Answers and analysis}
      \CoverWorkLabel[0.115]{0.300}{Atomic-action}{Local changes}
      \CoverWorkLabel[0.145]{0.490}{Multi-step}{Dependent actions}
      \CoverWorkLabel[0.175]{0.680}{Workflow}{Stages and handoffs}
      \CoverWorkLabel[0.210]{0.870}{Workspace}{Connected work objects}
      \fi
      \else
      \CoverWorkLabel{0.120}{Conversation-level}{Answers and analysis}
      \CoverWorkLabel{0.312}{Atomic-action}{Local changes}
      \CoverWorkLabel{0.500}{Multi-step}{Dependent actions}
      \CoverWorkLabel{0.710}{Workflow}{Stages and handoffs}
      \CoverWorkLabel{0.903}{Workspace}{Connected work objects}
      \fi
      \fi
      \fi
    \end{tikzpicture}%
  }%
  \fi
  \fi
  \fi
}

\begin{document}
\newgeometry{top=1.5cm,bottom=1.6cm,left=2.5cm,right=2.5cm,footskip=20pt}
\AddToShipoutPictureBG*{\WorkingAgentCover}
\begingroup
\CoverTypography
\maketitle
\vfill
\clearpage
\endgroup
\restoregeometry
\tableofcontents
\newpage
\FloatBarrier
\FloatBarrier
\section{Introduction}
\label{sec:introduction}



Advances in foundation models’ reasoning, planning, and tool-use capabilities are expanding the scope of work that agents can undertake~\cite{foundationmodels}. Their responsibilities increasingly extend beyond short, narrowly scoped tasks to larger assignments that require them to maintain an overall objective, coordinate multiple stages, and carry the work through to completion. Furthermore, recursive self-improvement~\cite{dreamrsi} explores how agents can improve their own models and runtime systems, bringing capability improvement itself into the range of work agents undertake.

Humans are already delegating complex productivity tasks to agents across software engineering~\cite{sweagent}, office work~\cite{sheetcopilot,pptagent}, web and computer use~\cite{webarena,osworld,cogagent}, enterprise operations~\cite{workarena}, and scientific research~\cite{scienceagentbench}. Frontier agents such as Codex and Claude Code~\cite{codexdocs,claudedocs} illustrate this shift by supporting multi-day, cross-session work with suitable orchestration and taking on parts of result assessment and feedback-driven revision previously handled by humans. Across these tasks, decisions often depend on evidence in related files, records, or applications and on states established by earlier actions. A work-centric perspective traces these dependencies from the initial request through execution to a verifiable outcome.

From this perspective, we introduce the concept of a \textbf{work agent}, defined as a system that uses a foundation model as its decision-making core, relies on a runtime harness to organize environment observations and action execution, and carries out delegated tasks within a stateful digital workspace. The foundation model provides understanding, reasoning, and decision-making capabilities; the harness manages context construction, tool invocation, memory, skills, execution control, and error recovery; and the workspace contains interconnected objects such as files, applications, records, tasks, and other actors. A work agent aims to produce work outcomes that satisfy task requirements, with execution results verified through observable evidence. This perspective connects task objectives, model capabilities, system mechanisms, environment states, and verification, shifting the unit of analysis from isolated model outputs or tool calls to the complete process of accomplishing digital work.

Work agents operate under different conditions depending on the tasks they undertake. To characterize these differences, we introduce two conceptually distinct yet complementary dimensions: \circlednum{182}~\textbf{work persistence} and \circlednum{183}~\textbf{work openness}. Work persistence concerns the conditions under which delegated responsibility can be considered complete, distinguishing work that concludes upon delivery from responsibilities that remain active across multiple stages, feedback cycles, or external events. Work openness describes the extent to which task-relevant objects, information, interfaces, and rules are specified before execution, or must instead be discovered and interpreted during execution. These working conditions impose different demands on an agent's ability to understand its environment, manage task dependencies, maintain execution state, and verify outcomes. Based on the functional requirements that arise throughout the work process, we organize the core capabilities of work agents into five interrelated families: perception and understanding, reasoning and decision making, planning and coordination, execution and state management, and verification, recovery, and replanning. Together, these capabilities support the complete process of interpreting work requirements, organizing and executing actions, and verifying results, forming an iterative execution cycle driven by environmental feedback.

The performance of work agents depends not only on their core capabilities but also on how foundation models and runtime harnesses work together. From an optimization perspective, some capabilities can be learned by models through training data, interaction experience, and feedback, while others can be supported by external mechanisms such as tool interfaces, memory management, and execution control. Improving work agents therefore requires considering model optimization, harness optimization, and their joint optimization. From an evaluation perspective, agent performance is reflected not only in the correctness of answers or individual actions but also in the quality of work artifacts, changes to environment states, workflow completion, and satisfaction of task requirements. Systematic evaluation across these dimensions provides a more comprehensive understanding of the practical performance and reliability of work agents.

As agents move from executing bounded tasks toward sustained work delegation, their responsibilities increasingly extend across multiple stages and events. In this setting, agents must maintain continuity despite environmental changes, interruptions, and external feedback, while also leveraging experience accumulated during execution to improve their capabilities without disrupting ongoing responsibilities. Work execution and capability improvement thus form two interacting loops, raising new research challenges for the long-term development of work agents. Personal work agents, human digital twins, and organizational work systems provide research settings for sustaining goals between requests, representing a particular person's judgment, and coordinating responsibilities across roles. Future research must address reliability, environmental adaptation, and outcome verification during sustained execution, while also considering the practical value of work outcomes and the human and computational resources required to produce them. These challenges will shape the transition from completing individual tasks to undertaking more persistent, complex, and valuable digital work.

Based on this perspective, the main contributions of this paper are summarized as follows:

\begin{itemize}
    \item \textbf{A unified, work-centric framework for agent systems.}
    We present a unified perspective centered on digital work, characterizing the structure and execution of work agents through six elements: task, model, harness, workspace, work outcome, and verification. We further describe the scope, organization, and execution conditions of digital work through five work forms and two complementary dimensions, work persistence and work openness.

    \item \textbf{A systematic account of core capabilities and optimization for reliable work.}
    We identify five core capability families required by work agents and examine their roles under different working conditions. We further examine training environments, data and reward design, model optimization, harness optimization, and their joint optimization, highlighting how model learning and external system support contribute to reliable work execution.

    \item \textbf{An outcome-oriented perspective on agent evaluation.}
    We systematically analyze existing benchmarks, evaluation metrics, judging methods, and execution environments, examining how they assess work artifact quality, environment-state changes, task completion, and constraint satisfaction. This provides a structured perspective for understanding the practical performance and reliability of work agents.

    \item \textbf{Future research directions toward sustained work delegation.}
    We explore emerging applications, including personal work agents, human digital twins, and organizational work systems, and discuss key challenges involving persistent responsibility, memory and adaptation, outcome verification, and the value of work in open environments. We further examine the interaction between work execution and capability improvement to outline research directions for next-generation work agents.
\end{itemize}

This paper focuses on digital knowledge work across office productivity, professional services, enterprise operations, software engineering, and scientific research. Section~\ref{foundation} introduces the foundations of work agents, including their system components, working conditions, and core capabilities. Section~\ref{workforms} examines five forms of digital work undertaken by agents. Section~\ref{optimization} examines training environments, model optimization, harness optimization, and their joint optimization. Section~\ref{evaluation} covers evaluation benchmarks, methodologies, metrics, and execution environments. Section~\ref{frontiers} discusses future applications, technical frontiers, and the interaction between work execution and capability improvement. Section~\ref{conclusion} concludes the paper.

\FloatBarrier
\FloatBarrier

\section{Foundation of Work Agents}\label{foundation}

Section~\ref{sec:introduction} argued that the performance of a system that carries out digital work depends not only on its model but also on its interfaces, the information available to it, the organization of its state, and the verification of its results. Studying such systems therefore requires a common account of what the system consists of, the conditions under which it works, what it must be able to do, and how it differs from related lines of agent research. This section develops this account in four steps. Section~\ref{sec:foundation-system-view} identifies the elements involved when digital work is delegated to an agent and defines a work agent. Section~\ref{sec:foundation-working-conditions} introduces two dimensions of working conditions, work persistence and work openness, which describe when a responsibility may close and how much the agent must find out before and while it acts. Section~\ref{capabilities} identifies five core capabilities that the work process requires and examines how the two dimensions raise their demands. Section~\ref{related-paradigms} compares work agents with long-horizon, computer-use, and office agents.

\subsection{Definition of Work Agents}\label{sec:foundation-system-view}
\label{sec:foundation-definition-scope}\label{sec:foundation-six-elements}\label{formalization}

\begin{figure}[!t]
\centering
\includegraphics[width=1.0\linewidth]{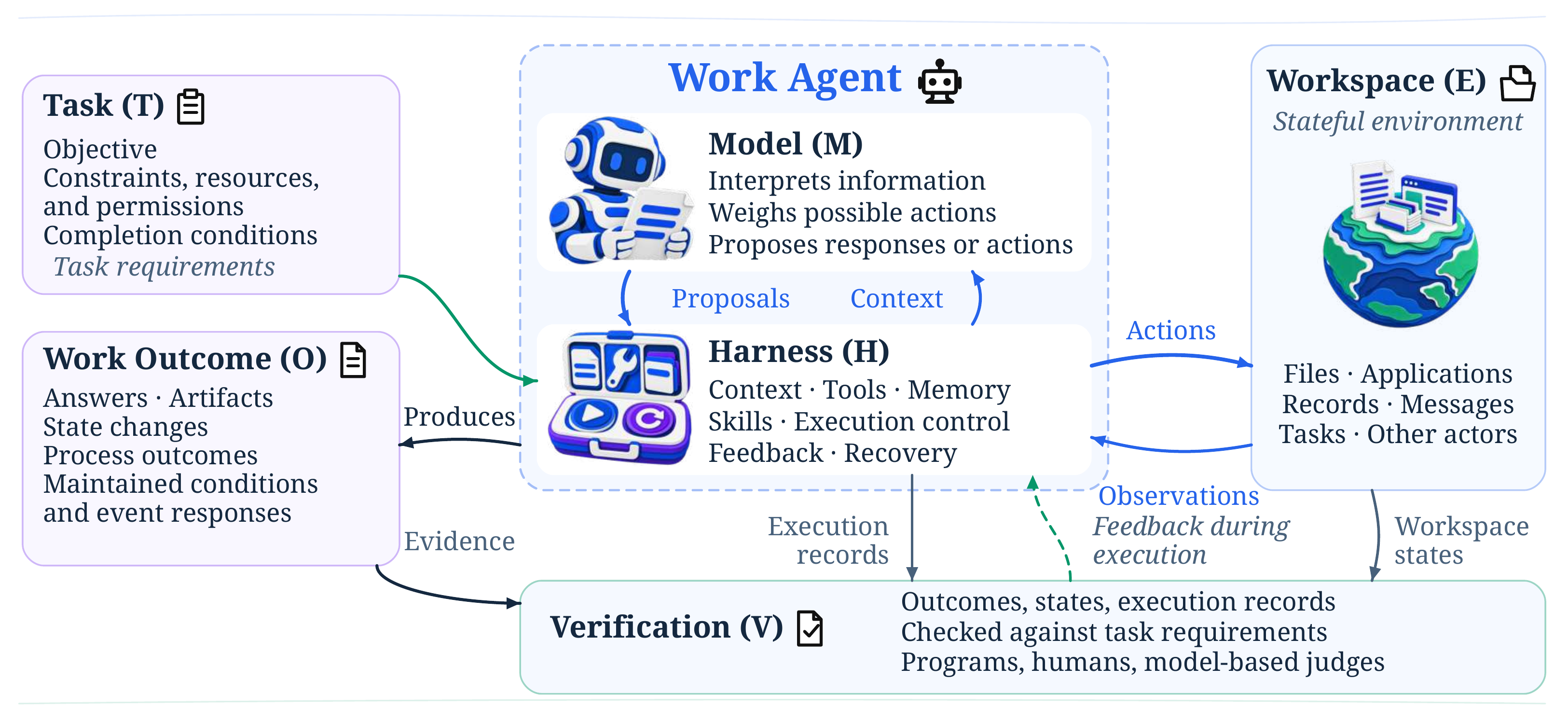}
\caption{A system view of work agents. The model and harness form the work agent within the dashed boundary. The task specifies the objective, constraints, resources, permissions, and completion conditions. The harness supplies context to the model, carries out proposed actions in the stateful workspace, and receives observations. Outcomes include answers, artifacts, state changes, and process outcomes. For ongoing responsibilities, they also include maintained conditions and event responses over an agreed interval. Verification assesses evidence from these outcomes, workspace states, and execution records against task requirements. Programs, humans, or model-based judges apply task-specific criteria. Verification may occur during execution, with feedback informing subsequent decisions, or afterwards to assess the work.}
\label{fig:working-agent-formalization}
\end{figure}

We begin with the system itself. By digital work we mean work carried out on digital objects such as files, applications, and records, for example updating a report or handling a service request. When a person or an organization hands such work to an agent, three things happen: the requester states what should be done, the agent acts on these objects to do it, and the result is checked against what was asked. We call such work delegated: the agent carries it out on the requester's behalf and is accountable for the result. We first describe the six elements involved in this process, as shown in Figure~\ref{fig:working-agent-formalization}, then define a work agent in terms of them, and finally describe how the elements interact during execution.

\begin{elementbox}
\textcolor{seedblue}{\textbf{Task ($T$)}} specifies the objective, the relevant constraints, the available resources, the permissions, and the conditions for valid completion. For an ongoing responsibility, it also specifies the conditions to maintain, the agreed interval, and the criteria for closure or renewal.

\smallskip
\textcolor{seedblue}{\textbf{Model ($M$)}} serves as the system's decision-making core. Given context supplied by the harness, it interprets the available information, weighs possible courses of action, and proposes responses or actions~\cite{react,coala}.

\smallskip
\textcolor{seedblue}{\textbf{Harness ($H$)}} organizes context construction, tool use, memory, skills, execution control, feedback handling, and recovery, mediating between the model and the workspace~\cite{coala}.

\smallskip
\textcolor{seedblue}{\textbf{Workspace ($E$)}} is the stateful environment in which work takes place. It contains files, applications, records, messages, tasks, and other actors. These are linked to one another, and the agent accesses them through interfaces within its permissions. Both the agent's actions and external events change its state as work proceeds.

\smallskip
\textcolor{seedblue}{\textbf{Work Outcome ($O$)}} is the task-relevant result that the agent produces. Outcomes include answers, such as analyses, plans, and executable specifications; artifacts such as documents, spreadsheets, presentations, and code; state changes in applications or business records, such as updated records and sent messages; process outcomes, such as a completed process or the stages, approvals, and handoffs finished so far; and maintained conditions and event responses over an agreed interval. One assignment may combine several outcomes, such as an updated workbook, a report, and the corresponding changes to business records, and require them to remain consistent across related artifacts, records, and applications.

\smallskip
\textcolor{seedblue}{\textbf{Verification ($V$)}} assesses work against task requirements. It inspects evidence from answers, workspace states, artifacts, and workflow records (Section~\ref{evaluation-evidence}). Programs, humans, or model-based judges~\cite{agentjudge} assess this evidence against task-specific criteria, which may be organized as rules, rubrics, or checklists (Section~\ref{judging-procedures}). The checks cover task completion, correctness, completeness, quality, constraint satisfaction, and intended effects. For example, SWE-bench~\cite{swebench} executes tests against a modified repository, whereas PresentBench~\cite{presentbench} applies task-specific checklist items through a model judge. Verification may be performed by the agent during execution (Section~\ref{verification-recovery}) or after execution by the requester or a benchmark (Section~\ref{evaluation}); its judgments support completion decisions, quality measures, and diagnostic feedback for subsequent work.
\end{elementbox}

With these elements, we define a \textbf{work agent} as a system that uses a foundation model as its decision-making core and a harness to organize observations and actions in a stateful workspace. It pursues a delegated work objective under specified constraints, resources, and permissions, and its outcomes are assessed against task requirements through observable evidence. The agent itself consists of the model and the harness, shown as the dashed box in Figure~\ref{fig:working-agent-formalization}; the task, workspace, outcome, and verification define the work for which it is accountable. The definition covers a wide range of tasks. In the minimal case, a question-answering system counts as a work agent when answering is delegated to it and its answers are checked against task requirements. Its harness only assembles the question and the supplied material into context and returns the model's answer, and its workspace is that material, which the agent reads but does not change. A system that edits workbooks or carries business processes to completion uses the same elements with a fuller harness and a workspace that its actions change (Section~\ref{workforms}).

The six elements interact in a loop during execution. The harness constructs context from the task, workspace observations, and retained execution information and passes it to the model. The model proposes responses or actions; the harness carries out the actions, receives observations, and uses them with verification feedback to update the context for the next decision. At an assessment point $t$, execution and verification have distinct roles:
\begin{equation}
(T, E_0) \xrightarrow[\text{closed-loop execution}]{M,H} (\tau_{0:t}, E_t, O_t),
\label{eq:work-execution}
\end{equation}
\begin{equation}
S_t = V(T, E_0, \tau_{0:t}, E_t, O_t).
\label{eq:work-verification}
\end{equation}
Equation~\eqref{eq:work-execution} summarizes execution by the model--harness system; Equation~\eqref{eq:work-verification} defines an assessment of its evidence against the task requirements. Here, $E_0$ is the initial workspace state, $\tau_{0:t}$ is the observable sequence of actions and feedback up to $t$, $E_t$ is the workspace state at $t$, and $O_t$ is the outcome produced so far. The assessment $S_t$ records how the inspected evidence meets the task requirements. Assessment can mark completion or occur during an ongoing responsibility.

All work-agent settings share these elements and this loop; what differs is the conditions under which the loop runs, which Section~\ref{sec:foundation-working-conditions} examines.

\subsection{Working Conditions}\label{sec:foundation-working-conditions}

Across tasks, these conditions differ in two respects: whether the work ends once the agent delivers its result, and how much the agent must find out before and while it acts. We capture these two differences as \circlednum{182}~work persistence and \circlednum{183}~work openness. Section~\ref{work-persistence} defines work persistence and traces it through existing research, Section~\ref{work-openness} does the same for work openness, and Section~\ref{sec:foundation-conditions-combined} shows that the two dimensions are conceptually distinct, illustrates their four combinations, and relates them to the forms of work examined in Section~\ref{workforms}.

\subsubsection{Work Persistence}\label{work-persistence}

The first difference concerns when the agent may consider delegated work finished. In some tasks, the work is finished once the agent delivers its result, such as an answer, a confirmed change, or a tested patch. In others, the agent's delivery alone does not finish the work: other participants must still assess, approve, or contribute, or later events require earlier results to be revisited. \textbf{Work persistence} thus ranges from responsibility that closes at delivery to responsibility that remains open across stages and events. As persistence increases, the agent must carry state across waiting periods and events and must determine at each point which parts of its responsibility are still open.

Existing research covers both sides of this range. Responsibility closes at delivery when closure depends only on what the agent itself delivers. In early work on reading comprehension and question answering over text, tables, documents, and charts~\cite{rajpurkar2016squad100000questionsmachine,yang2018hotpotqadatasetdiverseexplainable,tatqa,finqa,docvqa,chartqa}, responsibility closes once an answer is returned. Tool-use research, from Toolformer~\cite{toolformer} and API-Bank~\cite{apibank} to Gorilla~\cite{gorilla}, extends the output from text to tool invocation. When an action changes the workspace, the work also requires a confirmed effect: in AndroidWorld~\cite{androidworld}, a messaging task counts as complete only when the intended message is recorded in the phone's SMS database. Responsibility can also close at delivery after a sequence of dependent actions. In SWE-bench~\cite{swebench}, producing a patch involves localization, editing, and testing; in MINT~\cite{mint}, tool outputs and language feedback across turns guide subsequent actions.

On the other side, closure depends on more than the agent's own delivery: on other participants or on later events. When it depends on other participants, responsibility cannot close until they provide the confirmation or contribution the task requires: $\tau$-bench~\cite{taubench} withholds specified database updates until the user confirms them, $\tau^2$-bench~\cite{tau2bench} requires the agent to coordinate with a simulated user who carries out part of the work, and TheAgentCompany~\cite{agentcompany} includes tasks that cannot proceed without information from simulated coworkers. When it depends on later events, earlier results or completed stages may need to be revisited: in OSWorld 2.0~\cite{yuan2026osworld2}, revised budget and vendor constraints arrive through a team chat and require the agent to reconsider affected entries in a purchase-order form. When the task defines an obligation over an agreed interval, the agent remains responsible for responding to such events throughout that interval: E-Commerce Bench~\cite{ecommercebench} spans a simulated year of operational decisions, and VibeLifeBench~\cite{vibelife} evaluates proactive and persistent assistance as events change the environment over simulated weeks.

Dividing the examples by what closure depends on also separates persistence from task difficulty and execution length. Answering a question can demand harder reasoning than submitting a form for approval, yet responsibility for the answer closes at delivery, whereas the workflow remains open until approval is received. The same holds for execution length, a central concern of long-horizon research (Section~\ref{sec:foundation-long-horizon}). A long run may pass through many dependent steps, yet its responsibility still closes when the result is submitted; an obligation to keep a service within its operating constraints may require only a short run each week, yet its responsibility remains open for a quarter. The two examples sit on opposite sides of the boundary: the first closes once its run delivers a result, whereas the second stays open across many runs until the agreed interval ends.

\subsubsection{Work Openness}\label{work-openness}

The two examples at the end of Section~\ref{work-persistence} illustrate when responsibility may close, but do not specify how much the agent must find out before and while it acts. The long run may operate on a repository whose relevant files are named in the request, or on one whose relevant files it must first locate; the weekly run may read the same fixed metrics each time, or have to trace which components a new incident affects. In some tasks, the task world, that is, the objects, interfaces, rules, and actors relevant to the task, is specified with the request. In others, the agent must establish during execution which objects are relevant and how they are represented and related. It may also need to find out where the relevant state is held, how other actors have changed it, and which actions it is authorized to take. \textbf{Work openness} thus ranges from a predefined task world to one discovered during execution. As openness increases, completing the work depends more on the agent's ability to build and update its understanding of the task world while acting in it.

As with persistence, existing research spans this range, and the examples below are ordered by how much the agent must establish during execution. Early research largely studied predefined task worlds. Reading-comprehension and question-answering tasks supply the passage, table, document, or chart to be analyzed~\cite{rajpurkar2016squad100000questionsmachine,yang2018hotpotqadatasetdiverseexplainable,tatqa,finqa,docvqa,chartqa}, and tool-use studies such as API-Bank~\cite{apibank}, Gorilla~\cite{gorilla}, and BFCL~\cite{patil2025bfcl} expose an explicit set of tools and interfaces. The agent reasons or acts within a space that has already been defined, and the difficulty lies in using the supplied material and tools correctly rather than in finding them.

Finding them becomes part of the task in web and computer-use environments. In WebArena~\cite{webarena}, the information needed for a task must be obtained by navigating and interacting with functional websites rather than read from the prompt. OSWorld~\cite{osworld} extends the setting to real computer environments, where the agent works across applications and heterogeneous digital objects. In Workspace-Bench~\cite{workspacebench}, the agent must also identify which files are relevant and how they depend on one another, so locating material across a workspace becomes the core of the work.

Beyond objects, other environments make the state of the task world, the actions of other actors, and authority part of what must be established. AppWorld~\cite{appworld} spreads the information a task needs across multiple applications, so the agent must determine which applications and records hold it before acting. In $\tau^2$-bench~\cite{tau2bench}, a simulated user acts on parts of the shared environment that the agent cannot operate, so the agent must find out from the user what has changed. Authority must be established when the request does not state who may do what. In EntCollabBench~\cite{entcollabbench}, instructions describe business goals rather than workflows, and each agent may use only its own department's tools, so the agent must infer which role is responsible for each step and delegate to it. Applicable permissions may likewise have to be discovered during execution: in a card-unlocking task in $\tau$-Knowledge's $\tau$-Banking domain~\cite{tauknowledge}, the agent must retrieve the relevant policy and inspect the card's status to determine that a security hold prevents it from unlocking the card and requires it to offer escalation to the security team.

\subsubsection{Combining Persistence and Openness}\label{sec:foundation-conditions-combined}

Having defined both dimensions, we now consider how they combine. The two dimensions answer different questions, as the two examples at the start of Section~\ref{work-openness} already suggest: openness concerns what the agent must find out, whereas persistence concerns when it may stop. The same workspace change can involve both dimensions: discovering which objects it affects concerns openness, while responsibility that remains open as further changes arrive concerns persistence. The two dimensions are thus conceptually distinct: a setting can be high on one and low on the other.

\begin{figure}[t]
  \centering
  \includegraphics[width=\linewidth]{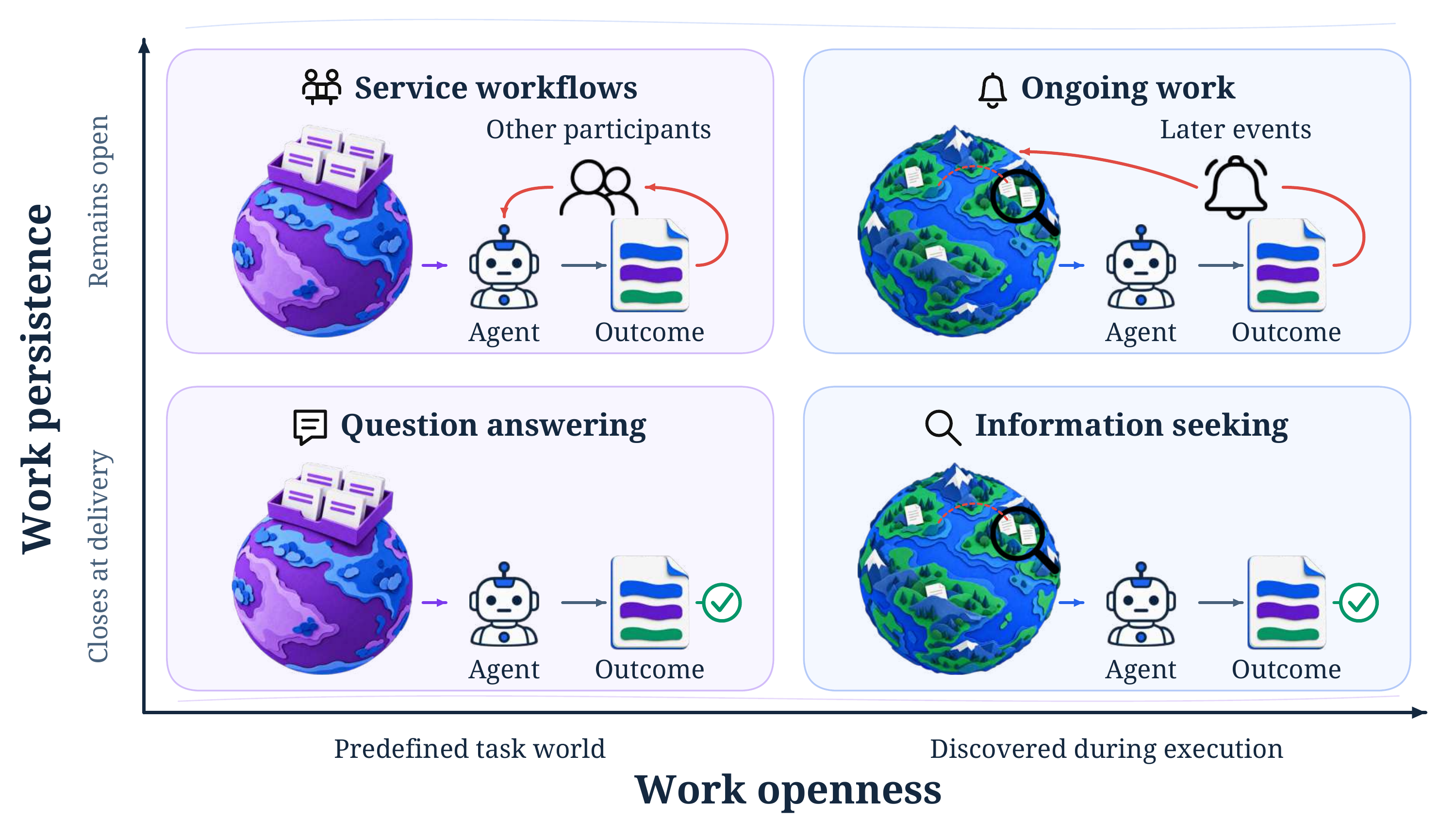}
  \caption{Work persistence and work openness as conceptually distinct dimensions of working conditions. Persistence ranges from responsibility that closes at delivery to responsibility that remains open across stages and events; openness ranges from a predefined task world to one discovered during execution. The four scenes show question answering (bottom left), information seeking (bottom right), service workflows (top left), and ongoing work under changing conditions (top right). Supplied packets distinguish the predefined task worlds; scattered material and magnifiers indicate that relevant objects and relationships must be found during execution. The same compact Agent marker appears in every scene. Checks at the terminal end mark responsibility that closes at delivery. Return paths through other participants (top left) or later events (top right) mark responsibility that remains open. The upper scenes include dependencies within bounded tasks as well as obligations extending beyond an initial delivery.}
  \label{fig:working-conditions}
\end{figure}

As shown in Figure~\ref{fig:working-conditions}, all four combinations appear in existing research. Question answering (bottom left) combines a predefined task world with responsibility that closes at delivery: the system answers from given material, as in SQuAD~\cite{rajpurkar2016squad100000questionsmachine} and DocVQA~\cite{docvqa}. Information seeking (bottom right) asks the agent to discover the relevant material during execution, yet responsibility still closes at delivery, as in GAIA~\cite{gaia}, whose questions require finding a source on the web, and Workspace-Bench~\cite{workspacebench}. Service workflows (top left) keep the task world predefined, yet responsibility remains open while the requested updates await user confirmation, as in $\tau$-bench~\cite{taubench}, where the tools, domain policy, and database are specified in advance. Ongoing work under changing conditions (top right) combines a task world discovered during execution with responsibility that remains open as later events arrive: revised constraints arriving mid-task in OSWorld 2.0~\cite{yuan2026osworld2} require the agent to find the affected entries, and in VibeLifeBench~\cite{vibelife}, some changes come without notification, so the agent must re-inspect the environment to discover them.

These combinations describe the conditions of work rather than its form. Section~\ref{workforms} distinguishes five work forms by the scope and organization of delegated work, from work at the Conversation Level, which answers questions over supplied material, to work at the Workspace Level, which operates across related objects in a workspace. Persistence and openness characterize the conditions under which each form is carried out: work at the Workflow Level, for example, can run in the predefined task world of $\tau$-bench~\cite{taubench} or require the agent to track a shared state that a simulated user also changes, as in $\tau^2$-bench~\cite{tau2bench}. Because both dimensions change what the agent must find out and keep track of, they also determine how demanding its core capabilities become, which Section~\ref{capabilities} examines.

\subsection{Core Capabilities of Work Agents}\label{sec:foundation-capabilities}\label{capabilities}

Whatever the working conditions of Section~\ref{sec:foundation-working-conditions}, carrying out the work process of Section~\ref{sec:foundation-system-view} raises five questions, each answered by one capability that recurs across office applications~\cite{sheetcopilot,pptagent}, software repositories~\cite{swebench,sweagent}, enterprise services~\cite{workarena,taubench}, and professional workspaces~\cite{workspacebench,agentcompany}: (1) what the work involves: Perception and Understanding interprets the objects and their relationships; (2) what to do next: Reasoning and Decision Making weighs goals, constraints, and evidence to decide how to proceed; (3) how dependent steps and contributions fit together: Planning and Coordination arranges dependencies and handoffs among steps and participants; (4) what has actually changed: Execution and State Management acts on the workspace through tools and interfaces while recording progress; and (5) whether the results satisfy the requirements: Verification, Recovery, and Replanning confirms outcomes and guides correction. Each capability is a function of the work process that the model, the harness, or their interaction can carry out, and in a simple task several are carried out in one step, as when a question-answering system reads the supplied material and returns a checked answer.

\begin{figure}[!t]
  \centering
  \includegraphics[width=\linewidth]{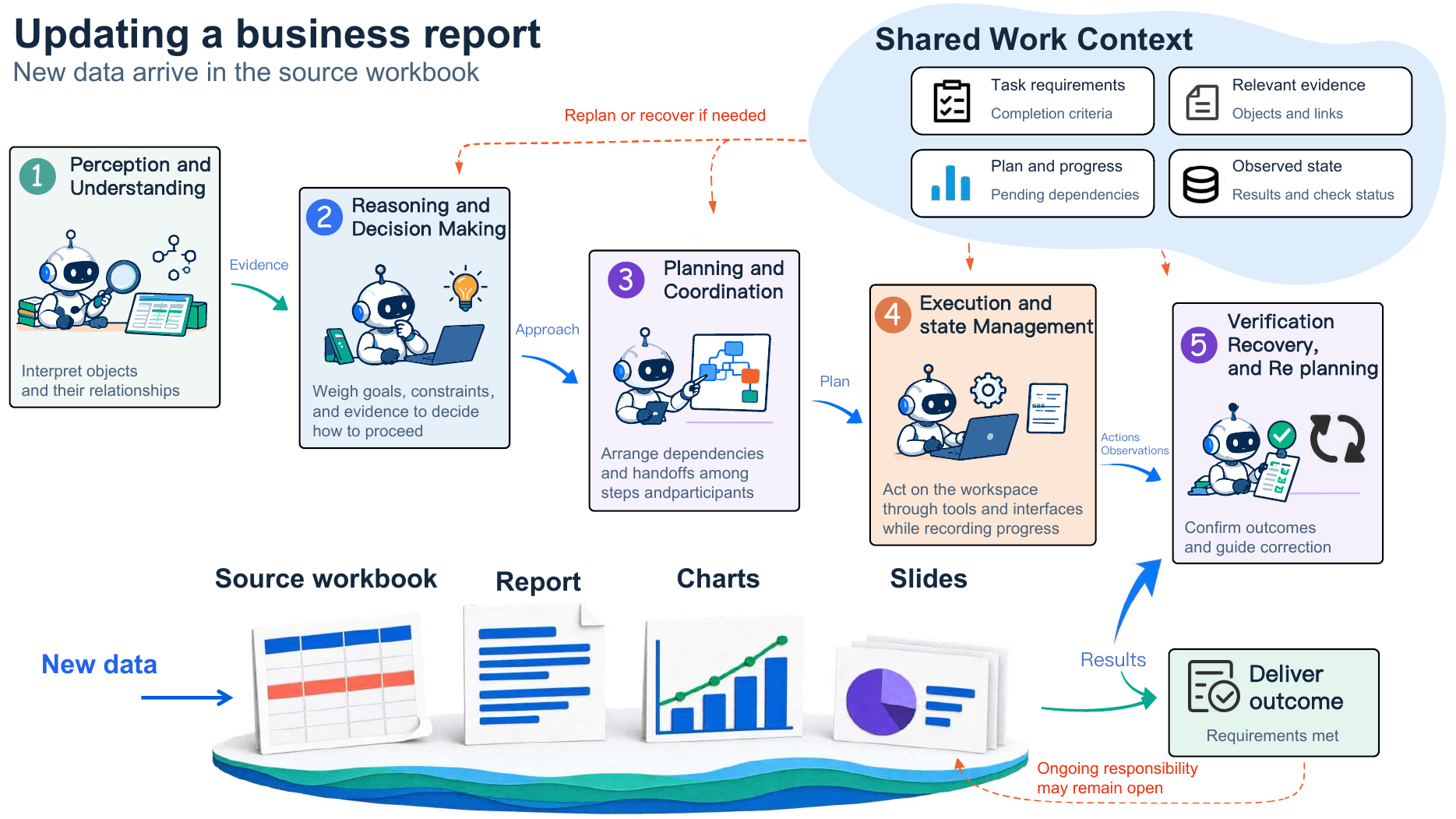}
  \caption{Five core capabilities supporting an illustrative business-report update. New data in the source workbook motivate revisions of the report, charts, and slides. The numbered panels identify the five functions, and solid arrows trace evidence, approaches, plans, actions, observations, and results. Shared work context retains task requirements, evidence and object relationships, plans and pending dependencies, and observed state with check status. Verification supports delivery when current requirements are met. Orange dashed paths indicate possible recovery, replanning, or continued work when an ongoing responsibility remains open after delivery.}
  \label{fig:working-capabilities}
\end{figure}

These capabilities form a cycle rather than a pipeline. The numbered panels in Figure~\ref{fig:working-capabilities} illustrate a pass through this cycle: Perception and Understanding passes evidence to Reasoning and Decision Making, which passes a chosen approach to Planning and Coordination; the resulting plan guides Execution and State Management in taking actions and recording observations. Verification, Recovery, and Replanning checks both the updated work objects and these execution records. When the current delivery requirements are met, the outcome can be delivered; otherwise, the work continues at the capability where the problem lies (Section~\ref{verification-recovery}). An ongoing responsibility remains active after delivery until its agreed closure conditions are met. Throughout, the capabilities read context and record updates through a shared work context, which holds task requirements with their completion criteria, relevant evidence about objects and links, plan and progress with pending dependencies, and observed state with results and check status.

The two dimensions of Section~\ref{sec:foundation-working-conditions} raise the demands on this cycle in different ways: persistence extends each capability across stages and events, whereas openness increases what must be discovered. Table~\ref{tab:conditions-capabilities} lists the resulting demands.

\begin{table}[t]
\centering
\fontsize{7.4}{9.2}\selectfont
\setlength{\tabcolsep}{3pt}
\renewcommand{\arraystretch}{1.15}
\arrayrulecolor{tableRule}
\caption{How work persistence and work openness raise the demands on the five core capabilities.}
\label{tab:conditions-capabilities}
\begin{tabularx}{\linewidth}{>{\raggedright\arraybackslash}m{0.23\linewidth}>{\raggedright\arraybackslash}X>{\raggedright\arraybackslash}X}
\toprule
\rowcolor{tableHeader}
\textcolor{tableInk}{\bfseries Capability} & \textcolor{tableInk}{\bfseries Higher persistence requires} & \textcolor{tableInk}{\bfseries Higher openness requires} \\
\midrule
\MethodName{Perception and Understanding} & Re-perceiving changed state and detecting stale evidence & Discovering relevant objects and relationships during execution \\
\midrule
\MethodName{Reasoning and Decision Making} & Revisiting earlier decisions when evidence or constraints change & Deciding whether to act or to gather more information \\
\midrule
\MethodName{Planning and Coordination} & Spanning waiting periods, handoffs, and events & Identifying dependencies and participants during execution \\
\midrule
\MethodName{Execution and State Management} & Keeping durable state across sessions; avoiding duplicate effects on retry & Establishing applicable interfaces, permissions, and authority during execution \\
\midrule
\MethodName{Verification, Recovery, and Replanning} & Re-verifying affected results; replanning when prerequisites change & Establishing which sources count as evidence \\
\bottomrule
\end{tabularx}
\end{table}

The following subsections examine the five capabilities in turn, using one task throughout: updating a business report, together with its source workbook and associated slides, when new data arrive. Each subsection describes what the capability does and what it passes on, examines representative work, and states what it provides. To show how the two dimensions raise the demands listed in Table~\ref{tab:conditions-capabilities}, each subsection then considers an extended version of the task: the agent is responsible for keeping the report up to date until it is approved, a data revision arrives during assessment in an upstream workbook that feeds the source workbook, and the request names neither that workbook nor the owner of the revised data.

\subsubsection{Perception and Understanding}\label{sec:foundation-perception-understanding}\label{perception-understanding}\label{workspace-understanding}\label{perception}

Before acting, an agent must know what the task refers to. Perception and Understanding interprets objects and relationships: it identifies the objects involved in a task, interprets the content and structure of each object, and traces its links to other objects. It produces evidence whose meaning and source can be traced, including where each target of an edit is located, for use by Reasoning and Decision Making. In the report task, a value in the workbook becomes usable evidence only once its header, unit, and reporting period are known, and an edit can be applied only once its target cells or slide element are located.

Existing work studies both the structure within an object and the links between objects. The structure within an object tells what a value means and where an edit applies. For meaning, SpreadsheetAgent~\cite{ren2026robustrealworldspreadsheetunderstanding} extracts hierarchical headers and layout cues from local spreadsheet regions, and PPTAgent~\cite{pptagent} analyzes the functional types and content structures of reference slides. For location, SpreadsheetLLM~\cite{spreadsheetllm} uses structural anchors, cell-address mappings, and format aggregation to locate relevant spreadsheet regions, and AutoCodeRover~\cite{autocoderover} searches program structure to locate issue-relevant classes and methods. The links between objects tell what else a task involves: RepoBench~\cite{repobench} evaluates the retrieval of relevant code across files, and Workspace-Bench~\cite{workspacebench} requires the agent to connect content spread over several files.

Perception and Understanding thus provides an actionable account of the task's objects and the links between them. In the report task, it links a reported value to its source cells and to the chart or slide that uses it, so that later checks can follow these links. When the revision arrives during assessment, perception must determine which values already cited in the report are affected; to do so, it must locate the upstream workbook that the revision changes.

\subsubsection{Reasoning and Decision Making}\label{sec:foundation-reasoning-decision}\label{reasoning}

Located evidence does not by itself determine what to do next. Reasoning and Decision Making weighs goals, constraints, and evidence to decide how to proceed among the available alternatives. Goals determine the desired result, and constraints limit acceptable actions; goals and some constraints are supplied with the task, while further constraints may need to be established during execution. Evidence establishes the current situation and accumulates as the work proceeds, and domain knowledge, learned by the model or retrieved, determines how it should be interpreted. The result is a chosen approach, which Planning and Coordination arranges into a plan. In the report task, such reasoning determines whether two figures are comparable, which calculation is appropriate, and whether more information is needed before an edit is made.

Existing work studies both domain-guided decisions and the use of evidence gathered during work. For domain-guided decisions, FinQA~\cite{finqa} requires domain knowledge to reason numerically over financial reports, and $\tau$-bench~\cite{taubench} makes domain policies and user requirements constrain tool-use decisions. For evidence gathered during the work, ReAct~\cite{react} interleaves reasoning with actions and observations, SheetAgent~\cite{chen2025sheetagentgeneralistagentspreadsheet} queries relevant spreadsheet subviews for its planner and uses code-execution results to inform subsequent operations, and CodeAct~\cite{codeact} expresses actions as executable code and feeds interpreter outputs and error messages into subsequent decisions.

Reasoning and Decision Making thus provides decisions that can be checked later. To make this possible, the harness records each decision in the shared work context, together with the goal it serves, the constraints that applied, and references to the supporting evidence. Later steps can then tell whether it worked and whether its basis still holds. In the report task, the record of a decision to recompute a growth rate shows which figures were compared and why they were judged comparable. The revision reopens decisions like this one, and the agent must decide whether it can act on the revision directly or must first gather more information.

\subsubsection{Planning and Coordination}\label{sec:foundation-planning-coordination}\label{planning}

When delegated work involves several dependent actions or contributions from other participants, each decision must be connected to the work that follows. Planning and Coordination arranges the dependencies and handoffs among these actions and contributions. Planning specifies subgoals, required inputs, and the order in which results become available~\cite{llmplusp,zeng2025routinestructuralplanningframework,liu2025workteamconstructingworkflowsnatural}; coordination obtains and routes the information, authorization, or artifacts supplied by other participants~\cite{coordination}. The result is a plan, which Execution and State Management carries out. In the report task, the analysis must be completed before the dependent charts are revised, and any information needed from collaborators must be obtained in time.

Representative work emphasizes different aspects of planning and coordination. Planning methods make dependencies explicit and revise them as results arrive: Data Interpreter~\cite{datainterpreter} represents workflows as hierarchical dependency graphs, records task results and status, and adjusts affected tasks as execution develops, and Agent S2~\cite{agents2} updates the remaining subgoals from the original task and new observations after each subgoal concludes. Coordination frameworks make one participant's output available as another's input: AutoGen~\cite{autogen} organizes role-based exchanges among agents, tools, and human input, and MetaGPT~\cite{metagpt} passes structured artifacts such as requirements and designs through a shared message pool and a subscription mechanism.

Planning and Coordination thus provides a plan that records the result each step produces, the inputs needed to produce it, and the steps or participants that will use it, including dependencies that are still pending. In the report task, the record leads from a revised data source directly to the analysis, charts, and slides that must be redone. Because responsibility lasts until approval, the plan must span the assessment period and take in the revision when it arrives, and the owner of the revised data must be identified during execution.

\subsubsection{Execution and State Management}\label{sec:foundation-execution-state}\label{execution}

Plans take effect only when they are applied to the workspace. Execution and State Management takes actions on workspace files, applications, and records through tools and interfaces, receives observations of their state, and records progress. Execution performs operations through code, tool interfaces, or graphical interfaces within the task's authorized scope; state management tracks the objects and versions involved, completed operations, returned results, and pending items. The two are connected because operations can change workspace state or reveal information that affects the next decision. The results are the artifacts produced and the records kept, which Verification, Recovery, and Replanning checks; in the report task, these are the revised workbook and the list of calculations and dependent edits already completed.

To retain the effects and observations on which later decisions depend, existing systems maintain action and state records during execution. SheetCopilot~\cite{sheetcopilot} executes spreadsheet operations through an observe--propose--revise--act state machine and updates its action and state records after each successful execution. InfiAgent~\cite{yu2026infiagentinfinitehorizonframeworkgeneralpurpose} extends this practice beyond a single run: it stores plans, intermediate artifacts, and tool outputs in a task directory and updates progress markers, so that an interrupted task can be resumed from the record.

Execution and State Management thus provides a record that tells which effects can be relied on. The record distinguishes an \textit{attempted action}, an \textit{accepted call}, a \textit{committed effect}, and a \textit{verified effect}: an attempted action has been issued; an accepted call has been acknowledged by the interface; a committed effect is reflected in the target state; and a verified effect has additionally been checked against the relevant requirement. These states need not coincide, so acceptance of a call alone does not establish task success. An operation whose effect is known to be committed becomes the starting point for the next one, whereas an operation with an uncertain effect remains pending until its state is inspected. In the report task, the saved workbook and its current version are the basis for the next edit, whereas a chart update that may not have been applied stays pending until the slide is checked. Because the work resumes when the revision arrives during assessment, the record must be kept across sessions so that resumed work does not repeat edits already made; before reading the upstream workbook, the agent must also establish that it is permitted to access it.

\subsubsection{Verification, Recovery, and Replanning}\label{sec:foundation-verification-recovery-replanning}\label{verification-recovery}

Recorded changes show what has happened but not whether it satisfies the task. Verification, Recovery, and Replanning confirms outcomes and guides correction. Its three functions proceed from assessment to corrective action: verification checks results against task requirements and completion criteria, recovery repairs the objects or operations responsible for a problem, and replanning revises the remaining steps when feedback shows that a prerequisite or condition has changed. We examine representative work for each function in turn.

Verification relates each requirement to evidence that can establish whether it has been met. The appropriate evidence depends on the property being checked, whether a benchmark defines the check or the agent runs it itself. Behavior is checked by execution: SWE-bench~\cite{swebench} runs tests against the modified repository, and SpreadsheetBench~\cite{spreadsheetbench} applies a spreadsheet transformation to several workbook inputs. Claims are checked against external information: CRITIC~\cite{critic} uses external tools to obtain evidence for critiquing and correcting its own outputs. Presentation is checked on the rendered artifact: DeepPresenter~\cite{deeppresenter} inspects rendered slides to identify issues such as overlap and overflow. The report task needs all three: executing its calculations, checking its claims against their sources, and inspecting its rendered slides.

When verification detects a problem, recovery repairs it. Self-Refine~\cite{selfrefine} revises the current output from generated feedback, whereas Reflexion~\cite{reflexion} retains reflections on task feedback to improve the next attempt. In a workspace, a repair must also target the object or operation responsible, such as a wrong formula or an overflowing slide. When the effect of an earlier operation is uncertain, the agent must first inspect the current state to decide whether to retain, repair, or retry it; this matters most for operations, such as sending a message, that produce an additional effect when repeated.

Some problems cannot be repaired locally because a prerequisite or condition has changed. Replanning then revises the remaining course of work: it uses the dependencies recorded in the plan (Section~\ref{planning}) to identify the steps the change affects and draws on Planning and Coordination to rebuild them. LATS~\cite{lats} searches over alternative reasoning and action paths, using environmental feedback and reflection to choose among them.

Verification, Recovery, and Replanning thus provides confirmed outcomes and corrections matched to the scope of each problem: a local defect receives a local repair and recheck, whereas a changed prerequisite leads to a revised plan and a check of its downstream results. After the revision, affected results already checked must be re-verified, and the agent must first establish which source counts as evidence for the revised figures.

With it, the cycle in Figure~\ref{fig:working-capabilities} closes: a confirmed result is delivered or becomes the basis for further work, and a remaining problem is directed to the capability it concerns, with missing evidence going back to Perception and Understanding, a questionable decision to Reasoning and Decision Making, a changed prerequisite to Planning and Coordination, and an execution error to Execution and State Management. Section~\ref{related-paradigms} relates this framework to existing agent paradigms.

\subsection{Work Agents and Related Agent Paradigms}\label{related-paradigms}

Relative to this framework, three established lines of agent research each focus on one facet of the work: long-horizon agents on sustaining execution over many steps, computer-use agents on the action interface, and office agents on the type of artifact produced. A work agent is instead defined by its delegated objective and by outcomes assessed against task requirements, regardless of how long it runs, which interface it uses, or what artifacts it produces. Sections~\ref{sec:foundation-long-horizon} to~\ref{sec:foundation-office} summarize the research focus of each paradigm and state what the work-agent view adds, and Section~\ref{sec:foundation-paradigm-relationships} shows how the perspectives combine on one shared task.

\subsubsection{Long-Horizon Agents}\label{sec:foundation-long-horizon}

The first paradigm concerns sustaining execution over many steps. Long-horizon agents study how to maintain task progress, context, and state over extended execution and how to handle dependencies whose consequences emerge in later stages. A recent synthesis organizes this research into three levels~\cite{longhorizonsynthesis}: the intra-context level covers coupled reasoning steps, the cross-context level covers state continuity beyond a single context window, and the cross-task level covers experience accumulation across tasks. Evaluations quantify this ability: the METR time-horizon analysis~\cite{metrhorizon} measures task length by the time human experts take to complete the task and estimates the length at which an agent reaches a given success rate, while RE-Bench~\cite{rebench} compares research-engineering performance under different time budgets.

To sustain such work, representative methods keep information available beyond a single context window. InfiAgent~\cite{yu2026infiagentinfinitehorizonframeworkgeneralpurpose} constructs its reasoning context from a snapshot of its stored task state and a fixed window of recent actions. On the evaluation side, LongMemEval~\cite{longmemeval} tests memory through timestamped user--assistant histories, including information extraction, multi-session reasoning, temporal reasoning, and updates to user information, and LoCoMo~\cite{locomo} evaluates very long-term conversational memory.

These methods and measures concern continuity of execution and retained information. Work persistence asks a different question: when responsibility may close (Section~\ref{work-persistence}). Persistent delegation requires tracking open obligations, responding to other participants and later events, and re-verifying earlier results against the current workspace.

\subsubsection{Computer-Use Agents}\label{sec:foundation-computer-use}

The second paradigm concerns the interface through which actions are taken. Computer-use agents study how to complete tasks through software interfaces in browsers, desktop applications, and operating systems. A central problem is to ground task intent in operations executable in the current interface: identifying relevant controls, locating action targets, and deciding what to do next from the resulting interface state~\cite{uground,wu2025guiactorcoordinatefreevisualgrounding,omniparser,lu2026uir1,lu2025uis1}. SeeClick~\cite{cheng-etal-2024-seeclick} studies the visual-grounding part of this process by locating screen elements from screenshots and instructions, and uses GUI-grounding pre-training to improve downstream task performance.

Grounding individual operations is not enough; the operations must also add up to the task outcome. Benchmarks in this line, such as WebVoyager~\cite{webvoyager}, therefore evaluate whole sequences of operations. WebArena~\cite{webarena} checks, depending on the task, whether an answer, page location, or website state satisfies the intended goal, and OSWorld~\cite{osworld} uses task-specific scripts to check the resulting files and application states.

The work-agent view places these interface operations within the delegated task as a whole. Computer use is one interface through which the harness acts on the workspace~\cite{oscopilot}. Beyond the interface, this view also asks when responsibility closes and which permissions and dependencies the agent must discover during execution, including dependencies on objects that the interface does not show. Relating interface operations to these questions shows what each operation contributes to the task.

\subsubsection{Office Agents}\label{sec:foundation-office}

The third paradigm concerns the type of artifact produced. Office agents, such as SheetCopilot, PPTAgent, and Huawei Cloud OfficeAce~\cite{sheetcopilot,pptagent,officeace}, create and modify common work artifacts such as documents, spreadsheets, and presentations. Task requirements concern both content and the structure and use of the objects that carry it: spreadsheet operations must act on the correct data and cells, while presentation production must balance content, layout, and overall coherence. The quality of the resulting artifact depends both on how individual objects are manipulated and on how the task is organized.

Representative studies address these two aspects. For object manipulation, SheetCopilot~\cite{sheetcopilot} controls spreadsheets through a set of atomic actions, and PPTAgent~\cite{pptagent} generates slides from reference presentations through code-based editing and evaluates content, design, and coherence. For task organization, OfficeBench~\cite{officebench} provides single- and multi-application tasks involving documents, spreadsheets, calendars, and email; its cross-application tasks require information obtained earlier to be used in later operations, and their results are verified through matching rules or execution-based checks.

Office agents thus take office objects and the workflows around them as their main setting. The work-agent view offers a framework for analyzing such work alongside software, service, and organizational work: it describes office outcomes in the same terms as other work outcomes (Section~\ref{sec:foundation-system-view}), so that an updated workbook, the corresponding changes to business records, and the approval required before publication can be analyzed as parts of one delegated task.

\subsubsection{Relationships among Paradigms}\label{sec:foundation-paradigm-relationships}

Having examined each paradigm separately, we now compare them on the same task. Table~\ref{tab:related-paradigms} uses the report task from Section~\ref{capabilities}, updating a business report when new data arrive; as its column ``What is tracked'' shows, the three related paradigms each track a different part of this task. A single system may need all three, as when an office agent completes a long reporting task through desktop applications and must maintain state over many steps. The work-agent view places these concerns within one work process, from task requirements through execution in the workspace to verified outcomes. This allows the paper to relate research across paradigms, identify which parts of the work each supports, and determine which capabilities must still be combined for a given delegated task.

\begin{table}[!htbp]
\centering
\fontsize{7.4}{9.2}\selectfont
\setlength{\tabcolsep}{3pt}
\renewcommand{\arraystretch}{1.15}
\arrayrulecolor{tableRule}
\caption{Work agents and related agent paradigms, compared through the task of updating a business report when new data arrive. These perspectives are complementary and can be combined within one system.}
\label{tab:related-paradigms}
\begin{tabularx}{\linewidth}{>{\raggedright\arraybackslash}m{0.17\linewidth}>{\raggedright\arraybackslash}X>{\raggedright\arraybackslash}X>{\raggedright\arraybackslash}X}
\toprule
\rowcolor{tableHeader}
\textcolor{tableInk}{\bfseries Perspective} & \textcolor{tableInk}{\bfseries Core question} & \textcolor{tableInk}{\bfseries What is tracked} & \textcolor{tableInk}{\bfseries Same task, different focus} \\
\midrule
\MethodName{Long-Horizon Agents} & How is work sustained over many steps? & Progress, prior results, and pending dependencies & Resume from prior results and incorporate new data. \\
\midrule
\MethodName{Computer-Use Agents} & How are actions grounded in software interfaces? & Targets, interface state, and action effects & Locate cells, edit values, and confirm the saved state. \\
\midrule
\MethodName{Office Agents} & How are office artifacts created and updated? & Content, structure, layout, and cross-artifact links & Update the workbook, report, and associated slides. \\
\midrule
\MethodName{Work Agents} & How is delegated work carried through to a verified outcome? & Task requirements, workspace state, progress, outcomes, and evidence & Plan the update, integrate new data, revise linked artifacts, and verify the agreed deliverables. \\
\bottomrule
\end{tabularx}
\end{table}

The next section examines the scope and organization of the work delegated to work agents.

\FloatBarrier

\section{Work Forms: What Work Do Agents Undertake?}
\label{workforms}

Section~\ref{foundation} describes the system elements, working conditions, and core capabilities of work agents. Together, these provide a foundation for explaining how an agent interprets a task, organizes its actions, and checks their effects. The demands on these capabilities depend on the work delegated to the system. We examine that work through three questions: what outcome must the agent produce, which objects and dependencies fall within its responsibility, and what conditions determine completion? The five work forms below describe how these responsibilities are organized.

\begin{figure}[!htbp]
\centering
\includegraphics[width=\linewidth]{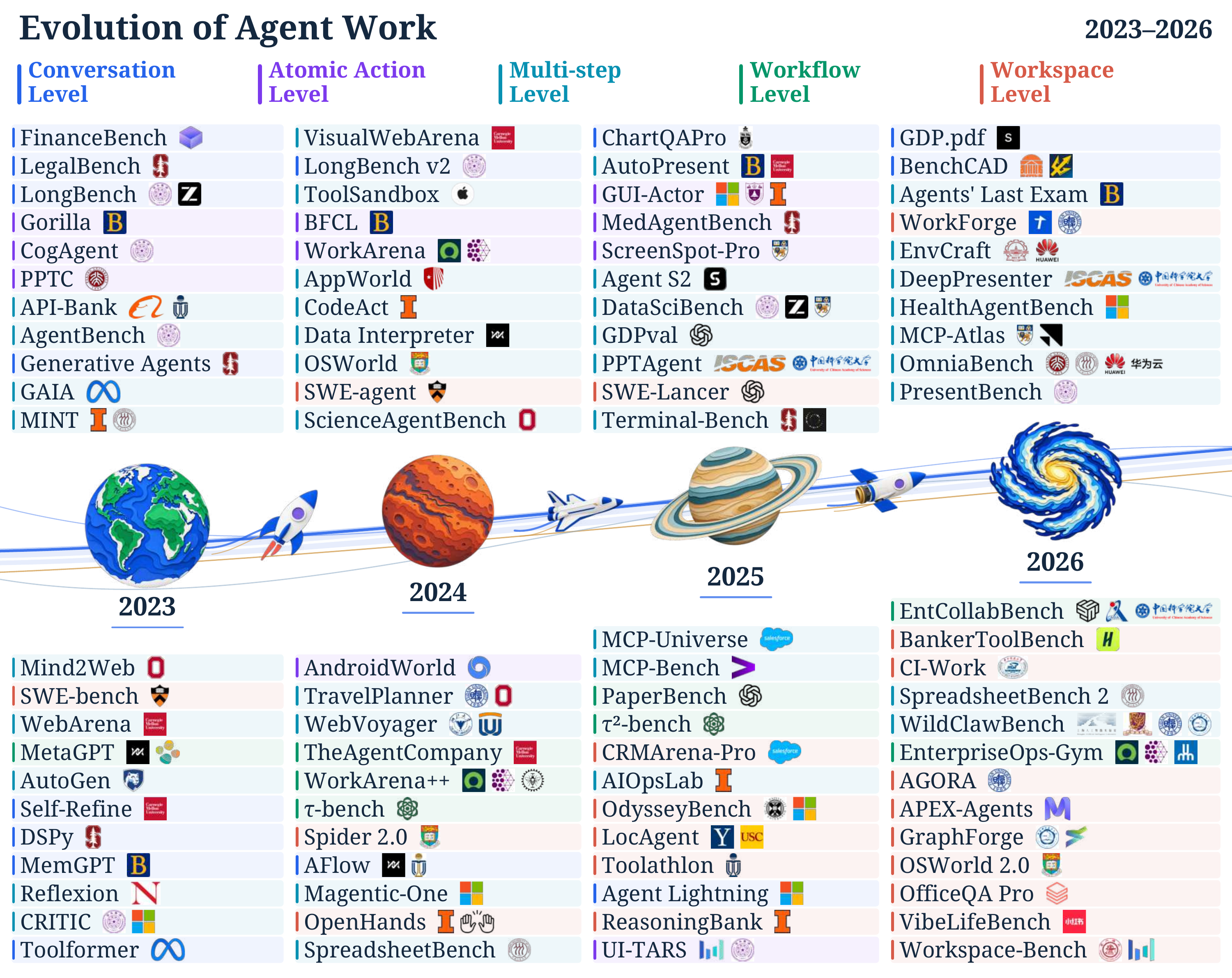}
\caption{Evolution of agent work through 90 representative works in 2023--2026. Colors follow the five work forms in Section~\ref{workforms} and classify selected task settings, not entire methods or frameworks. Method settings include Generative Agents interviews, Self-Refine FED dialogue, MemGPT MSC dialogue, numerical answers in Toolformer, DSPy, AFlow and Agent Lightning, CRITIC web-evidence QA, AutoGen and Reflexion ALFWorld tasks, Magentic-One GAIA tasks, a specified UI-TARS slide edit, ReasoningBank full purchase-history lookup, and an EnvCraft released literature-synthesis task. Benchmark settings include FinanceBench long context, BFCL single calls, WorkArena forms, PPTC individual edits, MedAgentBench blood-pressure documentation, AndroidWorld SMS, two-stage TravelPlanner, interactive AgentBench, feedback-based MINT, MetaGPT procedures, local GUI action or grounding for CogAgent, GUI-Actor and ScreenSpot-Pro, and BenchCAD numeric QA. Workspace examples use SWE-bench repository repair with SWE-agent and OpenHands, SWE-Lancer individual-contributor tasks, Spider 2.0 DBT projects, CRMArena-Pro records, Toolathlon cross-source work, OfficeQA Pro full-corpus search, GraphForge and WorkForge task synthesis, OdysseyBench history-referenced objects, BankerToolBench data-room discovery, CI-Work tool-based retrieval, and LocAgent repository localization. Terminal-Bench uses 2.0. Years refer to selected public versions; AIOpsLab uses its 2025 paper despite 2024 predecessors. Logos identify recorded first-author or joint-first-author institutions.}
\label{fig:workform-benchmark-map}
\label{fig:work-forms-evolution}
\end{figure}

Because a workflow can contain informational responses and local actions, Figure~\ref{fig:workform-benchmark-map} assigns each representative task setting to the broadest form it requires.

At the Conversation Level, the agent produces an informational result, whereas at the Atomic Action Level, it executes a local operation on a work object. Moving to the Multi-step Level, the agent combines such dependent actions to achieve a complete objective. This responsibility broadens at the Workflow Level, where the process also becomes part of the delegated work and the agent advances it through stages governed by procedural requirements and interactions with other participants. It extends further at the Workspace Level, where the request no longer specifies every object involved and the agent must establish which files, records, and applications bear on the work and how they are related. As the scope expands, the agent must manage more of the dependencies surrounding an outcome, from intermediate results to stage transitions and relationships among work objects. Table~\ref{tab:work-forms} summarizes the resulting work requirements.

\subsection{Conversation Level}
\label{conversation-level}

Work at the Conversation Level produces an informational result from a request and its associated evidence, such as an answer, analysis, classification, or judgment. The agent interprets the relevant material and applies the knowledge or computation needed to produce that result. Completion is assessed on the returned information and does not require modifying the source artifacts, external work objects, or their states.

\subsubsection{Producing Informational Outcomes}

In these tasks, the delegated deliverable is information that answers the requester's question or supports a judgment. It may range from a short factual answer to an assessment, explanation, or detailed proposal. SQuAD asks for an answer extracted from a supplied passage~\cite{rajpurkar2016squad100000questionsmachine}, FinQA requires a numerical answer derived from financial text and tables~\cite{finqa}, and MMTU includes classifications and judgments about whether a statement is supported~\cite{mmtu}. Professional requests extend these demands: AAAR-1.0 asks for equation checks and critiques of research papers~\cite{aaar_1}; FinLFQA requests financial analyses grounded in sources and calculations~\cite{finlfqa}; PRBench asks for justified legal and financial advice~\cite{prbench}; and MMLongBench includes summarization of supplied PDF documents~\cite{mmlongbench}. Responses can also take the form of plans or executable specifications. In Idea2Plan's direct-prompting setting, models develop a supplied research idea into a plan that includes experimental design, while TravelPlanner's sole-planning setting asks for a feasible itinerary based on supplied travel information~\cite{idea2plan,travelplanner}. WorfBench asks for a workflow graph of proposed subtasks and dependencies~\cite{worfbench}. Program outputs specify how an object can be constructed: BlenderLLM's CADBench asks for Blender scripts from modeling instructions, while the CADBench reconstruction benchmark asks for editable CAD programs from supplied images or meshes~\cite{du2024blenderllmtraininglargelanguage,doris2026cadbench}.

\subsubsection{Interpreting Evidence and Deriving Responses}

The complexity of a response task depends on both the source material to be understood and the reasoning needed to process it. Prose presents facts in linguistic context, while tables relate values through rows and columns~\cite{tapas,tablellama}. Document images carry information in page layout~\cite{mineru}, while charts encode values and relationships visually~\cite{chartqa}. Extracting a value or comparing entries therefore depends on interpreting these structures~\cite{docvqa,chartqapro,finqa}. In GDP.pdf, a footnote changes which employers belong in a chronological ranking~\cite{gdppdf}. Repository code and structured data also appear as supplied evidence in LongBench v2's multiple-choice comprehension tasks~\cite{longbench_v2}.

The scope of the supplied material determines where relevant evidence must be found, while relationships among its parts determine what must be connected. FinanceBench's long-context setting supplies the relevant filing in the prompt~\cite{financebench}; multi-hop and long-document tasks require linking supporting information across passages, pages, or documents~\cite{yang2018hotpotqadatasetdiverseexplainable,longbench,xl_docbench,wfwa0433,wfwa0419}. A synthesis must further combine these contributions into an account that preserves their meaning~\cite{mmlongbench,finlfqa}. ConvFinQA extends the dependency to conversation history, with follow-up calculations using values or results established in earlier turns~\cite{convfinqa}.

The requested inference determines how the evidence is used. A numerical answer may require computing with selected values, sometimes through an intermediate program~\cite{finqa,pal,pot}, while a legal conclusion may require applying an explicit rule to the facts~\cite{legalbench}. Scientific assessments draw on domain knowledge to interpret the evidence~\cite{jin2019pubmedqadatasetbiomedicalresearch,lu2022learnexplainmultimodalreasoning}. Plans require comparing available options against stated requirements and commonsense feasibility constraints, including whether the proposed choices can be combined into a workable arrangement~\cite{aaar_1,travelplanner}.

An informational response can explain why a report contains an incorrect value. Correcting that value extends the responsibility to changing the relevant workbook entry and checking that the update has taken effect.

\subsection{Atomic Action Level}
\label{atomic-action-level}

This transition from generating information to modifying state defines the Atomic Action Level, which assigns the agent a local objective with a clearly defined target and effect, such as correcting a spreadsheet cell, repositioning a slide element, or updating a record field. Here, the level is determined by the local scope of the work objective rather than the number of low-level execution steps: a single local action may still require several mouse, keyboard, or API operations to complete. Regardless of the interface mechanics, the agent must apply the change to the intended object while preserving any properties that the request leaves unchanged~\cite{delegate52}.

\subsubsection{Creating and Modifying Local Work Objects}

A request may add a bounded element to an existing artifact, such as a formula~\cite{wfwa0370,wfwa0454}, a conditional formatting rule~\cite{wfwa0425}, or a chart in WTM-Bench's single-artifact tasks~\cite{wtm_bench}. String transformations and spreadsheet data manipulation can likewise be specified through examples or natural language~\cite{wfwa0380,wfwa0381}. Existing content can also be edited: InstructExcel expresses spreadsheet operations as scripts, PPTC includes individual edits to slide text, styles, and positions, and MiLDEdit specifies design-layer changes that preserve surrounding layout and non-target content~\cite{instructexcel,pptc,mildedit}.

The same local scope applies to application records and communication state. WorkArena's form tasks request records with specified field values, while WorkBench includes scheduling a particular event and replying to or deleting a selected email~\cite{workarena,workbench}. Each request supplies a target and an intended change that execution must realize.

\subsubsection{Executing and Checking a Local Change}

To carry out these requests, the agent must identify the target and specify the operation through an available interface. ScreenSpot-Pro isolates target location in professional application screenshots~\cite{screenspotpro}. BFCL's single-call tasks examine function and argument specifications; MedAgentBench adds patient-specific parameters through a blood-pressure documentation request linked to the intended record~\cite{patil2025bfcl,medagentbench}.

After execution, completion depends on the resulting state: an artifact must contain the requested content or property, and an application record or message must reflect the intended change. In AndroidWorld's messaging task, the evaluator checks the destination number and message content in the phone's SMS database~\cite{androidworld}. Inspecting the resulting state during execution can likewise tell the agent whether the operation needs correction.

\subsection{Multi-step Level}
\label{multi-step-level}

Many work objectives cannot be met by a single local change: they depend on intermediate results that are not available at the start. Preparing a forecast, for example, requires importing data before constructing calculations and checking the resulting workbook. Intermediate operations establish information or state on which later decisions depend. Work at the Multi-step Level coordinates these dependent operations toward a complete objective, adapting subsequent actions to intermediate results and execution feedback~\cite{webshop}. Because only the objective is delegated, completion is assessed on the final result, whichever sequence of actions reaches it.

\subsubsection{Completing Objectives through Intermediate Results}

A complete objective may culminate in a usable artifact, an informational result assembled from acquired evidence, or a change to an operational environment. Intermediate results supply the components and inputs needed to complete that objective. In data-science tasks, data preparation and code execution lead to an executable analysis, as illustrated by DataSciBench and ScienceAgentBench~\cite{datascibench,scienceagentbench}. Artifact production also requires the parts to work together. Slide generation connects content selection and visual planning~\cite{wfwa0378,autopresent}; each slide must fit the deck's outline and surrounding narrative~\cite{pptagent,presentbench}. In a financial workbook, a value computed on one worksheet may supply calculations on another~\cite{zhu2026spreadsheetbench2evaluatingagents}. Completion therefore depends on both the individual transformations and the relationships among their results.

Intermediate results can also supply the evidence for an answer or plan. TravelPlanner's two-stage setting requires collecting travel options through tools before assembling an itinerary that satisfies the user's constraints~\cite{travelplanner}. GAIA includes questions that require locating a study on the web before recovering the requested fact~\cite{gaia}.

For tasks that change an environment, intermediate changes establish the state needed to complete the requested operation. Terminal-Bench~2.0 includes configuring systems and migrating databases~\cite{terminalbench}. In AIOpsLab's mitigation tasks, the objective is to restore a running service through interventions such as configuration changes or version rollbacks~\cite{aiopslab}.

Desktop and web tasks carry these intermediate results across applications. In OSWorld, payment records determine the recipients of a reminder email; WebArena's museum-itinerary example combines reference pages and a map before recording the route in a project README~\cite{osworld,webarena}.

\subsubsection{Managing Action Dependencies and Feedback}

These dependencies shape the order and content of subsequent actions. A tool result can supply arguments for a later call, as in MCP-Bench and MCP-Universe~\cite{wang2025mcpbenchbenchmarkingtoolusingllm,wfwa0513}. An interpretation of business context can determine which data to join or aggregate before a result is computed, as in DABstep~\cite{dabstep}. A change of state can enable a previously unavailable operation: ToolSandbox's messaging example requires cellular service to be enabled before the message can be sent~\cite{toolsandbox}. TimeArena's simulations add duration and resource constraints: agents must schedule around unfinished actions and occupied objects while advancing independent work~\cite{timearena}.

Feedback from execution can require revising the sequence. Execution feedback guides SheetAgent's workbook editing loop, while Spreadsheet-RL provides recalculated values during execution in Microsoft Excel~\cite{chen2025sheetagentgeneralistagentspreadsheet,spreadsheetrl}. In DeepPresenter, rendered slides guide revisions; VIGA likewise executes candidate programs and inspects the rendered scenes before continuing scene edits in BlenderBench~\cite{deeppresenter,yin2026visionasinversegraphics}. The agent must retain these observations and changes so later actions use the current state.

In many organizational settings, completing a task is only one step within a broader operational process. A finished financial forecast, for example, may need to pass through compliance verification, managerial assessment, and budget allocation before taking effect, where advancing the work depends on procedural rules and interactions with other participants.

\subsection{Workflow Level}
\label{workflow-level}

The Workflow Level widens what is delegated from a single objective to the operational process around it. The process carries the objective through stages with distinct responsibilities and conditions for advancement, and each stage establishes an output, decision, or confirmation on which another stage or participant relies. The agent therefore coordinates the handoffs between stages and the procedural requirements that govern them, including cases where intermediate findings change the remaining work. Because these stages belong to the delegated work, completion covers the process outcome as well as the final output (Section~\ref{sec:foundation-system-view}): a correct forecast still cannot take effect while a required approval is missing. CXBench scores final outcomes and process compliance separately in regulated customer-service tasks, where a run that reaches the correct final state but fails a process check, such as providing a required disclosure or obtaining consent, counts as a false pass~\cite{cxbench}.

The same criterion separates workflow stages from the stages an agent system introduces into its own execution. Anthropic's Research system, for example, runs a research stage, in which a lead agent decides how many search subagents to create and adds more when the evidence is insufficient, followed by a citation stage that attaches sources to the final report~\cite{anthropicresearchsystem}. The requester, however, asks for a sourced answer to a research question, and even an explicit request for multi-agent execution specifies how the system should work without defining any stage on which the result depends. Any other arrangement of stages that yielded an equally well-sourced report would complete the request just as well, so the research and citation stages organize the execution of a multi-step objective rather than forming part of the delegated work.

\subsubsection{Carrying Work through Functional Stages}

Stages group actions around an output or decision that subsequent work requires. A handoff can make an artifact available under an agreed specification, confirm an input, or authorize further work. In WorkArena++, employee-onboarding tasks connect user creation with ordering and assigning hardware~\cite{workarenapp}. A similar handoff occurs in an OfficeBench training task: the agent reads an Excel participant list, creates calendar events, and emails the training details to participants~\cite{officebench}.

The same dependence on stage outputs appears in artifact production. MetaGPT follows the standard operating procedures of a software team: a requirements document guides architecture and interface design, then implementation and testing~\cite{metagpt}. SlideRL uses research findings and an outline to guide slide generation and revision~\cite{sliderl}. EXP-Bench links experimental design and implementation to execution and conclusions, while PaperBench requires implementing and executing a paper's experiments to reproduce its empirical results~\cite{expbench,paperbench}. DAComp makes the handoffs explicit in pipeline tasks that connect cleaning, entity integration, and aggregation under specified data contracts~\cite{dacomp}.

\subsubsection{Coordinating Workflow Progress}

Once a stage has produced its output, assessment, confirmation, and policy conditions determine when work can advance. In CHI-Bench's healthcare administrative tasks, assessment can yield a decision or a request for missing information~\cite{chibench}; $\tau$-bench requires user confirmation before specified database updates~\cite{taubench}. In EnterpriseOps-Gym, warranty status determines whether the agent records a customer interaction and schedules a warranty discussion~\cite{enterpriseops}. AutomationBench's meeting-conflict task applies a spreadsheet priority policy, marks the lower-priority meeting for rescheduling, and reports the decision to Slack~\cite{automationbench}.

Progress also depends on other participants. In $\tau^2$-bench's telecommunications tasks, the agent coordinates its actions with changes made by a tool-using simulated user~\cite{tau2bench}. In EntCollabBench, roles with different tools and data access pass intermediate outputs to one another, while TheAgentCompany includes tasks requiring information from simulated coworkers~\cite{entcollabbench,agentcompany}. In Collaborative Gym, the agent's literature search and the human's feedback feed revisions to a shared related-work draft~\cite{cogym}. Retaining pending changes and their prerequisites lets the agent integrate each reply into the current workflow. These dependencies can extend across repeated operations: E-Commerce Bench connects procurement, fulfillment, returns, and cash-flow decisions over a simulated year~\cite{ecommercebench}.

\subsection{Workspace Level}
\label{workspace-level}

In the preceding forms, the request or task setup largely specifies which objects the work involves. As agents become able to work across entire code repositories and an organization's documents, emails, and chats~\cite{codexdocs,claudedocs,m365docs}, a request can state only the goal. The agent must then establish which objects are relevant and how their relationships affect completion. A report may depend on a particular workbook version, while an approved decision may need to be reflected in several records. The Workspace Level makes discovering and following these relationships part of the delegated responsibility. The agent locates relevant objects, reconciles their evidence and versions, and, when the task requires updates, preserves consistency among the affected objects.

\subsubsection{Locating Relevant Objects in a Workspace}

Relevant material may be distributed across documents and spreadsheets, message histories, code repositories, or linked business records. Workspace-Bench makes identifying relevant objects and their dependencies part of the task~\cite{workspacebench}. GraphForge constructs assignments requiring comparison, reconciliation, or derivation across files, while WorkGenesis requires drawing on public source files and synthesized companion attachments to produce occupational deliverables~\cite{graphforge,workgenesis}. OfficeQA Pro's full-corpus tasks require finding the pages needed for an answer, while DRBench's application-based tasks require retrieving enterprise material for a research report~\cite{officeqapro,drbench}. Finding the right page can itself require reading charts and layouts, as in ViDoRe, introduced with ColPali~\cite{colpali}. As a retrieved object points to further sources, the relevant context develops through these connections.

\subsubsection{Following Dependencies across Workspace Objects}

The connections uncovered during search determine how retrieved objects fit into the task. They may be explicit references, shared entities, or relationships of derivation and version. In CRMArena, identifying which order a customer refers to requires relating product information to that customer's purchase history~\cite{crmarena}. LOCA-bench makes these cross-source relationships explicit in a task adapted from Toolathlon: assembling an exam spreadsheet requires matching Canvas announcements and emails to the correct courses~\cite{toolathlon,zeng2026loca}. In Loc-Bench, repository tasks similarly require locating the files and functions relevant to an issue description~\cite{locagent}. Repository repair extends this process to modifying code and testing the affected components~\cite{swebench,sweagent,openhands,swelancer}. Spider~2.0's DBT tasks extend this dependence to data-transformation projects: the agent reads SQL, YAML, and Markdown files to complete a missing transformation and check the resulting database state~\cite{spider2}.

Using related objects together can also require reconciling what they say. Agora tasks require deriving a numerical answer from documents whose terminology, units, and time periods must be reconciled, while HERB queries connect people, products, and release events across heterogeneous enterprise records~\cite{agora,herb}. APEX-Agents makes source selection concrete in a valuation task that combines the implied share price in a discounted cash-flow model with earnings per share from a financial statement for a specified date~\cite{apex}. Tracing historical versions and derived-content relationships helps establish which sources support the conclusion~\cite{workspacebench}.

Object relationships also determine what must be rechecked when work changes. In OSWorld~2.0, revised budget and vendor constraints arriving through TeamChat require the agent to reconsider affected entries in a purchase-order form~\cite{yuan2026osworld2}. Linking the updated constraints in the chat to the affected form entries identifies which earlier work needs revision. Discovering these dependencies exercises work openness; maintaining the result as relevant inputs change exercises persistence (Section~\ref{sec:foundation-working-conditions}).

\subsection{Synthesis: Work Scope and Capability Demands}
\label{workforms-synthesis}

Across the five work forms, the foundational capabilities introduced in Section~\ref{foundation} support work over an expanding set of dependencies. As an assignment grows, an interpretation of the evidence can affect decisions and changes across several stages or objects. The agent must carry those dependencies through the work and use feedback from the resulting state to decide what can proceed and what needs revision. Work complexity thus scales through the web of dependencies that these capabilities must manage in concert.

Table~\ref{tab:working-agents-4} relates the five work forms to the capability requirements introduced in Section~\ref{capabilities}. Work at the Workspace Level requires tracking the sources and dependencies behind a result, with consistency checks extending to affected objects when updates are required.

\begingroup
\fontsize{7.4}{9.2}\selectfont
\setlength{\tabcolsep}{2.5pt}
\renewcommand{\arraystretch}{1.12}
\arrayrulecolor{tableRule}
\Needspace{150pt}
\begin{longtable}{@{}m{0.155\linewidth}m{0.153\linewidth}m{0.153\linewidth}m{0.153\linewidth}m{0.153\linewidth}m{0.153\linewidth}@{}}
\caption{Capability demands across the five Work Forms\label{tab:working-agents-4}\label{tab:work-forms}}\\
\toprule
\rowcolor{tableHeader}
\textcolor{tableInk}{\bfseries Work Form} & \textcolor{tableInk}{\bfseries Perception and Understanding} & \textcolor{tableInk}{\bfseries Reasoning and Decision Making} & \textcolor{tableInk}{\bfseries Planning and Coordination} & \textcolor{tableInk}{\bfseries Execution and State Management} & \textcolor{tableInk}{\bfseries Verification, Recovery, and Replanning} \\
\midrule
\endfirsthead
\multicolumn{6}{@{}l@{}}{\small\tablename~\thetable~(continued)}\\
\toprule
\rowcolor{tableHeader}
\textcolor{tableInk}{\bfseries Work Form} & \textcolor{tableInk}{\bfseries Perception and Understanding} & \textcolor{tableInk}{\bfseries Reasoning and Decision Making} & \textcolor{tableInk}{\bfseries Planning and Coordination} & \textcolor{tableInk}{\bfseries Execution and State Management} & \textcolor{tableInk}{\bfseries Verification, Recovery, and Replanning} \\
\midrule
\endhead
\midrule
\multicolumn{6}{r@{}}{\color{tableMuted}\scriptsize Continued on next page}\\
\endfoot
\bottomrule
\endlastfoot
\MethodName{Conversation Level} & Interpret supplied evidence and context & Derive supported answers and judgments & Organize local reasoning when needed & Produce the requested response or specification & Check evidence support, constraints, and output validity \\
\MethodName{Atomic Action Level} & Identify the target object and local state & Select an appropriate operation & Establish a valid local action path & Commit the intended local change & Confirm the target effect \\
\MethodName{Multi-step Level} & Track intermediate evidence and state & Adapt decisions to intermediate results & Decompose a coherent objective & Maintain state across related actions & Check intermediate and final outcomes \\
\MethodName{Workflow Level} & Identify stages, rules, and shared state & Reason about prerequisites and policies & Coordinate stages, branches, and handoffs & Commit consequential state transitions & Validate stage transitions and workflow outcomes \\
\MethodName{Workspace Level} & Discover relevant objects, relations, versions, and authority & Reconcile evidence across objects & Plan over object and dependency structure & Track object state; preserve consistency when updating related objects & Check source lineage, freshness, and dependencies \\
\end{longtable}
\endgroup

This dependency-centered view also clarifies where a work agent needs improvement: missing evidence can interrupt a sequence of actions, an incorrect decision can propagate into later stages, and a changed constraint can invalidate dependent outputs. Section~\ref{optimization} examines how experience from such work can improve the model, the harness, and their interaction, while Section~\ref{evaluation} considers how to assess the resulting work and the progress made by these improvements.

\section{Optimization: From Work Experience to System Capability}
\label{optimization}

\DeclareTableResource{paper}{compoworld}{https://arxiv.org/abs/2609.33665v2}{\PaperResourceIcon}{Paper}{resourcePaper}
\DeclareTableResource{paper}{workborne}{https://arxiv.org/abs/2610.14915}{\PaperResourceIcon}{Paper}{resourcePaper}
\DeclareTableResource{paper}{coworkenv}{https://arxiv.org/abs/2610.24960}{\PaperResourceIcon}{Paper}{resourcePaper}
\DeclareTableResource{code}{coworkenv}{https://github.com/HKUST-KnowComp/CoworkEnv}{\CodeResourceIcon}{Code}{resourceCode}
\DeclareTableResource{paper}{workforge}{https://arxiv.org/abs/2610.04906}{\PaperResourceIcon}{Paper}{resourcePaper}
\DeclareTableResource{paper}{workrover}{https://arxiv.org/abs/2610.18067}{\PaperResourceIcon}{Paper}{resourcePaper}
\DeclareTableResource{paper}{coworkgym}{https://arxiv.org/abs/2610.17670}{\PaperResourceIcon}{Paper}{resourcePaper}
\DeclareTableResource{paper}{mercorskyrl}{https://www.mercor.com/blog/training-frontier-knowledge-work-agents-a-397b-rl-training-guide-with-skyrl/}{\PaperResourceIcon}{Report}{resourcePaper}
\DeclareTableResource{code}{mercorskyrl}{https://github.com/Mercor-Intelligence/ApexAgents-SkyRL-Recipe}{\CodeResourceIcon}{Code}{resourceCode}
\DeclareTableResource{paper}{gigpo}{https://arxiv.org/abs/2505.10978}{\PaperResourceIcon}{Paper}{resourcePaper}
\DeclareTableResource{paper}{verse}{https://arxiv.org/abs/2610.02616}{\PaperResourceIcon}{Paper}{resourcePaper}
\DeclareTableResource{code}{verse}{https://github.com/wzekai/VERSE}{\CodeResourceIcon}{Code}{resourceCode}

\begin{figure}[!t]
  \centering
  \includegraphics[width=\linewidth]{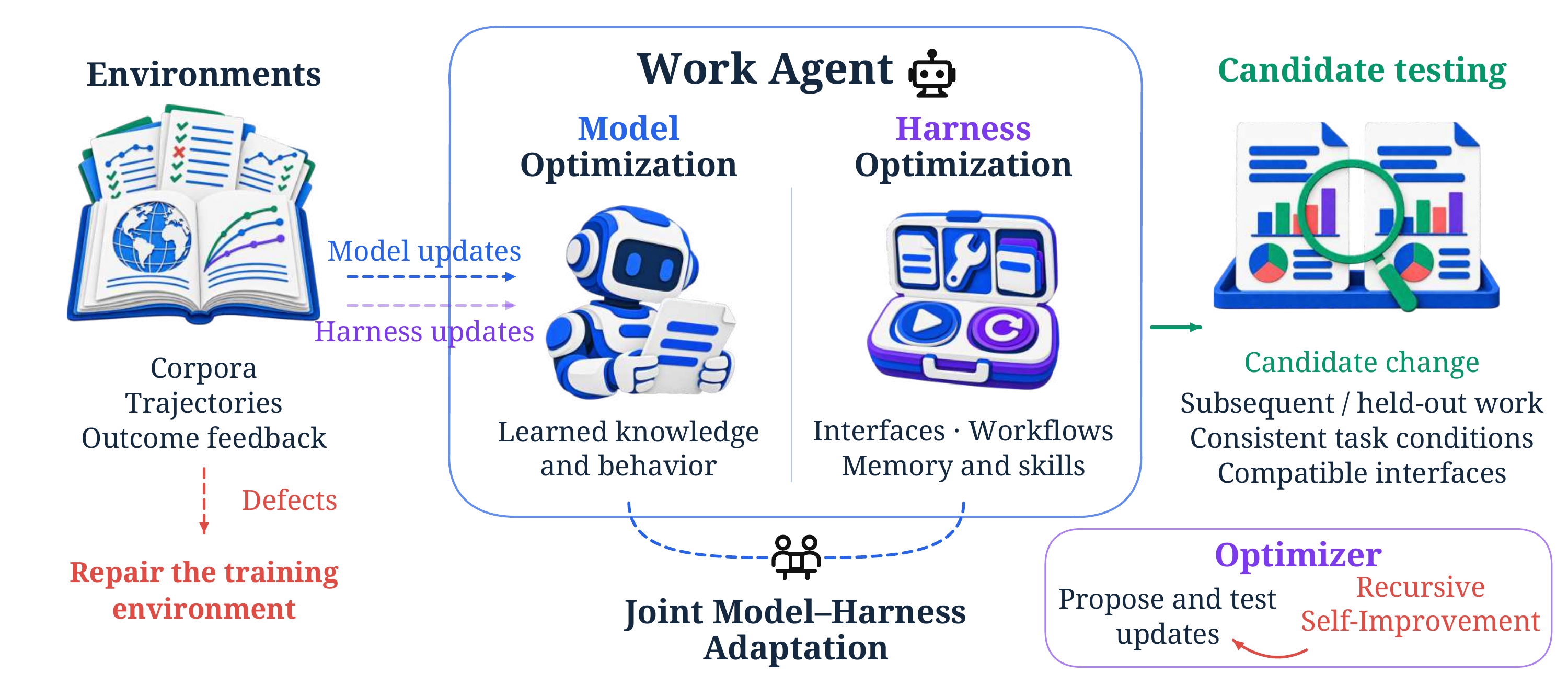}
  \captionof{figure}{Optimization targets in a work-agent system. Training environments, corpora, trajectories, and outcome feedback inform candidate changes to the model, harness, or their coordination. Model optimization changes learned knowledge and behavior; harness optimization changes interfaces, workflows, memory, and skills. Joint adaptation coordinates model and harness changes. Defects in the training environment may require environment repair. Comparisons on subsequent work under consistent task conditions and compatible interfaces test whether a candidate update helps. The optimizer proposes and tests candidate updates; optimizer-level recursive self-improvement also changes how updates are proposed and tested.}
  \label{fig:overview-4}
\end{figure}

The capabilities required by work agents must be cultivated via systematic optimization. As discussed in Section~\ref{foundation}, this optimization can take place at both the model and harness levels. Model updates improve interpretation, reasoning and action selection, while harness updates enhance the interfaces, tools, memory and execution workflows that support the model. These improvements draw on diverse corpora and experience gathered during work, including observations of the environment, action trajectories and feedback on task outcomes. As illustrated in Figure~\ref{fig:overview-4}, Section~\ref{experience} examines the environments, data, and feedback that provide learning signals. Section~\ref{model-optimization} examines how pre-training, mid-training, and post-training develop model capabilities, while Section~\ref{harness-optimization} explores improvements to interfaces, workflows, and memory. Finally, Section~\ref{joint-optimization} examines how models and harnesses can adapt together and support continued self-improvement.



\subsection{Training Environments, Data, and Rewards}\label{experience}
Agents learn to perform work through environments, data and feedback. Environments provide tasks for interaction, while corpora offer relevant knowledge and trajectories show how actions affect task progress. Feedback helps agents identify successful decisions and outcomes. For example, when preparing a report from multiple workbooks, an agent needs to understand how source values relate to calculations and reported findings, and how to update affected results when the source data changes. Table~\ref{tab:experience-construction} compares methods for constructing environments, curating corpora and trajectories, and designing feedback.


\begingroup
\fontsize{7.6}{9.4}\selectfont
\setlength{\tabcolsep}{3pt}
\renewcommand{\arraystretch}{1.12}
\arrayrulecolor{tableRule}
\setlength{\LTpre}{7pt}
\setlength{\LTpost}{10pt}
\begin{longtable}{@{}>{\raggedright\arraybackslash}m{0.22\linewidth}>{\raggedright\arraybackslash}m{0.39\linewidth}>{\raggedright\arraybackslash}m{0.36\linewidth}@{}}
\caption{Constructing experience for work-agent optimization. Representative methods for constructing work environments, curating corpora and trajectories, and designing feedback. Rows describe the mechanisms used to produce training inputs and learning signals.}\label{tab:experience-construction}\\
\toprule
\rowcolor{tableHeader}
\textcolor{tableInk}{\bfseries Representative work} & \textcolor{tableInk}{\bfseries Main mechanism} & \textcolor{tableInk}{\bfseries Training inputs or feedback} \\
\midrule
\endfirsthead
\multicolumn{3}{@{}l}{\small\textit{Table \thetable\ continued.}}\\
\toprule
\rowcolor{tableHeader}
\textcolor{tableInk}{\bfseries Representative work} & \textcolor{tableInk}{\bfseries Main mechanism} & \textcolor{tableInk}{\bfseries Training inputs or feedback} \\
\midrule
\endhead
\midrule
\multicolumn{3}{r@{}}{\footnotesize\textit{Continued on the next page}}\\
\endfoot
\bottomrule
\endlastfoot

\rowcolor{tableHeader}
\multicolumn{3}{@{}l@{}}{\rule{0pt}{12pt}\textcolor{tableInk}{\bfseries Work Environment Construction}}\\*
\MethodEntry{daVinci-Env}{davincienv}\MethodEntry{Spreadsheet-RL}{spreadsheetrl} & Reconstruct tasks from repository changes or paired workbooks, with executable operations and result checks. & Initial workspaces, requested changes, and checkable outcomes. \\
\addlinespace[3pt]
\MethodEntry{NexForge}{nexforge}\MethodEntry{WorkGenesis}{workgenesis} & Use capability requirements and occupational activities to guide task and resource collection. & Professional requests with supporting work materials. \\
\addlinespace[3pt]
\MethodEntry{WorkForge}{workforge}\MethodEntry{WorkBorne}{workborne} & Expert workflows (WorkForge) and artifact transitions (WorkBorne) guide task construction from real files. & Source materials, dependent operations, and intended outputs. \\
\addlinespace[3pt]
\MethodEntry{Agent World Model}{agentworldmodel}\MethodEntry{AgentMercury}{agentmercury} & Connect databases, business entities, and tool interfaces across applications. & Executable tasks with related records and changing state. \\
\addlinespace[3pt]
\MethodEntry{EnvCraft}{envcraft}\MethodEntry{CompoWorld}{compoworld} & Build dependent task chains (EnvCraft) or compose reusable services through task graphs (CompoWorld). & Workflows that require information and results from earlier operations. \\
\addlinespace[3pt]
\MethodEntry{CoworkEnv}{coworkenv}\MethodEntry{Cowork-Gym}{coworkgym} & Validate requests, records, tools, and checks together (CoworkEnv); include a stateful user (Cowork-Gym). & Training interactions with workflow constraints, user input, and workspace changes. \\
\addlinespace[3pt]
\MethodEntry{WorkRover}{workrover} & Diagnose incompatible environment components and replay candidate repairs under fixed requirements. & Repaired environments that support further execution and data collection. \\
\addlinespace[3pt]

\rowcolor{tableHeader}
\multicolumn{3}{@{}l@{}}{\rule{0pt}{12pt}\textcolor{tableInk}{\bfseries Corpus and Trajectory Curation}}\\*
\MethodEntry{daVinci-Dev}{davincidev} & Link issues, repository context, and edits; combine reconstructed histories with real tool executions. & Context-rich work records and action--observation sequences, including failed attempts. \\
\addlinespace[3pt]
\MethodEntry{AgentTrek}{agenttrek} & Filter web tutorials, replay their instructions in real environments, and verify the resulting trajectories. & Demonstrations linking task instructions to observations and executed actions. \\
\addlinespace[3pt]
\MethodEntry{EnterpriseLab}{enterpriselab} & Derive tool dependencies from schemas and pass valid returned values into later calls. & Executable tool sequences and task descriptions that preserve data dependencies. \\
\addlinespace[3pt]
\MethodEntry{OpenThoughts-Agent}{openthoughtsagent} & Compare task sources, teachers, mixtures, and filters through downstream training. & Selected trajectory mixtures evaluated for transfer across agent tasks. \\
\addlinespace[3pt]
\MethodEntry{CurateEvo}{curateevo} & Use development-set failures to revise augmentation, filtering, and refinement. & Updated SFT and RL data and inference-time memory from a fixed raw corpus. \\
\addlinespace[3pt]

\rowcolor{tableHeader}
\multicolumn{3}{@{}l@{}}{\rule{0pt}{12pt}\textcolor{tableInk}{\bfseries Feedback Design}}\\*
\MethodEntry{Spreadsheet-RL}{spreadsheetrl}\MethodEntry{Anchor}{anchor} & Recalculate workbook outputs (Spreadsheet-RL) or check business constraints and solution quality (Anchor). & Outcome signals from the resulting workspace state. \\*
\addlinespace[3pt]
\MethodEntry{GraphForge}{graphforge}\MethodEntry{SlideRL}{sliderl} & Link rubrics to sources (GraphForge); combine content, structure, rendering, and specification recovery (SlideRL). & Artifact feedback on factual support, presentation, and fulfillment of the brief. \\*
\addlinespace[3pt]
\MethodEntry{CompoWorld}{compoworld} & Give more weight to rubric items with lower pass rates within a rollout group. & Partial-progress reward; the maximum is reached only when all criteria pass. \\*
\addlinespace[3pt]
\MethodEntry{VERA}{vera}\MethodEntry{WorkRover}{workrover} & Check stage requirements (VERA); trace verified progress back to supporting turns (WorkRover). & Stage-level feedback and credit for earlier decisions that support verified progress. \\
\addlinespace[3pt]
\end{longtable}
\endgroup

\subsubsection{Work Environment Construction}\label{task-env-synthesis}

Training environments convert work requirements into tasks that agents can execute and evaluate. Each task includes an initial workspace, available actions, a request, and checks for evaluating the results. These components can be constructed from existing work records. For example, daVinci-Env reconstructs software tasks from repositories, and Spreadsheet-RL uses paired workbooks and an Excel runtime~\cite{davincienv,spreadsheetrl}. Real-world tasks often depend on information spread across documents, applications, and people. Therefore, training environments should preserve these dependencies so that agents can learn which actions to take and how to verify their results.

\textbf{Grounding tasks in work materials.}
Recent pipelines connect work materials to requirements in several ways. NexForge organizes task generation around capabilities and task forms, while WorkGenesis builds occupational requests with supporting resources~\cite{nexforge,workgenesis}. GraphForge represents evidence across retrieved files as a graph, then uses it to generate instructions and source-linked rubrics~\cite{graphforge}. WorkForge starts from expert workflow plans, retrieves real files, and identifies facts that guide the instructions, solution plans, and programmatic and semantic checks~\cite{workforge}. WorkBorne organizes tasks around artifact transitions: a specified actor transforms input materials toward reference outputs~\cite{workborne}.
Together, these approaches cover different aspects of task construction, from generating realistic requirements to grounding them in evidence and defining the expected results.

\textbf{Constructing shared and changing state.}
Tasks spanning multiple applications also require consistent states and operations. Agent World Model creates database-backed tool environments, EnvCraft builds sequences of dependent tasks, and AgentMercury constructs business environments with shared entities and constraints~\cite{agentworldmodel,envcraft,agentmercury}. CoworkEnv jointly generates requests, service records, completion checks, and permitted actions, then validates their consistency before collecting teacher trajectories~\cite{coworkenv}. Cowork-Gym introduces a stateful user who can provide information, revise requirements, and modify the shared workspace~\cite{coworkgym}. Such changes can affect later actions. For example, an updated customer record may change which operation is valid, while a revised workbook may invalidate an earlier analysis. WorkRover repeatedly checks and repairs consistency among the context, task, harness, and verifier~\cite{workrover}. These methods highlight the importance of maintaining a valid and consistent environment as tasks progress.

CompoWorld extends environment construction to tasks involving multiple services~\cite{compoworld}. It builds reusable services with typed states and common interfaces, uses a world model for tools that are difficult to implement reliably, and samples dependency graphs across services. Each service maintains its own state, while the agent transfers information between services through observations and actions. This allows cross-service dependencies to be controlled alongside the number of services. In a matched comparison using the same backbone, training procedure, and 1K SFT trajectories, AutomationBench success increases from 12.83\% with single-service environments to 33.33\% with composed environments.

\textbf{Scaling coverage, depth, and validation.}
Training environments can scale by adding more workspaces, more tasks within each workspace, or more rollouts per task. These choices expose agents to different materials, task dependencies, and ways to complete or recover from work. WorkBorne compares broader artifact coverage with more transitions per corpus under matched trajectory budgets~\cite{workborne}. Broader coverage performs better at 1K, 3K, and 5K trajectories, while both strategies achieve similar results at 8K. WorkForge reports that increasing training data from 1K to 16K trajectories improves ALE by 13.0 points and WildClaw by 8.4 points~\cite{workforge}. Its comparisons using equal numbers of trajectories also favor longer interactions, although these trajectories differ in task content and decision depth. Together, these findings highlight the importance of tracking both task diversity and interaction structure alongside training volume.

Reported dataset sizes also refer to different outputs of generation pipelines. WorkForge constructs 16,665 tasks, GraphForge retains 2,169 training trajectories from 2,153 selected tasks, and WorkBorne produces 7,882 task environments and 15,485 trajectories from 1,061 artifact corpora~\cite{workforge,graphforge,workborne}. These counts are difficult to compare without specifying what is measured and at which stage. Thus, we suggest reporting generation attempts, accepted tasks, usable trajectories, and verification costs to assess how efficiently environment construction produces useful training experience.

Validation is equally important as environments scale. An executable task may still contain incomplete requirements, inconsistent sources, or a verifier that accepts incorrect results. Validation must establish both that a task can be completed and that its verifier accepts correct solutions while detecting errors. It must also cover any new materials introduced during scaling. In WorkForge's matched 1K-trajectory ablation, combining programmatic and semantic checks achieves an ALE score of 22.11, compared with 15.62 for semantic-only and 14.89 for programmatic-only selection~\cite{workforge}. GraphForge's corruption audit shows that deleting a worksheet reduces the target rubric score by 0.377, while finer row-level, numerical, and citation errors change scores by less than 0.03 on a 0--1 scale~\cite{graphforge}. These results suggest combining source-based judgments with exact checks of critical values and using deliberate errors to test verifier reliability. Failed checks and their dependencies can also guide environment repairs, as discussed in Section~\ref{self-improving}.

Beyond scale and validation, training environments vary in difficulty, openness, persistence, and business fidelity. Difficulty reflects the decisions and dependencies involved in completing a task, while openness reflects what the agent must discover during execution. Persistence concerns how long the task state and responsibilities remain relevant, and business fidelity describes how closely the materials and constraints reflect real work. Human participants may also provide clarification, corrections, approvals, or final assessments at different stages. A useful training set should combine controlled tasks that expose specific failures with interactive workspaces that require multiple capabilities. Evaluation on new business tasks can then test whether these capabilities transfer to unfamiliar materials and conditions.

\subsubsection{Corpus and Trajectory Curation}\label{trajectory-curation}

Corpora provide the materials agents work with, while trajectories record how agents use those materials to complete tasks. Documents, code, and completed artifacts offer domain knowledge and examples, while interaction records capture observations, actions, and their results. Curation selects and organizes these sources according to the capabilities being trained. For example, repository-level data can link issues to relevant files and edits, while execution trajectories also capture tool outputs and failed attempts leading to those changes~\cite{davincidev}. Preserving this context helps models learn what actions to take and why. Task, workspace, and trajectory identifiers support coverage analysis and data splits based on shared sources. Interactive tasks also require records of changing requirements and actions taken by other participants~\cite{coworkgym}. CoworkEnv collects teacher demonstrations from validated environments, where SFT improves the mean Claw-Eval score from 46.6 to 64.1. The same environments also support student RL comparisons~\cite{coworkenv}.

Generated trajectories should preserve the dependencies between actions. AgentTrek uses tutorials and guided replay to construct web interaction trajectories~\cite{agenttrek}. EnterpriseLab derives tool dependencies from enterprise schemas and reuses identifiers returned by earlier calls in later operations~\cite{enterpriselab}. TDScaling organizes tools into business scenarios and generates examples for underrepresented domains and reasoning patterns~\cite{tdscaling}. Exploration offers another way to collect experience. SEE builds a verified graph of mobile interface transitions and generates tasks involving interruptions and recovery~\cite{see}. Scenario-based generation provides workflows with clear goals, while exploration captures states that standard demonstrations may miss. Successful trajectories show how to complete tasks, and recovery segments demonstrate how to proceed when execution is disrupted.

Data selection should reflect the capabilities being trained. Artifact checks can filter for successful outcomes, while failed attempts may still provide useful examples for recovery and state tracking. OpenThoughts-Agent evaluates task sources, teachers, data mixtures, and filtering strategies through downstream training, including comparisons under matched token budgets~\cite{openthoughtsagent}. CurateEvo uses development-set failures to improve data augmentation, filtering, and refinement~\cite{curateevo}. As policies improve, repeated easy successes provide less useful information, while errors involving changed inputs, missed requirements, or costly recovery can guide further data collection. Construction and development data should remain separate from held-out evaluation tasks. When tasks share files, splitting data by workspace or source family provides a stronger test of whether models can generalize to unfamiliar materials and requirements.

\subsubsection{Feedback Design}\label{reward-signals}

Feedback turns work requirements into learning signals.  When success can be measured from the resulting workspace, direct checks are possible. Spreadsheet-RL recalculates edited workbooks in Excel and compares relevant regions against target workbooks~\cite{spreadsheetrl}. Anchor derives business tasks and verifiers from constraint programs to check feasibility, solution quality, and relationships among records~\cite{anchor}. Other tasks require judgments about content and presentation. Preference-based reward learning uses human comparisons~\cite{rlhumanprefs}, GraphForge links rubric items to source evidence, and SlideRL evaluates slide structure, rendering, and content~\cite{graphforge,sliderl}. CogEvol also checks rendered content and browser interactions~\cite{cogevol}. The appropriate feedback depends on the output, which may require checks of its values, organization, and behavior. Requirements specify what counts as success, while verifiers assess the supporting evidence. Keeping these roles separate allows missing checks or parsing errors to be corrected without changing the task itself.

Feedback can also focus on requirements that agents frequently miss. The Completion-Focused Rubric Reward of CompoWorld assigns greater weight to criteria with lower pass rates within a rollout group~\cite{compoworld}. It rewards partial progress but gives the maximum reward only when all criteria are satisfied. This encourages agents to address the remaining steps needed to complete a workflow.

Final outcomes alone do not reveal which actions contributed to success. Intermediate feedback helps connect results to earlier decisions~\cite{verifystepbystep,thinkprm}. VERA restores workspace states and checks intermediate requirements using executable tests and rubrics~\cite{vera}. WorkRover's Verifier-to-Turn Credit Assignment uses recorded dependencies to assign bounded positive credit to earlier turns that contributed to verified progress~\cite{workrover}. For example, reading a source may receive credit if its information supports a later correct change. Stage checks identify when progress occurs, while dependency records help determine which earlier decisions contributed to it. Both require sufficient records of workspace states and their relationships.

Reliable feedback must also account for changing requirements. Cowork-Gym links requirements to success conditions, supporting evidence, and checkers while allowing users to introduce new constraints~\cite{coworkgym}. Maintaining current requirements alongside relevant artifacts helps agents detect omissions when earlier instructions leave the active context or source data changes. Verification also has a cost. Structural and state checks can detect common errors efficiently, while semantic inspection and human assessment handle requirements that need more context. As execution becomes faster and more parallel, verification may become a bottleneck. Thus, we propose using dependencies among sources and artifacts to determine which checks need to be repeated, which evidence remains valid, and when further assessment is necessary. Feedback design should also account for the verification and rework required before an output can be accepted.

\subsection{Model Optimization}\label{model-optimization}

\begin{figure}[!t]
    \centering
    \includegraphics[width=\linewidth]{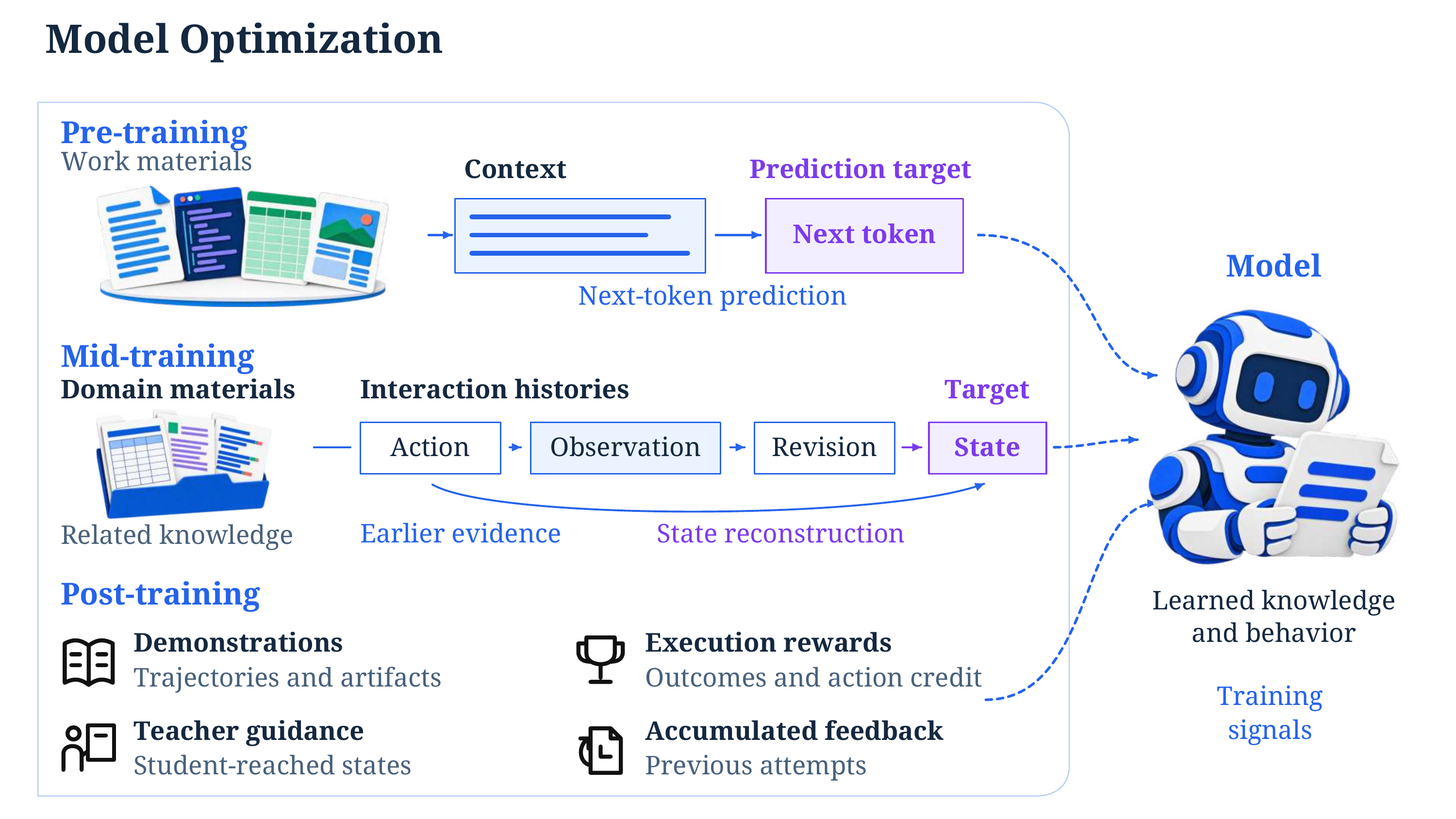}
    \captionof{figure}{Training emphases that develop the model. Illustrative samples distinguish visible context from prediction targets and feedback. Pre-training builds semantic, computational, structural, and visual knowledge. Mid-training enriches domain knowledge, introduces interaction patterns, and develops long-context use in preparation for subsequent optimization. Post-training uses demonstrations, execution rewards, teacher guidance, and accumulated feedback to supervise decisions, trajectories, artifacts, or earlier attempts. Links to the model denote learning signals, not live execution. Inputs and objectives can overlap across stages; the samples do not prescribe a shared dataset or universal training recipe. Supervision granularity differs across the four post-training groups.}
    \label{fig:model-optimization}
\end{figure}

Model optimization develops the knowledge and decision-making capabilities agents need to perform work. Pre-training builds broad semantic and computational knowledge; mid-training adapts this foundation to domain materials, interaction patterns, and long context; post-training strengthens behavior through demonstrations, execution feedback, and teacher guidance. Figure~\ref{fig:model-optimization} connects these roles. The model first learns about work objects, then how to use their interaction histories, and how to improve decisions from task outcomes. The stages differ in their data mixtures, prediction targets, and training schedules, while token prediction and agent trajectories contribute across stages.

\subsubsection{Pre-training}\label{pre-training}

Pre-training establishes the world knowledge and general capabilities that work agents rely on to understand unfamiliar tasks and environments. Through large-scale learning from text, code, mathematics, and other materials, models acquire factual knowledge, domain concepts, common conventions, and computational skills. These are essential for professional work, where requests often assume knowledge that users do not explicitly provide. For example, an agent revising a budget report must understand financial categories and units, interpret relationships between tables and written claims, and recognize how formulas or code produce numerical results. This background knowledge allows agents to interpret workspace materials, understand task requirements, and reason about possible actions before adapting to specific workflows. Work materials can in turn enrich pre-training with structured and multimodal prediction signals.
For autoregressive language models, pre-training uses the standard next-token prediction (NTP) objective. Given a token sequence $x$, the model learns to predict each token from its preceding context by minimizing
\begin{equation}
\mathcal{L}_{\mathrm{NTP}}(\theta)
= -\mathbb{E}_{x\sim\mathcal{D}}\sum_{t=1}^{|x|}\log p_{\theta}(x_t\mid x_{<t}),
\label{eq}
\end{equation}
where $\mathcal{D}$ denotes the pre-training corpus. Under this objective, the selection and organization of training data determine what knowledge and relationships the model encounters. DeepSeek-V4, for example, combines web data with mathematics, programming, and long technical documents~\cite{deepseekv4}. Work materials further expose cell addresses, formulas, layout, and relations between inputs and computed outputs. DocLLM uses text-infilling pre-training over document text and bounding-box layout~\cite{docllm}. Such structure can support executable prediction targets: TAPEX pre-trains a model to predict SQL execution results from queries and serialized tables, improving downstream table reasoning~\cite{tapex}. This illustrates how structured data can develop computational skills alongside factual knowledge.

Additionally, work agents must recognize relevant knowledge when it appears in different forms. A task requirement may be described briefly in a user request, explained in detail in a reference document, or reflected in the content of an output artifact. Kimi K2 addresses this challenge by rewriting knowledge-rich passages in different styles and from different perspectives while preserving their original meaning~\cite{kimik2}. Its early-checkpoint experiment reports 28.94\% SimpleQA accuracy for ten distinct rephrasings used once, versus 23.76\% for ten repetitions of the original text. For work agents, this principle could extend to aligned views of the same content, such as source cells, rendered charts, and report passages. Preserving their correspondences could provide training signals for maintaining semantic consistency across formats and associating claims with supporting data.

Working environments also contain visual information that agents must interpret alongside text and code. Kimi K2.5 jointly trains on text and visual inputs, including screenshots and image--code pairs; its pre-training corpus also includes dense OCR, multi-page documents, GUI screenshots, and fine-grained grounding annotations~\cite{kimik25}. Dense work artifacts require reading small text, distinguishing visually similar elements, and interpreting spatial relationships. Pairing rendered workbooks or slides with native object structure could extend this supervision to locating cells, interpreting chart elements, and connecting visible content to its underlying representation. Such data could strengthen transferable perception and understanding capabilities (Figure~\ref{fig:working-capabilities}), establishing the knowledge foundation for subsequent reasoning, planning, execution, and verification.

\subsubsection{Mid-training}\label{mid-training}

\textbf{Targeted Knowledge Enrichment.}
Work agents often encounter tasks that require knowledge beyond the information provided in a user request. Understanding an unfamiliar repository, interpreting business records, or following domain-specific conventions may depend on knowledge that is underrepresented in general pre-training data. Mid-training addresses this gap by increasing the density of specialized and long-tail knowledge while preserving the relationships among relevant materials. GLM-5 organizes related files, issues, pull requests, and commits into training sequences, then increases the proportion of long documents and synthetic agent trajectories in later stages~\cite{glm5}. Step 3.5 Flash begins with software engineering and tool-use data at 32K context, retaining 21\% pre-training replay, before expanding to 128K with long documents, synthetic reasoning, and agent data~\cite{step35flash}. Kimi K2.5 similarly combines high-quality text and multimodal materials while extending context from 32K to 256K~\cite{kimik25}. These approaches increase exposure to specialized work knowledge while using replay to preserve broader capabilities. For work agents, organizing knowledge around related files, requirements, and revisions is particularly useful, as understanding a work object often requires information distributed across multiple sources.

\textbf{Agent Interaction Familiarization.}
Domain knowledge alone cannot prepare an agent for the interactions required to complete work. An agent must also understand what a tool call changes, how to interpret its response, and what to do when an operation fails or produces an unexpected result. Mid-training can introduce these interaction patterns before post-training begins to optimize task behavior. DeepSeek-V4 incorporates agentic coding data at this stage~\cite{deepseekv4}. Nemotron combines repositories, pull requests, edit histories, and build logs in a 500B-token mixture at 256K context, including 10\% agent traces and 20\% pre-training replay~\cite{nemotronmidtraining}. With subsequent SFT held fixed, this stage improves SWE-bench Verified Pass@1 from 60.7\% to 64.5\%. daVinci-Dev further combines trajectories reconstructed from pull requests with trajectories collected through actual tool execution~\cite{davincidev}. Its mid-training uses both successful and failed executions, while later SFT focuses on test-passing trajectories. This distinction is important for work agents, as unsuccessful interactions can expose errors and intermediate states that success-only demonstrations may omit. Learning from these records helps models become familiar with the observation, action, and state-change patterns involved in execution. This provides a stronger starting point for post-training, where feedback can focus on improving decisions and completing tasks reliably.

\textbf{Long-Context Learning and Post-Training Readiness.}
Work agents frequently perform tasks in which later decisions depend on information or changes from much earlier interactions. For example, modifying a source workbook may affect several calculations and invalidate parts of a previously generated report. Simply extending the context window does not ensure that a model can identify these dependencies or reconstruct the current workspace state. TrajLong addresses this problem by transforming interaction histories into retrieval, information-combination, and file-state reconstruction tasks using full-token next-token prediction~\cite{trajlong}. Under a fixed 40B-token budget and identical later SFT, these tasks outperform raw-trajectory training on all seven reported ablation benchmarks. Function-aware fill-in-the-middle training similarly uses code dependencies to select functions or function groups for reconstruction, improving the mean across SWE-bench Verified and Lite from 13.95 to 15.60 in a controlled 7B comparison~\cite{functionfim}. The choice of prediction targets also matters. In Nemotron's 100B-token ablation, masking tool responses while retaining them as context raises SWE-bench Verified Pass@1 from 27.5\% to 37.8\% after identical SFT~\cite{nemotronmidtraining}. This distinction is relevant to agent training because tool responses provide evidence for subsequent decisions, even when the model does not need to generate those responses itself. Efficient access to long histories presents another challenge. Qwen3.8-Flash-Next introduces sparse attention during continued training at 256K context to reduce the cost of accessing relevant information~\cite{qwen38next}, connecting this stage to the memory mechanisms in Section~\ref{memory-skill-optimization}. Ultimately, the value of mid-training should be measured by its contribution to later optimization. Step 3.5 Flash reports stronger and more stable RL gains after incorporating task knowledge and tool-use patterns~\cite{step35flash}, while Nemotron compares recipes through identical SFT runs because intermediate model metrics do not reliably predict downstream agent performance~\cite{nemotronmidtraining}.

\subsubsection{Post-training}\label{post-training}

Post-training develops the policies that allow work agents to apply their knowledge to real tasks. Beyond understanding a request, an agent must select appropriate actions, respond to changes in the workspace, and determine whether its results satisfy the requirements. Different training methods provide supervision for these decisions. Demonstrations show how tasks can be completed, execution rewards evaluate the consequences of actions, teacher guidance improves decisions at states encountered by the agent, and retrospective learning extracts useful information from previous attempts. Table~\ref{tab:post-training-mechanisms} summarizes these methods according to their learning signals and update mechanisms.


\textbf{Learning from demonstrations.}\label{sft}
Supervised fine-tuning (SFT) teaches agents to follow instructions, use tools, and produce completed work through examples~\cite{agenttuning,agentflan,fireact}. Synthetic instructions provide additional supervision before task-specific interaction~\cite{selfinstruct}. For work agents, demonstrations can capture both the steps taken during execution and the structure of an accepted deliverable. Kimi K2 generates tool-use trajectories by constructing domains, tools, and tasks, simulating interactions, and filtering the results, while also incorporating real execution environments into post-training~\cite{kimik2}. Interactive demonstrations teach agents how to respond to observations and tool feedback~\cite{uitars,uimate}. Artifact-based supervision instead connects task requirements to completed outputs. CogEvol trains on verified representations of slides and interactive pages~\cite{cogevol}, while GraphForge uses source-linked rubrics to select generated solutions for rejection fine-tuning~\cite{graphforge}. CoworkEnv provides curated teacher executions that improve multi-turn performance on Claw-Eval, with the same executable environments supporting subsequent RL~\cite{coworkenv}. Together, these approaches teach agents how to carry out tasks and what acceptable results should look like.

\textbf{Learning through execution rewards.}\label{agentic-rl}
Reinforcement learning improves an agent's policy through the outcomes of its own interactions~\cite{ppo,retool,searchr1}. Web navigation, device control, and repository repair provide different environments in which actions can be executed and evaluated~\cite{webrl,digirl,swerl}. For artifact-producing tasks, Spreadsheet-RL evaluates the resulting workbook, while SlideRL combines structural, visual, and content feedback to assess generated slides~\cite{spreadsheetrl,sliderl}. However, work agents often complete tasks through many dependent steps, making it difficult to determine which earlier decisions contributed to the final result. Intermediate rewards and credit assignment help address this problem. ADMIRE derives adaptive milestones from successful GUI trajectories, while HiPER learns hierarchical policies with subgoals and associated value estimates~\cite{admire,hiper}. GiGPO combines episode-level advantages with comparisons of actions taken from repeated environment states~\cite{gigpo}. WorkRover uses recorded dependencies to assign credit to earlier turns that support verified progress~\cite{workrover}. These methods provide different ways to connect final outcomes with intermediate decisions, allowing agents to learn from both completed work and progress made along the way.

Reliable execution feedback also requires accurate records of the interactions being trained on. Mercor and SkyRL's knowledge-work recipe uses 1,928 expert-created tasks and emphasizes environment repair, exact token histories, and small-set overfitting tests before scaling RL~\cite{mercorskyrl}. Its loss ablations compare weighting at the token, sequence, and prompt-group levels, with prompt-group weighting reducing the influence of unusually long trajectories. The resulting 397B model improves held-out APEX-Agents Pass@1 from 16.11\% to 27.29\%. These practices are particularly important for work agents, where tool failures, inconsistent environments, and long execution histories can make rewards unreliable. Preserving the exact observations and actions behind each reward helps distinguish policy errors from infrastructure failures and keeps learning aligned with the interaction that actually occurred.


\textbf{Learning from teacher guidance.}\label{opd}
Demonstrations alone may not cover the states an agent encounters during its own execution, especially after incorrect actions or unexpected tool responses. On-policy distillation addresses this gap by providing teacher guidance at states reached by the student~\cite{gkd}. TCOD gradually increases student-controlled interaction, while Guided-OPD reduces teacher-controlled turns and combines teacher imitation with distillation on student-generated states~\cite{tcod,turnopd}. This allows supervision to follow the developing policy more closely~\cite{dagger,daggerllm}. Step-wise on-policy distillation adjusts guidance based on student--teacher divergence~\cite{sod}. Other methods improve supervision at specific decisions. LiteGUI selects a human-verified valid action that matches the student's attempted action, while Gated Hindsight Distillation uses a later screenshot to guide an earlier decision~\cite{litegui,ghd}. Foundation-model training also uses on-policy distillation to combine specialized capabilities or preserve them across training stages~\cite{deepseekv4,glm5}. For work agents, these methods are useful because guidance can address errors and partial progress encountered during actual execution, rather than relying only on previously collected demonstrations.

\textbf{Learning from accumulated feedback.}
Work agents can also learn from information accumulated across attempts, including corrections, execution feedback, and completed results~\cite{star,selfrewarding,agentq}. SDPO provides textual feedback to a self-teacher and distills its improved predictions into the acting policy, while Skill-SD additionally conditions the teacher on short summaries of earlier attempts~\cite{rlsd,skillsd,lu2026sdar}. RePro uses completed attempts to teach models which intermediate decisions indicate useful progress~\cite{repro}. These approaches reuse information revealed during execution to improve future behavior~\cite{retrodistill}. For example, when generating a report from multiple sources, demonstrations can teach the initial workflow, execution rewards can assess the completed report, and teacher guidance can help correct errors encountered during revision. Retrospective feedback can then turn these corrections into useful experience for subsequent tasks. Section~\ref{self-improving} extends this process beyond model updates to systems that also improve their environments, external procedures, and methods of learning.

\begingroup
\fontsize{7.6}{9.4}\selectfont
\setlength{\tabcolsep}{3pt}
\renewcommand{\arraystretch}{1.12}
\arrayrulecolor{tableRule}
\setlength{\LTpre}{7pt}
\setlength{\LTpost}{10pt}
\begin{longtable}{@{}>{\raggedright\arraybackslash}m{0.22\linewidth}>{\raggedright\arraybackslash}m{0.33\linewidth}>{\raggedright\arraybackslash}m{0.42\linewidth}@{}}
\caption{Post-training methods for work agents, grouped by their supervision and update mechanism. The final column describes the information used for learning and its purpose.}\label{tab:post-training-mechanisms}\\
\toprule
\rowcolor{tableHeader}
\textcolor{tableInk}{\bfseries Representative work} & \textcolor{tableInk}{\bfseries Learning mechanism} & \textcolor{tableInk}{\bfseries Learning signal and purpose} \\
\midrule
\endfirsthead
\multicolumn{3}{@{}l}{\small\textit{Table \thetable\ continued.}}\\
\toprule
\rowcolor{tableHeader}
\textcolor{tableInk}{\bfseries Representative work} & \textcolor{tableInk}{\bfseries Learning mechanism} & \textcolor{tableInk}{\bfseries Learning signal and purpose} \\
\midrule
\endhead
\midrule
\multicolumn{3}{r@{}}{\footnotesize\textit{Continued on the next page}}\\
\endfoot
\bottomrule
\endlastfoot

\rowcolor{tableHeader}
\multicolumn{3}{@{}l@{}}{\rule{0pt}{12pt}\textcolor{tableInk}{\bfseries Demonstrations and accepted work products}}\\*
\MethodEntry{AgentTuning}{agenttuning}\MethodEntry{Agent-FLAN}{agentflan} & Supervised mixtures of agent interactions. & Action formats, tool-use patterns, and instruction-following behavior. \\
\addlinespace[3pt]
\MethodEntry{CogEvol}{cogevol}\MethodEntry{CoworkEnv}{coworkenv} & Supervision from checked artifacts or curated teacher executions. & Deliverable structure and multi-turn behavior grounded in an executable task. \\
\addlinespace[3pt]
\MethodEntry{GraphForge}{graphforge} & SFT followed by rubric-selected rejection fine-tuning. & Source-linked acceptance criteria select useful self-generated solutions. \\
\addlinespace[3pt]
\rowcolor{tableHeader}
\multicolumn{3}{@{}l@{}}{\rule{0pt}{12pt}\textcolor{tableInk}{\bfseries Execution rewards and credit assignment}}\\*
\MethodEntry{Spreadsheet-RL}{spreadsheetrl}\MethodEntry{SlideRL}{sliderl} & Policy rollouts receive artifact-level rewards. & Numerical, structural, visual, and semantic properties of the produced artifact. \\
\addlinespace[3pt]
\MethodEntry{ADMIRE}{admire}\MethodEntry{HiPER}{hiper} & Milestone-based feedback and hierarchical subgoal learning. & Progress through multi-step tasks and decisions between subgoals. \\
\addlinespace[3pt]
\MethodEntry{GiGPO}{gigpo} & Episode advantages plus relative action credit at repeated anchor states. & Global outcome quality and local action effectiveness in comparable states. \\
\addlinespace[3pt]
\MethodEntry{WorkRover}{workrover} & Verifier-to-turn credit assignment. & Checked progress and supporting tool/artifact dependencies. \\
\addlinespace[3pt]
\MethodEntry{Mercor / SkyRL}{mercorskyrl} & Asynchronous RL with exact token accounting and group-balanced loss. & Rewards remain aligned with the sampled multi-turn interaction. \\
\addlinespace[3pt]
\rowcolor{tableHeader}
\multicolumn{3}{@{}l@{}}{\rule{0pt}{12pt}\textcolor{tableInk}{\bfseries Teacher guidance at student-reached states}}\\*
\MethodEntry{TCOD}{tcod}\MethodEntry{Guided-OPD}{turnopd} & Increase student-controlled interaction while transferring teacher guidance. & Supervision follows the states encountered by the developing policy. \\
\addlinespace[3pt]
\MethodEntry{SOD}{sod}\MethodEntry{LiteGUI}{litegui}\MethodEntry{GHD}{ghd} & Adapt teacher guidance using divergence, verified actions, or future observations. & Targets for correcting errors observed during execution. \\
\addlinespace[3pt]
\rowcolor{tableHeader}
\multicolumn{3}{@{}l@{}}{\rule{0pt}{12pt}\textcolor{tableInk}{\bfseries Feedback-conditioned and retrospective learning}}\\*
\MethodEntry{SDPO}{rlsd}\MethodEntry{SDAR}{lu2026sdar} & Feedback or compact experience summaries condition a self-teacher. & Improved continuations transfer lessons from earlier attempts to the policy. \\*
\addlinespace[3pt]
\MethodEntry{RePro}{repro} & Completed attempts supervise retrospective progress learning. & Evidence about which intermediate decisions support later success. \\
\addlinespace[3pt]
\end{longtable}
\endgroup

\subsection{Harness Optimization}\label{harness-optimization}

Work agents rely on external support to inspect information, execute operations, and manage work across multiple steps. Even with a capable model, missing evidence, unreliable tools, or poorly organized procedures can prevent successful task completion. Harness optimization improves this external support through observation and action interfaces, execution workflows, and reusable experience. Interfaces determine what the model can observe and how it interacts with the workspace. Workflows organize execution, verification, recovery, and delivery, while context management, memory, and skills preserve task progress and useful procedures across interactions. Execution traces reveal where these components need improvement, such as missing information, repeated tool errors, or unnecessary rework. Table~\ref{tab:harness-mechanisms} compares representative methods based on what they change and how feedback guides their optimization.


\begingroup
\fontsize{7.6}{9.4}\selectfont
\setlength{\tabcolsep}{3pt}
\renewcommand{\arraystretch}{1.12}
\arrayrulecolor{tableRule}
\setlength{\LTpre}{7pt}
\setlength{\LTpost}{10pt}
\begin{longtable}{@{}>{\raggedright\arraybackslash}m{0.22\linewidth}>{\raggedright\arraybackslash}m{0.33\linewidth}>{\raggedright\arraybackslash}m{0.42\linewidth}@{}}
\caption{External mechanisms for organizing and improving work execution. The middle column identifies what each method changes; the final column explains its effect on later decisions.}\label{tab:harness-mechanisms}\\
\toprule
\rowcolor{tableHeader}
\textcolor{tableInk}{\bfseries Representative work} & \textcolor{tableInk}{\bfseries Editable object and feedback} & \textcolor{tableInk}{\bfseries Effect on work execution} \\
\midrule
\endfirsthead
\multicolumn{3}{@{}l}{\small\textit{Table \thetable\ continued.}}\\
\toprule
\rowcolor{tableHeader}
\textcolor{tableInk}{\bfseries Representative work} & \textcolor{tableInk}{\bfseries Editable object and feedback} & \textcolor{tableInk}{\bfseries Effect on work execution} \\
\midrule
\endhead
\midrule
\multicolumn{3}{r@{}}{\footnotesize\textit{Continued on the next page}}\\
\endfoot
\bottomrule
\endlastfoot

\rowcolor{tableHeader}
\multicolumn{3}{@{}l@{}}{\rule{0pt}{12pt}\textcolor{tableInk}{\bfseries Observation and action interfaces}}\\*
\MethodEntry{SWE-agent}{sweagent}\MethodEntry{TACO}{taco} & Agent-oriented repository operations and execution-driven output filters. & Show the state needed for decisions while controlling inspection and context cost. \\
\addlinespace[3pt]
\MethodEntry{Trace-Free+}{tracefree}\MethodEntry{JTPRO}{jtpro} & Learned tool descriptions or coordinated instruction/description edits. & Align tool choice and arguments with the intended operation. \\
\addlinespace[3pt]
\MethodEntry{PRISM}{prism} & Prompts and executable middleware, revised from traces and state changes. & Repair recurring calling errors and enforce operation prerequisites. \\
\addlinespace[3pt]
\rowcolor{tableHeader}
\multicolumn{3}{@{}l@{}}{\rule{0pt}{12pt}\textcolor{tableInk}{\bfseries Workflows and execution control}}\\*
\MethodEntry{GEPA}{gepa}\MethodEntry{AFlow}{aflow} & Feedback-guided prompt revision and workflow-graph search. & Reorganize reasoning, tool use, and inspection across a task. \\
\addlinespace[3pt]
\MethodEntry{Meta-Harness}{metaharness}\MethodEntry{AutoDesign}{autodesign} & Harness code or design procedures, revised from repeated execution failures. & Turn recurring defects into reusable production and recovery changes. \\
\addlinespace[3pt]
\MethodEntry{JIT-Agent}{jitagent} & A learned designer produces task-specific harnesses. & Organize execution using the frozen agent's reward, latency, and cost. \\
\addlinespace[3pt]
\rowcolor{tableHeader}
\multicolumn{3}{@{}l@{}}{\rule{0pt}{12pt}\textcolor{tableInk}{\bfseries Context and reusable experience}}\\*
\MethodEntry{MemGPT}{memgpt} & Movement between active context and external storage. & Recover information whose relevance extends beyond the current context window. \\
\addlinespace[3pt]
\MethodEntry{ReasoningBank}{reasoningbank}\MethodEntry{ACE}{acecontext} & Retrieved strategies and a playbook updated over time. & Carry lessons from successful and failed executions into later tasks. \\
\addlinespace[3pt]
\MethodEntry{SkillOpt}{skillopt}\MethodEntry{SkillAlpha}{skillalpha}\MethodEntry{CodeSkill}{codeskill} & Validated or learned edits to procedural documents and skill libraries. & Improve the instructions that guide recurring work operations. \\*
\addlinespace[3pt]
\MethodEntry{SkillSmith}{skillsmith} & Library composition and the boundary between procedural skills and tools. & Coordinate reusable procedures under dependencies and a context budget. \\
\addlinespace[3pt]
\end{longtable}
\endgroup

\subsubsection{Observation and Action Interfaces}\label{interface-optimization}

Observation interfaces determine what information an agent can inspect before making a decision. For work agents, this often involves locating relevant evidence within large repositories, documents, or tool outputs. SWE-agent introduces agent-oriented repository operations that make software files easier to search, inspect, and edit~\cite{sweagent}. TACO further improves observations by discovering terminal-output filters, refining them during execution, and reusing effective rules across tasks~\cite{taco}. These filters reduce unnecessary output while preserving relevant state and error information. The main consideration is whether an interface provides the evidence needed for the next decision without consuming excessive context or requiring repeated inspection.

Action interfaces determine how an agent translates its decisions into operations on files, applications, and services. Unclear tool descriptions or inconsistent argument conventions can cause repeated errors, even when the intended operation is appropriate. Trace-Free+ trains a tool-description generator, gradually moving from supervision based on execution traces to generation using tool descriptions alone~\cite{tracefree}. JTPRO jointly optimizes global instructions and tool descriptions using execution feedback~\cite{jtpro}. The former learns tool-use conventions that can transfer across tool sets, while the latter coordinates instructions within a particular environment. Both aim to improve tool selection and argument generation, helping agents carry out the operations required by a task.

Executable support can also correct errors during tool execution. PRISM analyzes tool traces, state changes, and errors to improve prompts, middleware, or both~\cite{prism}. Middleware can correct arguments, check prerequisites, and provide error information that supports recovery. These mechanisms are particularly important when work agents modify shared records or related objects. For example, checking a file version before writing helps ensure that an edit applies to the same object the agent previously inspected. Tool responses should also indicate what changed so the agent can decide whether to continue or recover. Evaluating interfaces with the same model and tasks helps isolate improvements from external support. In the Mercor/SkyRL study, correcting tool behavior and observation handling improves the base agent before weight training, while the training gains also transfer to a held-out code-based harness~\cite{mercorskyrl}. Such comparisons help distinguish improvements in the execution environment from improvements learned by the model.

\subsubsection{Workflows and Execution Control}\label{workflow-optimization}\label{learned-harness-design}

Workflow optimization improves how an agent organizes and executes work across multiple steps~\cite{agentsquare,mesharness}. The editable components range from prompts~\cite{ape,opro,bpo} and connections between model calls~\cite{dspy,mipro} to executable harness code~\cite{adas,aevo}. Prompt optimization can use feedback-guided revision~\cite{protegi,textgrad}, reinforcement learning~\cite{rlprompt}, or evolutionary search~\cite{evoprompt,promptbreeder}. More recent approaches extend these ideas to complete execution procedures. GEPA uses execution traces and outcomes to refine prompts~\cite{gepa}, AFlow searches over workflow graphs~\cite{aflow}, and Meta-Harness modifies harness code~\cite{metaharness}. These methods provide different levels of control over how an agent performs work~\cite{tracegenopt}. For example, an improved workflow may inspect additional evidence before editing, verify intermediate results before dependent operations, or introduce a recovery step after failure. Its effectiveness should be measured by whether these changes improve execution on subsequent tasks.

Artifact production illustrates how feedback from individual tasks can improve reusable workflows. AutoDesign uses rendering, validation, and critic feedback to revise a poster, then analyzes recurring failures to propose changes to its design harness~\cite{autodesign}. This separates improving the current deliverable from improving the procedure used for future tasks. JIT-Agent takes a related approach by training a separate model to construct task-specific harnesses using execution reward, latency, and cost from a frozen executor~\cite{jitagent}. The designer learns how to organize execution and produces an explicit procedure for the acting agent. These approaches show how experience from completed work can guide the design of more effective workflows without requiring changes to the underlying executor.

Execution control must also preserve progress when work is interrupted, requirements change, or responsibilities are transferred. For ongoing tasks, a workflow needs clear conditions for continuing, revising previous results, and determining when delivery is complete. These conditions relate to the requirements and responsibilities discussed in Section~\ref{foundation}. For example, a report awaiting source confirmation should retain that dependency, while a completed deliverable should be checked against the latest requirements before the task closes. Repeated inspections, retries, and human corrections provide evidence of where workflow changes could reduce delays and rework. Methods based on retrospective self-preference, adaptive curricula, and task-dependent harness routing further use feedback to select and deploy suitable procedures~\cite{rho,activesaddler,adaptiveautoharness}. Evaluating both the quality of accepted results and the effort required to produce them helps ensure that workflow optimization improves real task execution.

\subsubsection{Long Context, Memory and Skills}\label{memory-skill-optimization}

Context management determines which parts of a growing work history remain available for each decision. Work agents may accumulate lengthy tool outputs, repeated observations, and information that becomes outdated as the workspace changes. A larger context window alone does not resolve these issues. MemGPT manages information between active context and external storage, while TACO adjusts the amount and form of tool output retained during execution~\cite{memgpt,taco}. For ongoing work, the active context must also preserve current requirements, relevant object versions, and unresolved dependencies. These records help agents determine whether earlier observations and decisions remain valid. Effective context management therefore supports efficient access to information and reliable continuation after interruptions or revisions.

Memory optimization focuses on preserving useful experience across tasks. ReasoningBank extracts strategies from successful and failed attempts for later retrieval~\cite{reasoningbank}, while ACE maintains an evolving playbook through generation, reflection, and curation~\cite{acecontext}. Skills extend this idea by representing reusable procedures that guide future execution. SkillOpt improves procedural documents through validated edits~\cite{skillopt}, SkillAlpha learns editing policies and evaluates them through downstream execution~\cite{skillalpha}, and CodeSkill extracts, revises, and maintains a library of coding skills~\cite{codeskill}. These approaches allow agents to reuse lessons from previous work rather than rediscovering procedures for every task. Execution feedback also helps determine which instructions remain useful and which should be revised as new situations arise.

As reusable experience accumulates, agents must decide which procedures to retain, how to combine them, and when to apply them. SkillSmith addresses interactions and dependencies among skills under a limited context budget and considers which operations should be represented as skills or executable tools~\cite{skillsmith}. Stable operations can be implemented as tools, while skills provide guidance on when and how to use them. This distinction is important for work agents that repeatedly perform related tasks with different materials and constraints. Task-specific state and reusable experience also require different update schedules. A change to a source workbook should immediately update the current task record, while a revised reporting procedure should be validated across multiple tasks before broader reuse. Recording the sources and dependencies of both enables targeted updates. For example, a reporting skill may describe how to reconcile information from several workbooks, while the task record identifies which workbook version has been checked and which approval remains pending. Maintaining this distinction allows agents to reuse established procedures while keeping their execution aligned with the current workspace.

\subsection{Model--Harness Coupling and Joint Optimization}\label{joint-optimization}

\begin{figure}[!t]
    \centering
    \includegraphics[width=\linewidth]{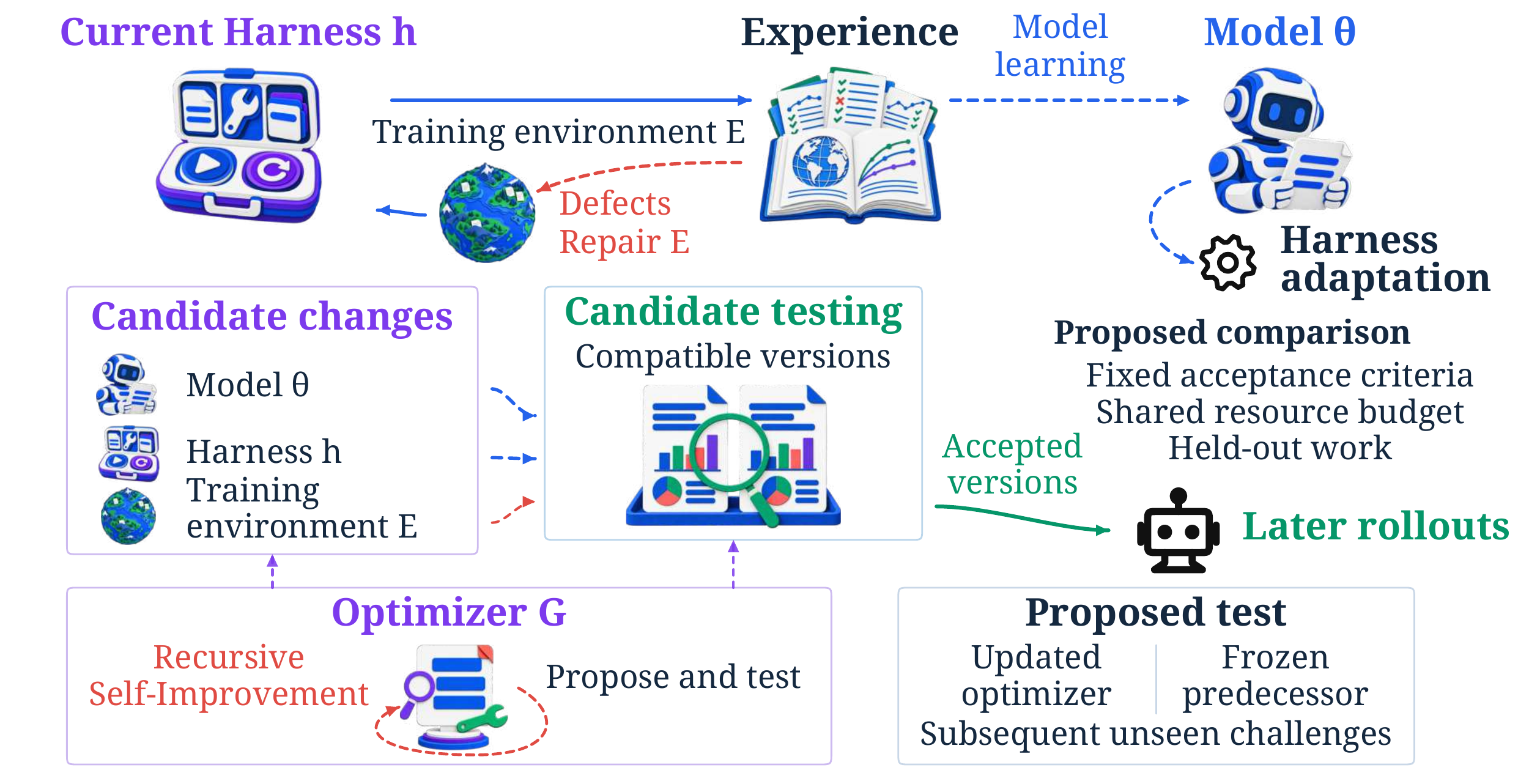}
    \captionof{figure}{Model--harness coupling and improvement across training executions. The current harness shapes experience used to train the model; model updates can motivate harness adaptation. Evidence informs candidate changes to executor weights $\theta$, harness $h$, or training environment $E$. Optimizer $G$ proposes and tests changes, and accepted versions supply subsequent rollouts. The paths illustrate possible component updates, without prescribing their order or requiring every target to change. A separate relation denotes recursive self-improvement of $G$'s search and diagnostic/testing procedures. Fixed acceptance criteria, held-out work, and a shared resource budget are our proposed comparison conditions. We further propose comparing the updated optimizer with its frozen predecessor on subsequent unseen challenges.}
    \label{fig:recursive-improvement}
\end{figure}

The effectiveness of a work agent depends on how well its model and harness work together. The model must learn to use the information and tools provided by the harness, while the harness must offer interfaces and procedures that the model can use reliably. This raises an important question of where improvements should be implemented. A capability may be learned through model training, provided through retrieved context, encoded as a reusable skill, or implemented directly in control code. We examine how models are trained within a harness, how the two components can be optimized together, and how repeated execution can improve both the system and its optimization process. Table~\ref{tab:joint-optimization} summarizes the components updated by these methods, while Figure~\ref{fig:recursive-improvement} illustrates their feedback loops.


\begingroup
\fontsize{7.6}{9.4}\selectfont
\setlength{\tabcolsep}{3pt}
\renewcommand{\arraystretch}{1.12}
\arrayrulecolor{tableRule}
\setlength{\LTpre}{7pt}
\setlength{\LTpost}{10pt}
\begin{longtable}{@{}>{\raggedright\arraybackslash}m{0.22\linewidth}>{\raggedright\arraybackslash}m{0.33\linewidth}>{\raggedright\arraybackslash}m{0.42\linewidth}@{}}
\caption{Joint adaptation and self-improvement, organized by the component changed in each update. $\theta$ denotes executor weights, $h$ its harness, $\phi$ a separate designer's weights, $E$ a training environment package, and $G$ the optimizer's procedure. The last group gives within-study comparisons of the role and cost of verification.}\label{tab:joint-optimization}\\
\toprule
\rowcolor{tableHeader}
\textcolor{tableInk}{\bfseries Representative work} & \textcolor{tableInk}{\bfseries Updated object} & \textcolor{tableInk}{\bfseries How updates use feedback} \\
\midrule
\endfirsthead
\multicolumn{3}{@{}l}{\small\textit{Table \thetable\ continued.}}\\
\toprule
\rowcolor{tableHeader}
\textcolor{tableInk}{\bfseries Representative work} & \textcolor{tableInk}{\bfseries Updated object} & \textcolor{tableInk}{\bfseries How updates use feedback} \\
\midrule
\endhead
\midrule
\multicolumn{3}{r@{}}{\footnotesize\textit{Continued on the next page}}\\
\endfoot
\bottomrule
\endlastfoot

\rowcolor{tableHeader}
\multicolumn{3}{@{}l@{}}{\rule{0pt}{12pt}\textcolor{tableInk}{\bfseries Policy and harness adaptation}}\\*
\MethodEntry{UI-Copilot}{uicopilot}\MethodEntry{Agent Lightning}{agentlightningv1} & $\theta$ within a supplied execution harness. & Training consumes decisions made through the available observations, tools, and control flow. \\
\addlinespace[3pt]
\MethodEntry{BetterTogether}{bettertogether}\MethodEntry{Fast-Slow Training}{fastrl} & Prompts in $h$ and executor weights $\theta$. & Alternate external prompt search with model updates, refreshing the conditions for each. \\
\addlinespace[3pt]
\MethodEntry{Co-Harness}{coharness}\MethodEntry{VERA}{vera} & $\theta$ and external procedures or skills in $h$. & Improved procedures shape training rollouts; policy outcomes guide later external revisions. \\
\addlinespace[3pt]
\MethodEntry{UI-Mem}{uimem}\MethodEntry{SkillRL}{skillrl}\MethodEntry{SLIM}{slim} & $\theta$ and reusable memory or skills. & Experience retained outside the model guides learning and changes as the policy improves. \\
\addlinespace[3pt]
\MethodEntry{SIA}{sia} & Select an update to $\theta$ or $h$. & Failure traces and reward changes guide the choice between policy learning and scaffold edits. \\
\addlinespace[3pt]
\rowcolor{tableHeader}
\multicolumn{3}{@{}l@{}}{\rule{0pt}{12pt}\textcolor{tableInk}{\bfseries Learned designers and adaptive practice}}\\*
\MethodEntry{JIT-Agent}{jitagent}\MethodEntry{DMRL}{dmrl} & Designer weights $\phi$; generated harnesses or skill documents in $h$. & Frozen-executor outcomes train a separate design or editing policy; DMRL uses business feedback. \\
\addlinespace[3pt]
\MethodEntry{Socratic-SWE}{socraticswe} & Shared generation/solving policy and a trace-derived skill registry. & Role-specific objectives connect task generation to the solver's experience. \\
\addlinespace[3pt]
\MethodEntry{CurateEvo}{curateevo} & Data augmentation, filtering, and refinement procedures. & Development failures guide which training material is retained and produced next. \\
\addlinespace[3pt]
\rowcolor{tableHeader}
\multicolumn{3}{@{}l@{}}{\rule{0pt}{12pt}\textcolor{tableInk}{\bfseries Environment evolution and optimizer-level recursion}}\\*
\MethodEntry{WorkRover}{workrover} & Environment components in $E$; separate policy post-training. & Versus applying all repairs: cross-task regressions are 0.7\% versus 7.0\%; normalized model-token cost is 1.00 versus 0.59. \\*
\addlinespace[3pt]
\MethodEntry{VERSE}{verse} & Executor harness $h$ and optimizer procedure $G$; model weights fixed. & Controlled study, test accuracy (\%): baseline 39.20; verification only 39.81; self-evolution only 31.79; both 41.05. One evolution run per condition. \\
\addlinespace[3pt]
\end{longtable}
\endgroup

\subsubsection{Harness-Aware Policy Training}\label{harness-aware-training}

A training harness determines what a policy can observe and do. Tool schemas define available operations and their arguments, context management controls access to previous interactions, and execution rules determine how actions affect the workspace. Harness-aware training allows models to learn under these conditions. Spreadsheet-RL trains agents with a spreadsheet runtime and toolset, while UI-Copilot provides structured observations, progress summaries, and a copilot for retrieval and calculation~\cite{spreadsheetrl,uicopilot}. Through these interactions, the policy learns when to rely on internal reasoning and when to use external tools. The deployment harness must therefore provide the interfaces and support that the policy has learned to use.

Training systems can also optimize a policy while preserving the agent's existing execution process. Agent Lightning connects policy optimization with agent execution, allowing the harness to manage interactions while the training system learns from the resulting decision records~\cite{agentlightningv1}. The available tools, memory access, and execution rules all influence the experience collected during training. For example, a model trained with compressed observations needs to learn how to retrieve information that is no longer available in its active context. The training harness therefore shapes the situations and decisions the policy learns to handle, even when only model weights are updated.

Models can also learn to construct tools for future tasks. SMITH alternates between tool construction and tool use, adds generated tools to a reusable pool, and rewards their validity and downstream usefulness~\cite{smith}. As the available tools evolve, the policy learns how to use the capabilities it creates. For work agents, this provides a way to turn repeated operations into reusable executable tools while learning when to construct and invoke them~\cite{wfwa0458}. Feedback from subsequent execution helps determine which generated tools are useful, allowing the model to improve based on their contributions to later tasks.

\subsubsection{Joint Model--Harness Adaptation}\label{joint-prompt-weight}

Joint adaptation coordinates model learning with changes to external support, which can be revised at a different pace~\cite{optimas}.  BetterTogether alternates between prompt and weight optimization, while Co-Harness uses improved harnesses to generate model-training trajectories and then adapts the harness to the updated model~\cite{bettertogether,coharness}. Fast-Slow Training similarly alternates prompt search with blocks of reinforcement-learning updates~\cite{fastrl}. In these approaches, each component is optimized using experience collected with the current version of the other. This interaction matters because a procedure that compensates for an earlier model's limitations may become unnecessary after training, while a stronger model may be able to use workflows that were previously unreliable.

Memory and skills provide concrete components for joint adaptation. UI-Mem extracts experience during GUI reinforcement learning and updates memory guidance as the policy improves, while SkillRL maintains a skill library alongside policy training~\cite{uimem,skillrl}. VERA alternates rubric-guided model updates with validated additions, revisions, and removals in a skill library organized around work stages~\cite{vera}. SIA uses execution trajectories and reward changes to decide between scaffold editing and policy reinforcement~\cite{sia}. These approaches allow improvements to target the model, its external support, or both, depending on the observed failures and the capabilities already available.

An important question is where learned experience should be stored. Component Routing compares experience incorporated through weight updates with the same experience provided as context~\cite{routingexperience}. In its evaluated GUI settings, recurring patterns for locating interface elements favor weight updates, while state-dependent procedures and facts benefit more from context. The effectiveness of transferred procedures also depends on the receiving model. For work agents, this suggests learning recurring behavioral patterns in model weights while keeping changing facts, task commitments, and operating rules in external records. To evaluate these choices, we propose comparing $(M_0,H_0)$, $(M_1,H_0)$, $(M_0,H_1)$, and $(M_1,H_1)$ under consistent task conditions and compatible interfaces. For ongoing work, each configuration can resume from the same saved workspace after an identical source change or user intervention. Comparing these runs with fresh starts would reveal whether updates preserve unfinished dependencies, support necessary rechecks, and improve task completion while accounting for verification costs and human intervention.

\subsubsection{Self-Improvement and Recursive Self-Improvement}\label{self-improving}

Work agents may operate over extended periods while the environments in which they work continue to change. New tools, data formats, business rules, and organizational processes can gradually make previously reliable procedures less effective. Although context and memory can help agents recognize these changes, substantial shifts may require persistent updates to their models, workflows, skills, and verification mechanisms. Self-improvement addresses this need by using execution experience to adapt the agent and its supporting environment. Recursive self-improvement (RSI) goes further by improving the mechanisms through which the system identifies limitations, develops new capabilities, and evaluates its own progress. These two directions offer complementary benefits for work agents. Self-improvement supports continued adaptation to changing work, while RSI offers a potential path toward more fundamental growth in agent capabilities.

\textbf{Adapting the source of practice.}
Self-improvement allows agents to accumulate experience and revise the capabilities and procedures they use in later work. Socratic-SWE extracts strategies and recurring failures into a skill registry, then alternates task generation and solving through a shared policy with role-specific objectives~\cite{socraticswe}. Generated tasks guide subsequent training, while new execution traces inform later task generation. CurateEvo uses development failures to improve data augmentation and filtering~\cite{curateevo}, while SkillRL and SLIM update a procedural library alongside policy training~\cite{skillrl,slim}. These methods improve future behavior by changing what the agent practices and which procedures it can reuse. Beyond benchmark tasks, DMRL learns a policy that edits the skill instructions of a frozen advertising-system agent using online outcomes and estimates of longer-term effects~\cite{dmrl}. Such feedback connects self-improvement to operational needs, where an agent must adjust its behavior as requirements and working conditions evolve.

\textbf{Maintaining reliable training environments.}
Continued learning also requires environments that provide valid tasks and feedback. GraphForge revises generated tasks and rubrics using execution feedback before collecting training trajectories~\cite{graphforge}, while WorkForge constructs and filters environments through a forward generation pipeline~\cite{workforge}. WorkRover supports repeated environment repair by diagnosing compatibility failures and testing candidate changes through Factorial Compatibility Replay~\cite{workrover}. It selects the smallest repair set that satisfies fixed requirements and preserves necessary dependencies, then validates the result through fresh execution. Its natural-defect study reports 70.3\% safe repairs, compared with 60.7\% for full regeneration, with rates remaining within 70.3--72.4\% across three repair generations. Table~\ref{tab:joint-optimization} further compares regressions and token costs. These methods highlight the need to distinguish failures caused by the policy from defects in the training environment. An incorrect decision may require further learning, while missing materials or inconsistent verifiers require environment repair. Maintaining reliable learning environments helps agents improve from experience as their training needs change.



\textbf{From adaptive improvement to RSI.}
Self-improvement can also be guided directly by the outcomes of real work. DMRL learns a policy that edits the skill instructions of a frozen advertising-system agent using online outcomes and estimates of longer-term effects~\cite{dmrl}. This illustrates how business feedback can guide improvements to reusable procedures. RSI introduces a further possibility by allowing the optimization process itself to evolve. VERSE explores this direction by keeping the executor and optimizer model weights fixed while modifying the executor harness and the optimizer's prompts, skills, tools, hooks, and notes~\cite{verse}. Its controlled comparison evaluates execution verification and optimizer self-evolution separately and together, with the combined setting achieving the best results. Without executable checks, self-evolution selects the initial harness and produces a lower test result. Although these results are based on one evolution run per condition, they illustrate how changes to the optimization process can influence subsequent improvements.

These studies provide different perspectives on how improvement should be evaluated. DMRL connects procedural updates to business outcomes~\cite{dmrl}, WorkRover examines the reliability and cost of repeated environment repairs~\cite{workrover}, and VERSE separates the effects of optimizer evolution and executable verification~\cite{verse}. For work agents, we propose extending these comparisons to longer periods and changing work conditions. Self-improvement can be evaluated by how effectively an agent adapts while preserving unfinished responsibilities and maintaining the quality of accepted work. RSI requires an additional comparison between the updated optimizer and its frozen predecessor, testing whether improvements to the optimization process enable more effective responses to subsequent, unseen challenges. Such evaluations should measure task success, adaptation costs, verification effort, and the recurrence of previously resolved failures. Table~\ref{tab:optimization-diagnosis} links observed failures to candidate optimization targets and the work-agent capabilities discussed in Section~\ref{capabilities}.

\par\medskip
\noindent\begin{minipage}{\linewidth}
\centering
\fontsize{7.4}{9.2}\selectfont
\setlength{\tabcolsep}{3pt}
\renewcommand{\arraystretch}{1.15}
\arrayrulecolor{tableRule}
\captionof{table}{Choosing an optimization target from observed failures. The comparisons are proposed tests; the cited studies provide methods for the corresponding changes.}
\label{tab:optimization-diagnosis}
\begin{tabularx}{\linewidth}{@{}>{\raggedright\arraybackslash}m{0.25\linewidth}>{\raggedright\arraybackslash}m{0.28\linewidth}>{\raggedright\arraybackslash}X@{}}
\toprule
\rowcolor{tableHeader}
\textcolor{tableInk}{\bfseries Failure evidence} & \textcolor{tableInk}{\bfseries Candidate change} & \textcolor{tableInk}{\bfseries How to test the update} \\
\midrule
An error recurs with the required evidence visible & Train interpretation or action patterns~\cite{davincidev,step35flash} & Same harness; held-out objects and requests testing the recurring decision \\
\midrule
Evidence is absent, truncated, or difficult to retrieve & Revise observations and retrieval~\cite{sweagent,taco} & Same model; evidence recovery, task success, and context cost \\
\midrule
A dependency is lost after interruption or source revision & Revise workflows and external state storage~\cite{metaharness,memgpt} & Resume from saved state; check unfinished dependencies and affected artifacts again \\
\midrule
An artifact marked complete fails an independent requirement check & Revise requirement capture and verification~\cite{coworkgym,vera} & Independent requirement checks; missed defects, false rejection, and assessment effort \\
\midrule
Replay diagnoses incompatible context, task, harness, or verifier components & Repair the training environment~\cite{workrover} & Check that the repaired task runs, preserves requirements, and supports policy transfer to other environments \\
\bottomrule
\end{tabularx}
\end{minipage}
\par\medskip

Together with the continuation protocol, these comparisons suggest evaluating work agents across successive changes in tasks, tools, requirements, and shared workspaces. Targeted failures can test specific improvements, while held-out workflows and later business tasks can assess their transfer. Beyond immediate task success, evaluations should track preserved dependencies, necessary rechecks, human intervention, and the time and cost of accepted work. For RSI, repeated evaluations should further determine whether improvements to the optimization process help agents acquire capabilities needed for increasingly unfamiliar work. This distinction provides a basis for studying both sustained adaptation and longer-term capability growth.

\section{Evaluation and Resources: Assessing Work Agents}\label{evaluation}

Rigorous evaluation is essential for understanding the capabilities and limitations of current models on work tasks and guiding the improvement of work agents. By verifying whether delegated work meets explicit acceptance criteria, evaluation reveals where agents succeed, where they fail, and which capabilities require further development. These findings provide empirical feedback for the improvement methods discussed in Section~\ref{optimization}. To serve this purpose, evaluation must clearly define the delegated work, specify its acceptance criteria, and collect appropriate evidence of completion. Answers, files, application state, and complete workflows have different requirements, so assessment must examine the relevant deliverables, workspace state, and execution process rather than rely on the final response alone. Because measured performance depends jointly on the model, tool interface, execution environment, and verification procedure, benchmark scores must be interpreted alongside the tasks and protocols that produced them.

This section covers task construction and domain coverage (Section~\ref{benchmarks}), the targets, dimensions, and methods used to evaluate work results (Section~\ref{evaluation-methods}), and the metrics used to summarize those judgments (Section~\ref{metrics}). It also introduces the execution frameworks and evaluation infrastructure that support task execution and verification (Section~\ref{frameworks}). The final part examines the relationships among task performance, execution cost, and economic value, including how human oversight affects deployment returns (Section~\ref{economics}).

\subsection{Benchmarks and Task Construction}\label{benchmarks}


A work-agent benchmark is scoped by two choices: how its tasks are constructed and which work domains they cover~\cite{agentbench,agentboard}. Task construction determines how requests, materials, and execution conditions are produced~\cite{mind2web,weblinx}, and thus how realistic and how controllable the resulting tasks are. Domain coverage determines the professional knowledge, work objects, and acceptance requirements involved, and thus how representative the benchmark is of the work agents are asked to perform.

\subsubsection{Task Construction Strategies}\label{task-construction-strategies}

Existing benchmarks mainly use four approaches: expert design, conversion of real requests, construction in interactive environments, and construction from professional workspaces. Each emphasizes a different task source and verification condition; Table~\ref{tab:working-agents-45} lists representative instances.

\textbf{Expert-designed and curated tasks.} Researchers or domain experts design tasks around knowledge, reasoning, and operational requirements, and supply reference answers, labels, or explicit acceptance criteria. GAIA, FinQA, and LegalBench cover information-seeking reasoning, financial computation, and legal judgment respectively~\cite{finqa,legalbench,gaia}; CADBench (BlenderLLM) refines modeling instructions into check items such as attributes and spatial relations~\cite{du2024blenderllmtraininglargelanguage}. This approach makes difficulty and requirements easy to control, but it does not necessarily include every interaction condition of a real delegation.

\textbf{Tasks built from real requests and existing task collections.} These methods extract tasks from real problems, user needs, or existing professional data. SWE-bench uses GitHub issues and their associated code repositories, and SpreadsheetBench curates workbook requests from user forums~\cite{swebench,spreadsheetbench}; CADBench (MIT) instead builds program-reconstruction tasks from existing CAD data~\cite{doris2026cadbench}. The advantage is that real work objects are preserved, although task coverage depends on the data source and the filtering procedure.

\textbf{Tasks built around applications and execution environments.} These methods place tasks in interactive software environments, constructing repeatably executable tasks by initializing application state, providing operation interfaces, and setting acceptance conditions. WebArena, OSWorld, AndroidWorld, and AppWorld provide web, desktop, mobile, and multi-application API environments respectively, and verify task results against the actual state after execution~\cite{webarena,osworld,appworld,androidworld}. Such methods can assess an agent together with the operations it actually performed, but task coverage remains limited by the environment design and by the states that can be verified.

\textbf{Tasks built from professional activities and workspaces.} These methods organize multi-step, multi-object tasks around occupational requests and actual delivery requirements. GDPval supplies professional work tasks authored by occupational experts, TheAgentCompany uses simulated enterprise services, and Workspace-Bench concerns multiple files and their dependencies~\cite{workspacebench,agentcompany,gdpval}; StartupBench and BankerToolBench further emphasize complete professional delivery and fine-grained acceptance~\cite{startupbench,bankertoolbench}. Compared with a single operation, these tasks come closer to a complex delegation, but their construction and verification costs are usually higher.

Taken together, the four approaches trade controllability for realism. Expert-designed tasks specify requirements most precisely, whereas workspace tasks reproduce the most of an actual delegation. In between, real requests contribute authentic content and environments contribute executable context. 

\begingroup
\footnotesize
\setlength{\tabcolsep}{3pt}
\renewcommand{\arraystretch}{1.12}
\begingroup
\fontsize{7.4}{9.2}\selectfont
\setlength{\tabcolsep}{2.5pt}
\renewcommand{\arraystretch}{1.12}
\arrayrulecolor{tableRule}
\Needspace{150pt}

\begin{longtable}{@{}>{\raggedright\arraybackslash}m{110.000pt}
 >{\raggedright\arraybackslash}m{115.000pt}
 >{\raggedright\arraybackslash}m{110.000pt}
 >{\raggedright\arraybackslash}m{115.000pt}@{}}
\caption{Task construction approaches, sources, and input--output forms in representative benchmarks and an associated agent system.}\label{tab:working-agents-45}\\
\toprule
\rowcolor{tableHeader}
\textcolor{tableInk}{\bfseries Benchmark / system} & \textcolor{tableInk}{\bfseries Task source} & \textcolor{tableInk}{\bfseries Input form} & \textcolor{tableInk}{\bfseries Output form} \\
\midrule
\endfirsthead
\multicolumn{4}{@{}l@{}}{\small\tablename~\thetable~(continued)}\\
\toprule
\rowcolor{tableHeader}
\textcolor{tableInk}{\bfseries Benchmark / system} & \textcolor{tableInk}{\bfseries Task source} & \textcolor{tableInk}{\bfseries Input form} & \textcolor{tableInk}{\bfseries Output form} \\
\midrule
\endhead
\midrule
\multicolumn{4}{r@{}}{\color{tableMuted}\scriptsize Continued on next page}\\
\endfoot
\bottomrule
\endlastfoot
\rowcolor{tableHeader}\multicolumn{4}{@{}l@{}}{\textcolor{tableInk}{\bfseries Expert-designed and curated tasks}}\\*
\specialrule{0.3pt}{0pt}{0pt}
\MethodEntry{GAIA}{gaia} & Cross-domain information and reasoning problems & Text prompt; optional files and external information & Concise text answer \\
\MethodEntry{DocVQA}{docvqa} & Document images with human-annotated questions & Document image and question & Text answer \\
\MethodEntry{FinQA}{finqa} & Financial reports and expert-designed questions & Question, report text, and tables & Numerical answer and reasoning program \\
\MethodEntry{FrontierFinance}{zhang2026frontierfinance} & Investor research queries written by finance professionals across six use cases & Query, as-of date, and access to filings, market data, and the web & Long-form investment research answer \\
\MethodEntry{LegalBench}{legalbench} & Legal cases, rules, and documents & Task instruction and legal text & Task-specific label or answer \\
\MethodEntry{CADBench (BlenderLLM)}{du2024blenderllmtraininglargelanguage} & Simulated modeling tasks and online forum requests & Natural-language modeling instructions & Blender script and 3D model \\
\MethodEntry{BlenderBench (VIGA)}{yin2026visionasinversegraphics} & Curated 3D scene reconstruction and editing tasks & Target images, scenes, and editing requirements & Blender scene and rendered result \\
\MethodEntry{BenchCAD}{benchcad} & Procedurally generated industrial part families anchored to engineering standards & Part images, CadQuery code, and editing instructions & CadQuery program, edited program, or numeric answer \\
\specialrule{0.5pt}{1pt}{0pt}
\rowcolor{tableHeader}\multicolumn{4}{@{}l@{}}{\textcolor{tableInk}{\bfseries Real requests and existing task collections}}\\*
\specialrule{0.3pt}{0pt}{0pt}
\MethodEntry{SWE-bench}{swebench} & GitHub issues and the associated code repositories & Issue description and codebase & Code patch \\
\MethodEntry{SWE-agent (system)}{sweagent} & Existing software-engineering tasks and execution environments & Issue description and executable repository & Code edits and patch \\
\MethodEntry{SpreadsheetBench}{spreadsheetbench} & Spreadsheet requests from user forums & Natural-language instruction and workbook & Modified workbook \\
\MethodEntry{CADBench (MIT)}{doris2026cadbench} & Existing CAD model and program datasets & Meshes, single-view or multi-view images & Executable, editable CAD program \\
\specialrule{0.5pt}{1pt}{0pt}
\rowcolor{tableHeader}\multicolumn{4}{@{}l@{}}{\textcolor{tableInk}{\bfseries Applications and execution environments}}\\*
\specialrule{0.3pt}{0pt}{0pt}
\MethodEntry{WebArena}{webarena} & Functional web applications & Request and browser environment & Answer or target website state \\
\MethodEntry{VisualWebArena}{visualwebarena} & Visual webpages and user requests & Request, browser, and visual content & Answer or target website state \\
\MethodEntry{OSWorld}{osworld} & Desktop applications and workflows & Request and initialized desktop environment & Changed files or application state \\
\MethodEntry{OSWorld 2.0}{yuan2026osworld2} & Long-horizon tasks in real desktop applications & Request, work files, and desktop environment & Cross-application state and complete deliverables \\
\MethodEntry{AndroidWorld}{androidworld} & Parameterized tasks in Android applications & Request and initialized mobile application & Changed application state \\
\MethodEntry{LOCA-bench (up to 256K)}{zeng2026loca} & Simulated MCP services and controlled state growth & Request, tool interfaces, and continuously growing context & Tool-operation results and target state \\
\specialrule{0.5pt}{1pt}{0pt}
\rowcolor{tableHeader}\multicolumn{4}{@{}l@{}}{\textcolor{tableInk}{\bfseries Professional activities and workspaces}}\\*
\specialrule{0.3pt}{0pt}{0pt}
\MethodEntry{PPTC}{pptc} & Presentation creation and editing tasks & Instructions and presentation state & Created or modified slide deck \\
\MethodEntry{PresentBench}{presentbench} & Presentation requests and background materials & Prompt and task materials & Generated slide deck \\
\MethodEntry{GDPval}{gdpval} & Professional work across occupations & Work request and reference materials & Professional deliverables \\
\MethodEntry{Workspace-Bench}{workspacebench} & Professional tasks with file dependencies & Request and multi-file workspace & Workspace deliverables \\
\MethodEntry{StartupBench}{startupbench} & Professional workflows with evidence of market adoption & Work request and workspace materials & Complete professional deliverables \\
\MethodEntry{Agents' Last Exam}{sun2026agentslastexam} & Professional tasks designed by industry experts & Work request, materials, and software environment & Verifiable professional work products \\
\end{longtable}
\endgroup
\par\endgroup

\subsubsection{Domain Coverage and Work Requirements}\label{evaluation-domains}

Task sources determine how tasks are built, not which work they cover. Domain coverage matters because the evidence needed to accept a completed task depends on the work object. In finance, law, and healthcare, the object is mainly an answer or a professional judgment, and acceptance rests on factual grounding, professional rules, and result correctness~\cite{finqa,legalbench,medagentbench}. In office and document work, it is the delivered file, whose content, native structure, and computational or visual effects must be checked~\cite{spreadsheetbench,pptc}. In customer service and enterprise operations, it is the state of business systems, subject to permissions and operational constraints~\cite{enterpriseops}.

Software engineering, data science, and scientific research usually produce executable or reproducible results, verifiable through tests, database queries, predictive outcomes, or scientific conclusions~\cite{swebench,scienceagentbench,dsbench,spider2,discoverybench}. CAD and 3D graphics additionally require program validity, geometric fidelity, and rendering quality to hold together~\cite{du2024blenderllmtraininglargelanguage,doris2026cadbench,yin2026visionasinversegraphics}. Web, desktop, mobile, and API settings, by contrast, are interaction settings reusable across domains; they assess the environment state after an operation rather than professional knowledge, and are therefore listed as a separate group~\cite{webarena,osworld,androidworld,appworld}.

Cross-domain benchmarks combine several work objects and acceptance requirements: GDPval examines occupational work products, Remote Labor Index authentic remote tasks, and Workspace-Bench and StartupBench complex workspace delivery~\cite{gdpval,rli,workspacebench,startupbench}. Table~\ref{tab:benchmark-metrics-judges} and Figure~\ref{fig:benchmark-domain-map} summarize the work objects and requirements of each group. Together, they illustrate the range from localized question answering to complete professional tasks.

Figure~\ref{fig:benchmark-domain-map} groups benchmarks by their primary work domain or setting and summarizes representative knowledge requirements.
\begin{figure}[p]
\centering
\scalebox{0.96}{%
\begin{tikzpicture}[x=\linewidth,y=1pt, every node/.style={inner sep=4pt,align=left,font=\fontsize{8}{10}\selectfont}]
\node[anchor=north west,rounded corners=3pt,draw=green!25!gray,fill=green!4,minimum height=54pt,text width=0.21\linewidth] (g0) at (0.06,0) {\textcolor{green!35!black}{\bfseries Finance \& accounting}\par{\fontsize{7}{8.5}\selectfont Reports, accounting, valuation}};
\node[anchor=north west,rounded corners=3pt,draw=green!25!gray,fill=green!4,minimum height=54pt,text width=0.69\linewidth,font=\fontsize{8}{9.6}\selectfont] (p0) at (0.29,0) {BankerToolBench~\cite{bankertoolbench}; ConvFinQA~\cite{convfinqa}; Finance Agent Benchmark~\cite{financeagentbenchmark}; Finance Agent v2~\cite{financeagentv2}; FinQA~\cite{finqa}; FrontierFinance~\cite{zhang2026frontierfinance}; TAT-QA~\cite{tatqa}};
\draw[gray!45] (0.035,-27) -- (g0.west);\draw[gray!45] (g0.east) -- (p0.west);
\node[anchor=north west,rounded corners=3pt,draw=green!25!gray,fill=green!4,minimum height=44pt,text width=0.21\linewidth] (g1) at (0.06,-56) {\textcolor{green!35!black}{\bfseries Law \& legal services}\par{\fontsize{7}{8.5}\selectfont Statutes, precedents, legal practice}};
\node[anchor=north west,rounded corners=3pt,draw=green!25!gray,fill=green!4,minimum height=44pt,text width=0.69\linewidth,font=\fontsize{8}{9.6}\selectfont] (p1) at (0.29,-56) {Harvey LAB~\cite{harveylab}; LawBench~\cite{lawbench}; Legal Research (vals)~\cite{valslegalresearch}; LegalBench~\cite{legalbench}};
\draw[gray!45] (0.035,-78) -- (g1.west);\draw[gray!45] (g1.east) -- (p1.west);
\node[anchor=north west,rounded corners=3pt,draw=green!25!gray,fill=green!4,minimum height=44pt,text width=0.21\linewidth] (g2) at (0.06,-102) {\textcolor{green!35!black}{\bfseries Healthcare \& medicine}\par{\fontsize{7}{8.5}\selectfont Clinical records, policies, biomedical data}};
\node[anchor=north west,rounded corners=3pt,draw=green!25!gray,fill=green!4,minimum height=44pt,text width=0.69\linewidth,font=\fontsize{8}{9.6}\selectfont] (p2) at (0.29,-102) {HealthAgentBench~\cite{healthagentbench}; MedAgentBench~\cite{medagentbench}; MedCode~\cite{medcode}; MedScribe~\cite{medscribe}; MLCR-AA~\cite{mlcraa}; PSEBench~\cite{psebench}};
\draw[gray!45] (0.035,-124) -- (g2.west);\draw[gray!45] (g2.east) -- (p2.west);
\node[anchor=north west,rounded corners=3pt,draw=cyan!30!gray,fill=cyan!4,minimum height=54pt,text width=0.21\linewidth] (g3) at (0.06,-148) {\textcolor{cyan!45!black}{\bfseries Office \& document work}\par{\fontsize{7}{8.5}\selectfont Document structure, layouts, office operations}};
\node[anchor=north west,rounded corners=3pt,draw=cyan!30!gray,fill=cyan!4,minimum height=54pt,text width=0.69\linewidth,font=\fontsize{8}{9.6}\selectfont] (p3) at (0.29,-148) {DocVQA~\cite{docvqa}; MMLongBench-Doc~\cite{mmlongbenchdoc}; OdysseyBench~\cite{odysseybench}; OfficeBench~\cite{officebench}; OfficeQA Pro~\cite{officeqapro}; OmegaUse-OfficeVal~\cite{omegauseofficeval}; PPTC~\cite{pptc}; PPTEval / PPTAgent~\cite{pptagent}; PresentBench~\cite{presentbench}; SpreadsheetBench~\cite{spreadsheetbench}};
\draw[gray!45] (0.035,-175) -- (g3.west);\draw[gray!45] (g3.east) -- (p3.west);
\node[anchor=north west,rounded corners=3pt,draw=cyan!30!gray,fill=cyan!4,minimum height=44pt,text width=0.21\linewidth] (g4) at (0.06,-204) {\textcolor{cyan!45!black}{\bfseries Customer service \& commerce}\par{\fontsize{7}{8.5}\selectfont Service policies, customer and order state}};
\node[anchor=north west,rounded corners=3pt,draw=cyan!30!gray,fill=cyan!4,minimum height=44pt,text width=0.69\linewidth,font=\fontsize{8}{9.6}\selectfont] (p4) at (0.29,-204) {CRMArena~\cite{crmarena}; CRMArena-Pro~\cite{crmapro}; E-Commerce Bench~\cite{ecommercebench}; $\tau^2$-bench~\cite{tau2bench}; $\tau$-bench~\cite{taubench}};
\draw[gray!45] (0.035,-226) -- (g4.west);\draw[gray!45] (g4.east) -- (p4.west);
\node[anchor=north west,rounded corners=3pt,draw=cyan!30!gray,fill=cyan!4,minimum height=54pt,text width=0.21\linewidth] (g5) at (0.06,-250) {\textcolor{cyan!45!black}{\bfseries Enterprise operations}\par{\fontsize{7}{8.5}\selectfont Business rules, workflows, shared records}};
\node[anchor=north west,rounded corners=3pt,draw=cyan!30!gray,fill=cyan!4,minimum height=54pt,text width=0.69\linewidth,font=\fontsize{8}{9.6}\selectfont] (p5) at (0.29,-250) {AutomationBench~\cite{automationbench}; CI-Work~\cite{ciwork}; EnterpriseBench~\cite{enterprisebench}; EnterpriseOps-Gym~\cite{enterpriseops}; FieldWorkArena~\cite{fieldworkarena}; TheAgentCompany~\cite{agentcompany}; WorkArena~\cite{workarena}; WorkArena++~\cite{workarenapp}; WorkBench~\cite{workbench}};
\draw[gray!45] (0.035,-277) -- (g5.west);\draw[gray!45] (g5.east) -- (p5.west);
\node[anchor=north west,rounded corners=3pt,draw=blue!25!gray,fill=blue!3,minimum height=44pt,text width=0.21\linewidth] (g6) at (0.06,-306) {\textcolor{blue!40!black}{\bfseries Software engineering}\par{\fontsize{7}{8.5}\selectfont Code, tests, dependencies, execution}};
\node[anchor=north west,rounded corners=3pt,draw=blue!25!gray,fill=blue!3,minimum height=44pt,text width=0.69\linewidth,font=\fontsize{8}{9.6}\selectfont] (p6) at (0.29,-306) {SWE-bench~\cite{swebench}; Terminal-Bench 2.0~\cite{terminalbench}};
\draw[gray!45] (0.035,-328) -- (g6.west);\draw[gray!45] (g6.east) -- (p6.west);
\node[anchor=north west,rounded corners=3pt,draw=blue!25!gray,fill=blue!3,minimum height=44pt,text width=0.21\linewidth] (g7) at (0.06,-352) {\textcolor{blue!40!black}{\bfseries Data science \& data engineering}\par{\fontsize{7}{8.5}\selectfont Data semantics, analysis, query workflows}};
\node[anchor=north west,rounded corners=3pt,draw=blue!25!gray,fill=blue!3,minimum height=44pt,text width=0.69\linewidth,font=\fontsize{8}{9.6}\selectfont] (p7) at (0.29,-352) {DSBench~\cite{dsbench}; Spider 2.0 (agentic)~\cite{spider2}};
\draw[gray!45] (0.035,-374) -- (g7.west);\draw[gray!45] (g7.east) -- (p7.west);
\node[anchor=north west,rounded corners=3pt,draw=blue!25!gray,fill=blue!3,minimum height=44pt,text width=0.21\linewidth] (g8) at (0.06,-398) {\textcolor{blue!40!black}{\bfseries Scientific research}\par{\fontsize{7}{8.5}\selectfont Research methods and scientific data}};
\node[anchor=north west,rounded corners=3pt,draw=blue!25!gray,fill=blue!3,minimum height=44pt,text width=0.69\linewidth,font=\fontsize{8}{9.6}\selectfont] (p8) at (0.29,-398) {DiscoveryBench~\cite{discoverybench}; ScienceAgentBench~\cite{scienceagentbench}};
\draw[gray!45] (0.035,-420) -- (g8.west);\draw[gray!45] (g8.east) -- (p8.west);
\node[anchor=north west,rounded corners=3pt,draw=blue!25!gray,fill=blue!3,minimum height=44pt,text width=0.21\linewidth] (g9) at (0.06,-444) {\textcolor{blue!40!black}{\bfseries Computer-aided design \& 3D graphics}\par{\fontsize{7}{8.5}\selectfont Geometry, CAD programs, rendered scenes}};
\node[anchor=north west,rounded corners=3pt,draw=blue!25!gray,fill=blue!3,minimum height=44pt,text width=0.69\linewidth,font=\fontsize{8}{9.6}\selectfont] (p9) at (0.29,-444) {CADBench (BlenderLLM)~\cite{du2024blenderllmtraininglargelanguage}; CADBench (MIT)~\cite{doris2026cadbench}; BlenderBench (VIGA)~\cite{yin2026visionasinversegraphics}; BenchCAD~\cite{benchcad}};
\draw[gray!45] (0.035,-466) -- (g9.west);\draw[gray!45] (g9.east) -- (p9.west);
\node[anchor=north west,rounded corners=3pt,draw=violet!25!gray,fill=violet!4,minimum height=64pt,text width=0.21\linewidth] (g10) at (0.06,-490) {\textcolor{violet!45!black}{\bfseries General digital interaction}\par{\fontsize{7}{8.5}\selectfont Interfaces, tool contracts, application state}};
\node[anchor=north west,rounded corners=3pt,draw=violet!25!gray,fill=violet!4,minimum height=64pt,text width=0.69\linewidth,font=\fontsize{8}{9.6}\selectfont] (p10) at (0.29,-490) {AgentIF-OneDay~\cite{agentifoneday}; AndroidWorld~\cite{androidworld}; AppWorld~\cite{appworld}; MCP-Atlas~\cite{mcpatlas}; MMTB~\cite{mmtb}; MultiAgentBench~\cite{multiagentbench}; Odysseys~\cite{odysseys}; OSWorld~\cite{osworld}; OSWorld 2.0~\cite{yuan2026osworld2}; LOCA-bench (up to 256K)~\cite{zeng2026loca}; ToolSandbox~\cite{toolsandbox}; VisualWebArena~\cite{visualwebarena}; WebArena~\cite{webarena}; WildClawBench~\cite{wildclawbench}; Windows Agent Arena~\cite{windowsarena}};
\draw[gray!45] (0.035,-522) -- (g10.west);\draw[gray!45] (g10.east) -- (p10.west);
\node[anchor=north west,rounded corners=3pt,draw=violet!25!gray,fill=violet!4,minimum height=64pt,text width=0.21\linewidth] (g11) at (0.06,-556) {\textcolor{violet!45!black}{\bfseries Cross-domain professional work}\par{\fontsize{7}{8.5}\selectfont Knowledge varies across roles and tasks}};
\node[anchor=north west,rounded corners=3pt,draw=violet!25!gray,fill=violet!4,minimum height=64pt,text width=0.69\linewidth,font=\fontsize{8}{9.6}\selectfont] (p11) at (0.29,-556) {\$OneMillion-Bench~\cite{onemillionbench}; APEX-Agents~\cite{apex}; Agents' Last Exam~\cite{sun2026agentslastexam}; CentaurBench~\cite{centaurbench}; CL-Bench~\cite{clbench}; GDP.pdf~\cite{gdppdf}; GDPval~\cite{gdpval}; JOBBench~\cite{jobbench}; OmniaBench~\cite{omniabench}; PRBench~\cite{prbench}; Remote Labor Index~\cite{rli}; Workspace-Bench~\cite{workspacebench}; StartupBench~\cite{startupbench}};
\draw[gray!45] (0.035,-588) -- (g11.west);\draw[gray!45] (g11.east) -- (p11.west);
\draw[gray!45] (0.035,-27) -- (0.035,-588);
\end{tikzpicture}%
}
\caption{Work domains, settings, and representative knowledge requirements across 79 benchmarks. Studies are grouped by their primary domain or work setting, following the groups of Table~\ref{tab:benchmark-metrics-judges}. Each study appears once for readability; the groups are non-exclusive and connecting lines indicate membership, not a temporal sequence.}\label{fig:benchmark-domain-map}
\end{figure}

\subsection{Evaluation Methodologies}\label{evaluation-methods}

Evaluating a work agent involves three decisions: which result to inspect, which properties to judge it on, and how to form the judgment. We refer to these as the evaluation target, the evaluation dimensions, and the evaluation method. Targets are the actual answers, artifacts, or environment states; dimensions include correctness, completeness, structure, visual quality, and constraint compliance. Section~\ref{evaluation-evidence} covers targets and dimensions, Section~\ref{judging-procedures} covers methods, and Section~\ref{metrics} discusses how judgments are aggregated into scores.

\subsubsection{Evaluation Targets and Evaluation Dimensions}\label{evaluation-evidence}

By the result inspected at final acceptance, benchmarks fall into four levels: answer, artifact, environment state, and workflow. A task may yield several kinds of evidence; its level is set by the object on which final acceptance is decided.

\textbf{Answer-level evaluation.} The result is text, a number, a label, or a structured response, checked for correctness, completeness, and consistency with reference information. ConvFinQA executes the model's reasoning program for a chain of interdependent questions about a financial report and compares the result with the gold answer~\cite{convfinqa}; MMLongBench-Doc asks questions over PDFs averaging 47.5 pages, some of which the document cannot answer~\cite{mmlongbenchdoc}. CRMArena has agents query a Salesforce organization but scores only the submitted answer~\cite{crmarena}. Answer-level evaluation suits well-defined outputs, but a correct answer does not establish that the operations behind it, such as CRMArena's queries, were correct.

\textbf{Artifact-level evaluation.} The result is a deliverable the agent created or modified, such as code, a document, or a model file. ScienceAgentBench runs each generated program and checks that it executes and that its output meets task-specific success criteria~\cite{scienceagentbench}; OmegaUse-OfficeVal requires Office deliverables to be valid, openable, and editable before scoring them against weighted requirements, with penalties for unintended damage~\cite{omegauseofficeval}. Because each artifact type has its own requirements of use, text similarity cannot replace structural and functional checks.

\textbf{Environment-state-level evaluation.} The result is the state of a website, application, or business system after execution. Terminal-Bench 2.0 tests the final container state rather than the agent's commands~\cite{terminalbench}, VisualWebArena checks the final webpage state with model-based checks where visual content matters~\cite{visualwebarena}, and WorkBench compares five sandbox databases with a ground-truth outcome while counting mismatching side effects~\cite{workbench}. Checking state rather than actions admits different action paths to the same goal, but the check must also catch unintended modifications and violated constraints.

\textbf{Workflow-level evaluation.} The result is a multi-step, cross-application, or multi-file delivery, and evaluation covers the final result, intermediate requirements, and dependencies among objects. ToolSandbox matches the trajectory against milestones that must occur and minefields that must not, scoring zero if any minefield is hit~\cite{toolsandbox}. MultiAgentBench uses an LLM-based detector to record which milestones a team reaches and who contributes~\cite{multiagentbench}, and Odysseys has an LLM judge assess the full trajectory against each rubric item~\cite{odysseys}. This level comes closest to a complete delegation, but it must make clear which process constraints belong to the acceptance conditions.

\FloatBarrier
\subsubsection{Evaluation Methods}\label{judging-procedures}

Given the collected evidence, automated benchmarks judge it mainly through rules, LLM judges, or agent judges; occupational experts assess open-ended work products or provide reference judgments.

\textbf{Rule-based evaluation.} Programs verify the result by answer matching, execution, or assertions over files and state. AutomationBench checks the end state of simulated business applications with deterministic assertions, including changes that must not occur, and gives no partial credit~\cite{automationbench}; OfficeBench saves the file system after each task and applies exact, fuzzy, or execution-based checks~\cite{officebench}. Such checks are clear and repeatable but establish only the requirements their tests cover.

\textbf{LLM-as-a-Judge.} A model judges the result against task requirements, reference material, and a rubric, which suits semantic, professional, and visual qualities that resist formalization~\cite{geval,prometheus}. MCP-Atlas has an LLM check each required factual claim in the final answer and passes a task at 0.75 claim coverage~\cite{mcpatlas}; Finance Agent v2 uses a jury of three frontier models~\cite{financeagentv2}; and CI-Work judges, entry by entry, whether an agent's final action conveys required information or discloses sensitive information~\cite{ciwork}. These scores depend on the criteria, the reference evidence, and the judge model, and are not objective fact.

\textbf{Agent-as-a-Judge.} An evaluating agent gathers evidence itself by reading files, calling tools, or inspecting traces~\cite{agentjudge}. Workspace-Bench lets its judge inspect output files, inputs, rubrics, and the execution trajectory with shell tools and file parsers~\cite{workspacebench}; BankerToolBench gives its judge the same tools as the evaluated agent, so it can run code to check spreadsheet formulas~\cite{bankertoolbench}; and StartupBench runs a separate judge session per rubric item, which agreed more often with domain experts than a single holistic judge~\cite{startupbench}. Agent judges reach evidence scattered across files, but extra tool execution can cause missed checks or unstable judgments.

These methods are often combined: WildClawBench pairs deterministic checks and side-effect audits with an LLM or VLM judge for reports and images~\cite{wildclawbench}, and CRMArena-Pro scores answers by exact match while a GPT-4o judge assesses confidentiality awareness~\cite{crmapro}. For open-ended professional work, expert judgment remains the reference; in Remote Labor Index, trained evaluators and subject experts decide, as a reasonable client would, whether an AI deliverable completes the project at least as well as the human one~\cite{rli}. Table~\ref{tab:benchmark-metrics-judges} links the targets, dimensions, methods, and metrics of each benchmark.

\begingroup
\footnotesize
\setlength{\tabcolsep}{3pt}
\renewcommand{\arraystretch}{1.12}
\begingroup
\fontsize{7.4}{9.2}\selectfont
\setlength{\tabcolsep}{2.5pt}
\renewcommand{\arraystretch}{1.12}
\arrayrulecolor{tableRule}
\Needspace{150pt}

\begin{longtable}{@{}>{\raggedright\arraybackslash}m{86.000pt}
 >{\raggedright\arraybackslash}m{45.000pt}
 >{\raggedright\arraybackslash}m{68.000pt}
 >{\raggedright\arraybackslash}m{82.000pt}
 >{\raggedright\arraybackslash}m{38.000pt}
 >{\raggedright\arraybackslash}m{128.000pt}@{}}
\caption{Evaluation targets, evaluation dimensions, evaluation methods, and model performance metrics in work-agent benchmarks. Metrics summarize only the requirements covered by the stated checks or rubrics.}\label{tab:benchmark-metrics-judges}\\
\toprule
\rowcolor{tableHeader}
\textcolor{tableInk}{\bfseries Benchmark} & \textcolor{tableInk}{\bfseries Level} & \textcolor{tableInk}{\bfseries Inspected work result} & \textcolor{tableInk}{\bfseries Main evaluation dimensions} & \textcolor{tableInk}{\bfseries Method} & \textcolor{tableInk}{\bfseries Model performance metrics} \\
\midrule
\endfirsthead
\multicolumn{6}{@{}l@{}}{\small\tablename~\thetable~(continued)}\\
\toprule
\rowcolor{tableHeader}
\textcolor{tableInk}{\bfseries Benchmark} & \textcolor{tableInk}{\bfseries Level} & \textcolor{tableInk}{\bfseries Inspected work result} & \textcolor{tableInk}{\bfseries Main evaluation dimensions} & \textcolor{tableInk}{\bfseries Method} & \textcolor{tableInk}{\bfseries Model performance metrics} \\
\midrule
\endhead
\midrule
\multicolumn{6}{r@{}}{\color{tableMuted}\scriptsize Continued on next page}\\
\endfoot
\bottomrule
\endlastfoot
\rowcolor{tableHeader}\multicolumn{6}{@{}l@{}}{\textcolor{tableInk}{\bfseries Finance \& accounting}}\\*
\MethodEntry{BankerToolBench}{bankertoolbench} & Artifact & financial models, presentations, and professional memoranda & Financial correctness, professional requirements, delivery quality & Agent & Weighted Rubric Score \\
\MethodEntry{ConvFinQA}{convfinqa} & Answer & multi-turn financial question answering & Numerical correctness, contextual reasoning & Rule & Execution Accuracy \\
\MethodEntry{Finance Agent Benchmark}{financeagentbenchmark} & Answer & financial retrieval and research answers & Conclusion correctness, evidence support & Rule & Accuracy \\
\MethodEntry{Finance Agent v2}{financeagentv2} & Answer & financial research results & Answer correctness, weighted checks with dealbreakers & LLM & Partial Credit, All-Pass \\
\MethodEntry{FinQA}{finqa} & Answer & numerical answer and reasoning program & Computational correctness, program correctness & Rule & Execution Accuracy, Program Accuracy \\
\MethodEntry{FrontierFinance}{zhang2026frontierfinance} & Answer & long-form investment research answers & Source-attributed factual accuracy, analytical completeness & LLM & Macro-averaged Rubric Qualification Rate (all and must-have rubrics), Cost, Latency \\
\MethodEntry{TAT-QA}{tatqa} & Answer & table and text question answering & Answer correctness, textual and numerical matching & Rule & Exact Match, F1 \\
\specialrule{0.5pt}{1pt}{0pt}
\rowcolor{tableHeader}\multicolumn{6}{@{}l@{}}{\textcolor{tableInk}{\bfseries Law \& legal services}}\\*
\MethodEntry{Harvey LAB}{harveylab} & Artifact & legal documents and professional deliverables & Legal professional requirements, content completeness & LLM & Task Pass Rate, Criteria Pass Rate \\
\MethodEntry{LawBench}{lawbench} & Answer & legal question answering and judgment & Legal understanding, reasoning, and text-processing correctness & Rule & Accuracy, F1, ROUGE-L, and other task metrics \\
\MethodEntry{Legal Research (vals)}{valslegalresearch} & Artifact & legal research results & Research correctness, professional requirement satisfaction & LLM & All-Pass Accuracy, Weighted Partial Credit \\
\MethodEntry{LegalBench}{legalbench} & Answer & legal labels and answers & Correctness of legal judgment & Rule & Accuracy, Balanced Accuracy \\
\specialrule{0.5pt}{1pt}{0pt}
\rowcolor{tableHeader}\multicolumn{6}{@{}l@{}}{\textcolor{tableInk}{\bfseries Healthcare \& medicine}}\\*
\MethodEntry{HealthAgentBench}{healthagentbench} & Workflow & medical analysis and tool execution & Task completion, correctness of medical results & Rule + LLM & Task Success Rate \\
\MethodEntry{MedAgentBench}{medagentbench} & Environment state & FHIR medical records & Query correctness, record-operation correctness & Rule & Task Success Rate \\
\MethodEntry{MedCode}{medcode} & Answer & clinical coding & Coding accuracy, agreement with reference codes & Rule & Coding Accuracy \\
\MethodEntry{MedScribe}{medscribe} & Artifact & clinical notes and SOAP documents & Clinical information completeness, documentation quality & LLM & Rubric Score \\
\MethodEntry{MLCR-AA}{mlcraa} & Answer & long medical record question answering & Completeness, no contradiction with the source record & LLM & Pass Rate, Completeness, Accuracy \\
\MethodEntry{PSEBench}{psebench} & Workflow & patient-safety event triage & Triage correctness, clause identification, evidence grounding & Rule + LLM & Verdict Accuracy, Clause Accuracy, Evidence Citation F1 \\
\specialrule{0.5pt}{1pt}{0pt}
\rowcolor{tableHeader}\multicolumn{6}{@{}l@{}}{\textcolor{tableInk}{\bfseries Office \& document work}}\\*
\MethodEntry{DocVQA}{docvqa} & Answer & document image question answering & Answer correctness & Rule & ANLS \\
\MethodEntry{MMLongBench-Doc}{mmlongbenchdoc} & Answer & long-document question answering & Long-document understanding, answer correctness & LLM + Rule & Overall Accuracy, Overall F1 \\
\MethodEntry{OdysseyBench}{odysseybench} & Workflow & cross-application office tasks & Task completion, correctness of file and application state & Rule & Pass Rate \\
\MethodEntry{OfficeBench}{officebench} & Workflow & cross-application Office operations & Task completion, state and file correctness & Rule & Pass Rate \\
\MethodEntry{OfficeQA Pro}{officeqapro} & Answer & professional PDF question answering & Answer correctness & Rule & Exact Match Accuracy \\
\MethodEntry{OmegaUse-OfficeVal}{omegauseofficeval} & Artifact & Office file deliverables & Content correctness, structure, and delivery requirements & Rule & Score, Time-Weighted Score, Price-Weighted Score \\
\MethodEntry{PPTC}{pptc} & Artifact & native PPTX & Instruction satisfaction, object attributes and positional relations & Rule & Instruction Accuracy, Session Accuracy \\
\MethodEntry{PPTEval / PPTAgent}{pptagent} & Artifact & presentations & Content quality, design quality, coherence & LLM & Content, Design, and Coherence Scores \\
\MethodEntry{PresentBench}{presentbench} & Artifact & presentations & Content quality, visual presentation & LLM & Overall Score, Dimension Scores \\
\MethodEntry{SpreadsheetBench}{spreadsheetbench} & Artifact & Excel workbooks & Computational behavior, correctness of target results & Rule & Execution Accuracy \\
\specialrule{0.5pt}{1pt}{0pt}
\rowcolor{tableHeader}\multicolumn{6}{@{}l@{}}{\textcolor{tableInk}{\bfseries Customer service \& commerce}}\\*
\MethodEntry{CRMArena}{crmarena} & Answer & answers to CRM queries & Answer correctness & Rule & Exact Match, F1, pass$^k$ \\
\MethodEntry{CRMArena-Pro}{crmapro} & Answer & answers in single- and multi-turn CRM tasks & Answer correctness, confidentiality awareness & Rule + LLM & Task Completion, Confidentiality Awareness \\
\MethodEntry{E-Commerce Bench}{ecommercebench} & Workflow & long-horizon e-commerce operation & Operating results, business decisions, execution performance & Rule & Year-end Total Assets \\
\MethodEntry{$\tau^2$-bench}{tau2bench} & Workflow & customer service and device operation & Goal completion, policy compliance, repeat stability & Rule & Task Success, pass$^k$ \\
\MethodEntry{$\tau$-bench}{taubench} & Workflow & customer service and database state & Target state, policy compliance, repeat stability & Rule & Task Success, pass$^k$ \\
\specialrule{0.5pt}{1pt}{0pt}
\rowcolor{tableHeader}\multicolumn{6}{@{}l@{}}{\textcolor{tableInk}{\bfseries Enterprise operations}}\\*
\MethodEntry{AutomationBench}{automationbench} & Workflow & SaaS automation tasks & Goal completion, constraint compliance & Rule & Task Success Rate \\
\MethodEntry{CI-Work}{ciwork} & Answer & agent's final information-sharing action & Conveyance of essential information, leakage of sensitive information & LLM & Conveyance Rate, Leakage Rate, Violation Rate \\
\MethodEntry{EnterpriseBench}{enterprisebench} & Workflow & enterprise question answering and interactive decisions & Information correctness, decision results, task completion & Rule + LLM & Task accuracy and interactive-decision scores \\
\MethodEntry{EnterpriseOps-Gym}{enterpriseops} & Environment state & enterprise databases and service state & Target state, integrity constraints, side effects & Rule & Task Success Rate \\
\MethodEntry{FieldWorkArena}{fieldworkarena} & Answer & judgments over field images and video & Answer correctness, numerical accuracy & LLM + Rule & Accuracy Rate (Correctness and Numerical Scores) \\
\MethodEntry{TheAgentCompany}{agentcompany} & Workflow & simulated enterprise collaboration tasks & Intermediate objectives, final completion & Rule + LLM & Task Completion Rate, Checkpoint Score \\
\MethodEntry{WorkArena}{workarena} & Environment state & ServiceNow application & Target state, task completion & Rule & Task Success Rate \\
\MethodEntry{WorkArena++}{workarenapp} & Workflow & complex ServiceNow tasks & Multi-step task completion, state correctness & Rule & Task Success Rate \\
\MethodEntry{WorkBench}{workbench} & Environment state & sandbox database state & Target state, side effects & Rule & Accuracy, Side-Effect Rate \\
\specialrule{0.5pt}{1pt}{0pt}
\rowcolor{tableHeader}\multicolumn{6}{@{}l@{}}{\textcolor{tableInk}{\bfseries Software engineering}}\\*
\MethodEntry{SWE-bench}{swebench} & Artifact & code patch and repository & Defect-repair correctness, regression behavior & Rule & \% Resolved / Issue Resolution Rate \\
\MethodEntryRes{Terminal-Bench 2.0}{terminalbench}{terminalbench2} & Environment state & final container state & Task completion, execution correctness & Rule & Resolution Rate \\
\specialrule{0.5pt}{1pt}{0pt}
\rowcolor{tableHeader}\multicolumn{6}{@{}l@{}}{\textcolor{tableInk}{\bfseries Data science \& data engineering}}\\*
\MethodEntry{DSBench}{dsbench} & Answer and artifact & data-analysis answers and prediction files & Analytical correctness, predictive performance & Rule + LLM & Analysis Accuracy, Relative Performance Gap \\
\MethodEntry{Spider 2.0 (agentic)}{spider2} & Workflow & database engineering tasks & Query execution, correctness of workflow results & Rule & Success Rate \\
\specialrule{0.5pt}{1pt}{0pt}
\rowcolor{tableHeader}\multicolumn{6}{@{}l@{}}{\textcolor{tableInk}{\bfseries Scientific research}}\\*
\MethodEntry{DiscoveryBench}{discoverybench} & Answer & scientific hypotheses and analytical conclusions & Correctness of scientific discovery, hypothesis matching & LLM & Hypothesis Matching Score \\
\MethodEntry{ScienceAgentBench}{scienceagentbench} & Artifact & scientific programs and computed results & Scientific task completion, code executability & Rule & Success Rate, Valid Execution Rate \\
\specialrule{0.5pt}{1pt}{0pt}
\rowcolor{tableHeader}\multicolumn{6}{@{}l@{}}{\textcolor{tableInk}{\bfseries Computer-aided design \& 3D graphics}}\\*
\MethodEntry{CADBench (BlenderLLM)}{du2024blenderllmtraininglargelanguage} & Artifact & Blender scripts and 3D models & Object attributes, spatial relations, instruction following, script validity & LLM + Rule & Attribute Score, Spatial Score, Instruction Score, Average Score, Syntax Error Rate \\
\MethodEntry{CADBench (MIT)}{doris2026cadbench} & Artifact & CAD programs and generated geometry & Geometric fidelity, executability, program compactness & Rule & IoU, SIoU, Chamfer Distance, Valid Shape Rate, Token Count, Operation Count \\
\MethodEntry{BlenderBench (VIGA)}{yin2026visionasinversegraphics} & Artifact & Blender scenes and rendered results & Rendering consistency, spatial relations, visual quality, task completion & Rule + LLM & Photometric Loss, N-CLIP Score, VLM Score \\
\MethodEntry{BenchCAD}{benchcad} & Artifact & CadQuery programs and edited solids & Geometric fidelity, essential-operation use, executability, edit fidelity & Rule & Execution Rate, Voxel IoU, Chamfer and Hausdorff Distance, Feature-F1, Essential-Op Recall, Normalized Edit Accuracy \\
\specialrule{0.5pt}{1pt}{0pt}
\rowcolor{tableHeader}\multicolumn{6}{@{}l@{}}{\textcolor{tableInk}{\bfseries General digital interaction}}\\*
\MethodEntry{AgentIF-OneDay}{agentifoneday} & Artifact & general task deliverables & Instruction following, degree of requirement satisfaction & LLM & Average Normalized Rubric Score \\
\MethodEntry{AndroidWorld}{androidworld} & Environment state & Android applications & Target state, task completion & Rule & Task Success Rate \\
\MethodEntry{AppWorld}{appworld} & Environment state & multi-application databases & Goal completion, preservation of non-target state & Rule & Task Goal Completion, Scenario Goal Completion \\
\MethodEntry{MCP-Atlas}{mcpatlas} & Answer & final answers to cross-server tool tasks & Claim correctness, coverage of necessary facts & LLM & Pass Rate, Mean Coverage \\
\MethodEntry{MMTB}{mmtb} & Workflow & multi-type tool calling & Tool selection, arguments, and interaction correctness & Rule & Action-type / Layer Accuracy, Multi-turn Accuracy \\
\MethodEntry{MultiAgentBench}{multiagentbench} & Workflow & multi-agent collaboration & Task milestones, task outcome, coordination quality & LLM + Rule & Task Score, Milestone KPI, Coordination Score \\
\MethodEntry{Odysseys}{odysseys} & Workflow & long-horizon web and file tasks & Item-wise requirement satisfaction, complete completion & LLM & Rubric Average, Perfect Rate \\
\MethodEntry{OSWorld}{osworld} & Environment state & desktop and file state & Target state, task completion & Rule & Task Success Rate \\
\MethodEntry{OSWorld 2.0}{yuan2026osworld2} & Workflow & cross-application desktop tasks with final files and state & Long-horizon completion, target-state correctness, cross-step constraints & Rule & Binary Task Completion, Partial Score \\
\MethodEntry{LOCA-bench (up to 256K)}{zeng2026loca} & Workflow & long-context tool tasks and environment state & Tool-operation correctness, context use, task completion & Rule & Task Accuracy (grouped by context length), Average Accuracy \\
\MethodEntry{ToolSandbox}{toolsandbox} & Workflow & tool calls and state milestones & Task completion, state correctness, constraint compliance & Rule & Milestone-based Task Score \\
\MethodEntry{VisualWebArena}{visualwebarena} & Environment state & visual website state and answers & Goal completion, correctness of visual interaction & Rule + LLM & Task Success Rate \\
\MethodEntry{WebArena}{webarena} & Environment state & website state and answers & Goal completion, answer correctness & Rule & Task Success Rate \\
\MethodEntry{WildClawBench}{wildclawbench} & Workflow & CLI and multimodal tool tasks & Execution correctness, state and constraint compliance & Rule + LLM & Overall Score \\
\MethodEntry{Windows Agent Arena}{windowsarena} & Environment state & Windows applications and files & Target state, task completion & Rule & Task Success Rate \\
\specialrule{0.5pt}{1pt}{0pt}
\rowcolor{tableHeader}\multicolumn{6}{@{}l@{}}{\textcolor{tableInk}{\bfseries Cross-domain professional work}}\\*
\MethodEntry{\$OneMillion-Bench}{onemillionbench} & Answer & expert-level text responses & Multidimensional professional quality, requirement satisfaction & LLM & Expert Score, Pass Rate, Economic Value \\
\MethodEntry{APEX-Agents}{apex} & Workflow & multi-application professional tasks & Task completion, satisfaction of itemized requirements & LLM & Pass@1, Pass@8, Pass$^8$, Mean Score \\
\MethodEntry{Agents' Last Exam}{sun2026agentslastexam} & Workflow & occupational software tasks and deliverables & Complete task completion, professional acceptance, degree of partial completion & Rule & Pass Rate, Score \\
\MethodEntry{CentaurBench}{centaurbench} & Artifact & deliverables produced directly or by a model-guided worker model & Relative work quality, quality of guidance to a weaker model & LLM & Mean Rank, Automation Rank, Augmentation Rank \\
\MethodEntry{CL-Bench}{clbench} & Answer & context-learning tasks & Context understanding, rule compliance, task completion & LLM & Solving Rate \\
\MethodEntry{GDP.pdf}{gdppdf} & Answer & professional PDF tasks & Document understanding, degree of requirement satisfaction & LLM & Strict All-Pass Rate, Graded Rubric Score \\
\MethodEntry{GDPval}{gdpval} & Artifact & occupational professional deliverables & Professional quality, relative acceptability & Expert assessment & Win Rate, Win-or-Tie Rate \\
\MethodEntry{JOBBench}{jobbench} & Artifact & deliverables built from multi-file workspaces & Satisfaction of chained requirements, delivery completeness & LLM & Rubric Score \\
\MethodEntry{OmniaBench}{omniabench} & Workflow & multi-type executable tasks & Goal completion, program and state correctness & Rule + LLM & Overall Pass@1, Route-level Scores \\
\MethodEntry{PRBench}{prbench} & Answer & professional task responses & Professional correctness, completeness, criterion satisfaction & LLM & Weighted Rubric Score, Min-Normalized Score \\
\MethodEntry{Remote Labor Index}{rli} & Artifact & end-to-end remote work products & Professional acceptance, degree of automated completion & Expert assessment & Automation Rate \\
\MethodEntry{Workspace-Bench}{workspacebench} & Workflow & multi-file workspaces & Requirement satisfaction, task completion, correctness of dependency relations & Agent + Rule & Rubric Pass Rate, Task Completion Rate, Node F1, Edge F1 \\
\MethodEntry{StartupBench}{startupbench} & Artifact & professional workspaces and complete delivery tasks & Importance-tiered requirement satisfaction, delivery completeness & Agent & Importance-Weighted Mean Task Score, Success Rate (score $\geq$ 90) \\
\end{longtable}
\endgroup
\par\endgroup

\FloatBarrier

\subsection{Evaluation Metrics}\label{metrics}\label{outcome-quality-metrics}

Evaluation methods judge each task; performance metrics aggregate these judgments across tasks or samples. Benchmarks use three kinds: task outcome scores, relative comparison metrics, and weighted performance metrics. All measure result quality or degree of completion; resource metrics such as cost, tokens, and latency are discussed in Section~\ref{performance-cost}.

\textbf{Task outcome scores.} These measure whether, or how far, a result meets the acceptance conditions. The most common is the share of successful tasks; in $\tau$-bench, a task succeeds only if the final database matches the ground truth and the agent conveys the required information~\cite{taubench}. Answer-oriented benchmarks use accuracy and matching scores, such as accuracy, F1, or ROUGE-L in LawBench, depending on the task~\cite{lawbench}. Rubric scores differ in strictness: GDP.pdf ranks models by a strict pass rate that requires every criterion, alongside a mean rubric score~\cite{gdppdf}. Some outcomes are continuous, such as year-end total assets after a simulated year of store operation in E-Commerce Bench~\cite{ecommercebench}. Over repeated attempts, pass@$k$ requires at least one of $k$ attempts to succeed, and pass$^k$, introduced by $\tau$-bench, requires all of them~\cite{taubench}; APEX-Agents reports both Pass@8 and Pass$^8$~\cite{apex}. Because they differ in unit and strictness, these scores are not directly comparable across benchmarks.

\textbf{Relative comparison metrics.} Without a unique reference answer, quality can be measured by preferences, win rates, or rankings among candidates. GDPval has occupational experts blindly compare model and professional deliverables, reporting how often the model's is rated better or as good~\cite{gdpval}. CentaurBench turns pairwise judgments from a panel of LLM judges into rankings, separately for automation, where the model produces the deliverable, and augmentation, where it guides a weaker worker model~\cite{centaurbench}. Such results depend on what is compared and how, and cannot be read apart from the reference set.

\textbf{Weighted performance metrics.} These aggregate results with explicit weights on criteria or on tasks. PRBench weights each expert criterion with an integer from $-10$ to $10$, so that undesirable content lowers the score~\cite{prbench}; OmegaUse-OfficeVal weights tasks by recorded human labor time or a task-price proxy~\cite{omegauseofficeval}. Criterion weights reflect how much each requirement matters and task weights how much each task is worth; both remain performance scores, not measures of economic return.

\begingroup
\footnotesize
\setlength{\tabcolsep}{3pt}
\renewcommand{\arraystretch}{1.15}
\begingroup
\fontsize{7.4}{9.2}\selectfont
\setlength{\tabcolsep}{2.5pt}
\renewcommand{\arraystretch}{1.12}
\arrayrulecolor{tableRule}
\Needspace{150pt}

\begin{longtable}{@{}>{\raggedright\arraybackslash}m{(\linewidth - 6\tabcolsep) * \real{0.19000}}
 >{\raggedright\arraybackslash}m{(\linewidth - 6\tabcolsep) * \real{0.35000}}
 >{\raggedright\arraybackslash}m{(\linewidth - 6\tabcolsep) * \real{0.19000}}
 >{\raggedright\arraybackslash}m{(\linewidth - 6\tabcolsep) * \real{0.27000}}@{}}
\caption{Categories of model performance metrics in work-agent benchmarks, with their units, aggregation, and required qualifications \label{tab:working-agents-44}}\\
\toprule
\rowcolor{tableHeader}
\textcolor{tableInk}{\bfseries Metric category} & \textcolor{tableInk}{\bfseries Specific metrics, units, and aggregation} & \textcolor{tableInk}{\bfseries What it measures} & \textcolor{tableInk}{\bfseries Required qualification} \\
\midrule
\endfirsthead
\multicolumn{4}{@{}l@{}}{\small\tablename~\thetable~(continued)}\\
\toprule
\rowcolor{tableHeader}
\textcolor{tableInk}{\bfseries Metric category} & \textcolor{tableInk}{\bfseries Specific metrics, units, and aggregation} & \textcolor{tableInk}{\bfseries What it measures} & \textcolor{tableInk}{\bfseries Required qualification} \\
\midrule
\endhead
\midrule
\multicolumn{4}{r@{}}{\color{tableMuted}\scriptsize Continued on next page}\\
\endfoot
\bottomrule
\endlastfoot
\MethodName{Task Outcome Score} & \textit{Success rates:} Task Success Rate, pass@$k$, pass$^k$, Pass Rate, and similar measures counting the proportion of successes or the success probability under repeated execution. \textit{Accuracy and matching:} Accuracy, Exact Match, Execution Accuracy, F1, ANLS, IoU, Chamfer Distance, and similar measures computed against reference results. \textit{Quality and partial credit:} Rubric Score, Overall Score, Partial Score, and similar measures aggregating item-level requirements or partial completion. \textit{Continuous outcomes:} final operating or work output counted according to the task definition & How far the model satisfies the task objective, the acceptance requirements, and the delivery quality standard & Acceptance conditions, matching units, and partial-credit rules must be stated. Representative benchmarks include \par\nobreak\MethodEntry{SWE-bench}{swebench} \par\nobreak\MethodEntry{FinQA}{finqa} \par\nobreak\MethodEntry{DocVQA}{docvqa} \par\nobreak\MethodEntry{PresentBench}{presentbench} \par\nobreak\MethodEntry{Workspace-Bench}{workspacebench} \par\nobreak\MethodEntry{OSWorld 2.0}{yuan2026osworld2} \par\nobreak\MethodEntry{Agents' Last Exam}{sun2026agentslastexam} \\
\MethodName{Relative Comparison Metrics} & Pairwise Preference, Win Rate, Tie Rate, Mean Rank, and similar measures aggregated through pairwise comparison, preference statistics, or model ranking & Model quality relative to a designated reference work product or to other models & Comparison objects, judging protocol, and ranking rules must be stated. Representative benchmarks include \par\nobreak\MethodEntry{GDPval}{gdpval} \par\nobreak\MethodEntry{CentaurBench}{centaurbench} \\
\MethodName{Weighted Performance Metrics} & Importance-Weighted Score, Time-Weighted Score, Price-Weighted Score, and similar measures that aggregate outcome scores using the importance of within-task requirements or explicit weights across tasks & Composite model performance after accounting for requirement importance or differences in task value & The source of the weights, the level at which they act, and the normalization procedure must be stated. Representative benchmarks include \par\nobreak\MethodEntry{BankerToolBench}{bankertoolbench} \par\nobreak\MethodEntry{StartupBench}{startupbench} \par\nobreak\MethodEntry{OmegaUse-OfficeVal}{omegauseofficeval} \\
\end{longtable}
\endgroup
\endgroup

Table~\ref{tab:working-agents-44} summarizes the computation and the interpretive scope of the three classes of metric. Overall, a metric should align with the object and the quality requirements actually accepted, and should state its success threshold, partial credit, comparison baseline, and the source of its weights, so that scores from different benchmarks are not treated as fully equivalent scales of capability.

\FloatBarrier

\subsection{Open-Source Harnesses and Execution Environments}\label{frameworks}

Evaluating a work agent depends not only on the model and the task but also on the harness configuration. An agent harness handles context management, tool invocation, and execution control, while an evaluation harness handles task initialization, run recording, and result verification. OpenHands, SWE-agent, Codex, and Claude Code supply agent execution mechanisms~\cite{openhands,sweagent,codexdocs,claudedocs}; BrowserGym, the SWE-bench harness, Harbor, and Inspect AI support standardized task execution and evaluation~\cite{swebench,browsergym,harbor,inspectai}.

The infrastructure of a benchmark is closely tied to its work objects. Software-engineering tasks depend on code repositories and test environments, web and desktop tasks on resettable interactive environments, and enterprise tasks check business state through APIs and databases, while office and design tasks require verifying native files, computed results, and the actual deliverables. Table~\ref{tab:benchmark-runtimes} summarizes the execution environments and verification mechanisms of representative benchmarks.

\begingroup
\footnotesize
\setlength{\tabcolsep}{3pt}
\renewcommand{\arraystretch}{1.12}
\begingroup
\fontsize{7.4}{9.2}\selectfont
\setlength{\tabcolsep}{2.5pt}
\renewcommand{\arraystretch}{1.12}
\arrayrulecolor{tableRule}
\Needspace{150pt}
\begin{longtable}{@{}>{\raggedright\arraybackslash}m{72.000pt}
 >{\raggedright\arraybackslash}m{128.000pt}
 >{\raggedright\arraybackslash}m{108.000pt}
 >{\raggedright\arraybackslash}m{142.000pt}@{}}
\caption{Harnesses and execution environments of representative work-agent benchmarks.}\label{tab:benchmark-runtimes}\\
\toprule
\rowcolor{tableHeader}
\textcolor{tableInk}{\bfseries Task type} & \textcolor{tableInk}{\bfseries Representative benchmarks} & \textcolor{tableInk}{\bfseries Agent harness / execution} & \textcolor{tableInk}{\bfseries Evaluation environment and verification} \\
\midrule
\endfirsthead
\multicolumn{4}{@{}l@{}}{\small\tablename~\thetable~(continued)}\\
\toprule
\rowcolor{tableHeader}
\textcolor{tableInk}{\bfseries Task type} & \textcolor{tableInk}{\bfseries Representative benchmarks} & \textcolor{tableInk}{\bfseries Agent harness / execution} & \textcolor{tableInk}{\bfseries Evaluation environment and verification} \\
\midrule
\endhead
\midrule
\multicolumn{4}{r@{}}{\color{tableMuted}\scriptsize Continued on next page}\\
\endfoot
\bottomrule
\endlastfoot
\MethodName{Software engineering} & \MethodEntry{SWE-bench}{swebench}
\MethodEntry{SWE-Lancer}{swelancer} & \MethodEntry{SWE-agent}{sweagent}
\MethodEntry{OpenHands}{openhands}
\MethodEntry{Codex}{codexdocs} & Docker, code repositories, and test programs \\
\midrule
\MethodName{Terminal tasks} & \MethodEntryRes{Terminal-Bench 2.0}{terminalbench}{terminalbench2} & \MethodEntry{Claude Code}{claudedocs}
\MethodEntry{Codex}{codexdocs}
\MethodEntry{Pi}{piagent} & \MethodEntry{Harbor}{harbor} isolated containers and verification scripts \\
\midrule
\MethodName{Web interaction} & \MethodEntry{WebArena}{webarena}
\MethodEntry{WorkArena}{workarena}
\MethodEntry{VisualWebArena}{visualwebarena} & Browser agents & \MethodEntry{BrowserGym}{browsergym} AgentLab, and website state checks \\
\midrule
\MethodName{Desktop and mobile} & \MethodEntry{OSWorld}{osworld}
\MethodEntry{OSWorld 2.0}{yuan2026osworld2}
\MethodEntry{AndroidWorld}{androidworld}
\MethodEntry{Windows Agent Arena}{windowsarena} & GUI and computer-use agents & Virtual machines, emulators, and application-state verification \\
\midrule
\MethodName{API and business operations} & \MethodEntry{AppWorld}{appworld}
\MethodEntry{$\tau$-bench}{taubench}
\MethodEntry{$\tau^2$-bench}{tau2bench} & Tool-calling agents & Application APIs, user simulators, and database verification \\
\midrule
\MethodName{Enterprise operations} & \MethodEntry{CRMArena}{crmarena}
\MethodEntry{AutomationBench}{automationbench}
\MethodEntry{EnterpriseOps-Gym}{enterpriseops} & API and MCP agents & Enterprise application sandboxes, business state, and constraint checks \\
\midrule
\MethodName{Tools and context} & \MethodEntry{MCP-Atlas}{mcpatlas}
\MethodEntry{ToolSandbox}{toolsandbox}
\MethodEntry{LOCA-bench}{zeng2026loca} & MCP and tool agents & Stateful tools, context configuration, and task verification \\
\midrule
\MethodName{Office artifacts} & \MethodEntry{SpreadsheetBench}{spreadsheetbench}
\MethodEntry{PPTC}{pptc}
\MethodEntry{PresentBench}{presentbench} & File, script, or GUI agents & Workbook computation, file parsing, and rendering \\
\midrule
\MethodName{Integrated office work} & \MethodEntry{OfficeBench}{officebench}
\MethodEntry{OmegaUse-OfficeVal}{omegauseofficeval} & GUI, API, or hybrid execution & Office workspaces and native-artifact inspection \\
\midrule
\MethodName{3D design} & \MethodEntry{CADBench (BlenderLLM)}{du2024blenderllmtraininglargelanguage}
\MethodEntry{CADBench (MIT)}{doris2026cadbench}
\MethodEntry{BlenderBench (VIGA)}{yin2026visionasinversegraphics}
\MethodEntry{BenchCAD}{benchcad} & Blender and CAD agents & Program execution, geometric comparison, and rendering assessment \\
\midrule
\MethodName{Enterprise collaboration} & \MethodEntry{TheAgentCompany}{agentcompany}
\MethodEntry{Workspace-Bench}{workspacebench} & Multi-tool workspace agents & Enterprise services, file dependencies, and task checkpoints \\
\midrule
\MethodName{Professional delivery} & \MethodEntry{StartupBench}{startupbench}
\MethodEntry{BankerToolBench}{bankertoolbench}
\MethodEntry{APEX-Agents}{apex} & Professional workspace agents & File inspection, professional rubrics, and agentic evaluation \\
\midrule
\MethodName{Occupational work} & \MethodEntry{GDPval}{gdpval}
\MethodEntry{Agents' Last Exam}{sun2026agentslastexam}
\MethodEntry{Remote Labor Index}{rli} & Configurable professional task agents & Professional work-product submission and task acceptance \\
\midrule
\MethodName{General evaluation} & \MethodEntry{AgentBench}{agentbench}
\MethodEntry{HAL}{hal}
\MethodEntry{Harness-Bench}{harnessbench} & Various model--harness configurations & Unified task environments, execution traces, and result scoring \\
\end{longtable}
\endgroup
\par\endgroup

Recent work has begun to examine the effect of the harness on evaluation results directly. HAL supports cross-benchmark comparison of different models and agent scaffolds~\cite{hal}; Harness-Bench further analyzes how model--harness combinations affect task performance and execution behavior~\cite{harnessbench}. These studies indicate that benchmark scores should be interpreted together with the execution framework, the tool configuration, and the resource budget, rather than attributed entirely to the underlying model.

\FloatBarrier

\subsection{Performance, Cost, and Economic Value}\label{economics}

Task success rates and delivery quality measure what an agent can do, but they do not capture its execution cost or its actual economic value. Recent work extends the evaluation of work agents along three levels: the performance--cost trade-off, the value of the professional work itself, and the utility of deployment.

\subsubsection{Trading Performance against Execution Cost}\label{performance-cost}

Task performance and execution cost do not always move together. HAL compares task scores and invocation cost jointly, so that a Pareto frontier can identify competitive system configurations under different cost constraints~\cite{hal}. The analysis applies both to model comparison and to complete agent systems.

HarnessTax further finds that when the same model is run under different harnesses, invocation cost can differ substantially even when task success rates are close~\cite{harnesstax}. Harness-Bench analyzes how the execution configuration affects task performance~\cite{harnessbench}, and automatic adaptation of instructions, tools, and the execution flow improves the cost efficiency of some smaller models~\cite{betterharnesses}. These results show that the performance--cost trade-off depends not only on the choice of model but also on the design of the harness.

\subsubsection{The Value of Professional Work}\label{work-value}

Execution cost reflects the resources needed to complete a task, whereas economic value also depends on the task itself. SWE-Lancer connects software-engineering tasks to real market pay~\cite{swelancer}; GDPval has occupational experts compare professional deliverables~\cite{gdpval}; and Remote Labor Index evaluates end-to-end automation using actual paid remote projects~\cite{rli}.

OmegaUse-OfficeVal further combines office-artifact quality, human completion time, and a task price proxy~\cite{omegauseofficeval}, while a task-completion time horizon uses human task duration to characterize the scale of work an agent can complete~\cite{metrhorizon}. The task distribution of existing agent benchmarks also diverges from the real labor market~\cite{agentrealwork}. Evaluating work value therefore has to account for task representativeness and professional acceptance, and annotation pay or human working hours should not be equated directly with actual economic return.

\subsubsection{Reliable Deployment and Agentic ROI}\label{agentic-roi}

High task performance and low invocation cost do not necessarily produce a positive return on deployment. The position that the real barrier to agent usability is agentic ROI argues that an agent's practical utility depends on delivery quality, the human time saved, and the cost of use, including the effort users spend describing, checking, and revising the work~\cite{agenticroi}.

READY further brings human oversight into reliability and cost evaluation, finding that agents with comparable autonomous accuracy may require different amounts of human assessment to meet the same deployment requirement~\cite{ready}. Agentic ROI therefore supplies a conceptual view of user utility, and READY an empirical complement in the form of joint human--machine cost under a reliability constraint. Actual deployment return must also account for failure repair, system integration, and long-term maintenance.

Overall, the economic evaluation of work agents needs to distinguish the resource cost of executing a task, the potential economic value of the work produced, and the actual deployment return once human oversight is taken into account. The performance--cost Pareto frontier, the harness tax, and agentic ROI complement conventional task-performance evaluation from the perspectives of system efficiency, execution configuration, and user utility respectively. Section~\ref{frontiers} develops how these systems can extend from completing assigned tasks to carrying responsibility as workspaces and requirements change.

\FloatBarrier

\section{Frontiers: Toward Next-Generation Work Agents}\label{frontiers}

A manager asks an agent to prepare a report for a recurring business meeting. Completing the first report requires finding the right records, resolving discrepancies, and explaining the results. Taking responsibility for later reports means following revisions, obtaining missing information, and acting before each deadline. The manager may also ask for a recommendation that reflects how a trusted analyst would weigh new evidence. Other teams must confirm inputs, resolve shared assumptions, and approve changes before the report can be used.

These requests extend delegation in three directions. Sustaining the manager's goals motivates \textit{personal work agents}; representing the analyst's judgment motivates \textit{human digital twins}; coordinating contributions toward the report motivates \textit{organizational work systems}. In each case, the agent must keep work moving as information and responsibilities change. Experience from one reporting cycle can improve the next, while changes to the method may affect results that others have already used.

\begin{figure}[!htb]
\centering
\includegraphics[width=\linewidth]{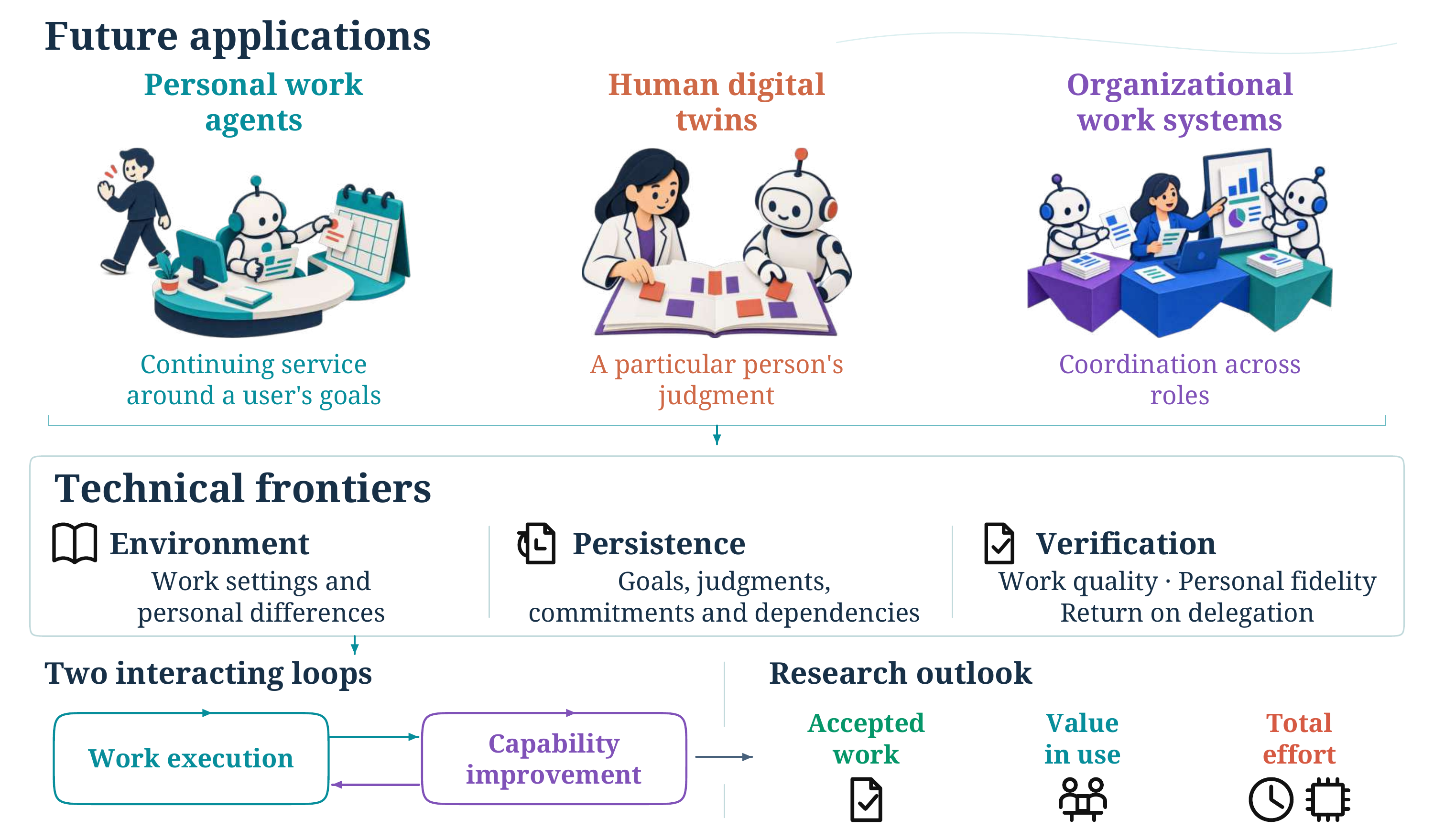}
\caption{Research directions for next-generation work agents. Continuing service around a user's goals, representation of a particular person's judgment, and coordination across organizational roles motivate advances in environment, persistence, and verification. Interacting execution and improvement loops support these responsibilities, while accepted work, value in use, and total effort guide the assessment of beneficial delegation.}
\label{fig:frontier-systems}\label{fig:continuing-work-system}
\end{figure}

We develop this agenda through four questions. Section~\ref{frontier-work-systems} examines what people can delegate in these applications. Section~\ref{frontier-technical} identifies the advances in environment, persistence, and verification they require. Section~\ref{frontier-loops} asks how execution experience can improve the agent while existing work continues. Section~\ref{frontier-value} asks under which work conditions this delegation is beneficial. Figure~\ref{fig:frontier-systems} maps the connections, and Table~\ref{tab:frontier-overview} summarizes the research questions.

\begin{table}[!htb]
\centering
\fontsize{7.4}{9.2}\selectfont
\setlength{\tabcolsep}{2.5pt}
\renewcommand{\arraystretch}{1.12}
\arrayrulecolor{tableRule}
\caption{Research agenda: what people can delegate, which technical advances support it, how execution and learning interact, and where delegation is beneficial.}
\label{tab:frontier-overview}
\begin{tabularx}{\linewidth}{@{}>{\raggedright\arraybackslash}m{0.225\linewidth} >{\raggedright\arraybackslash}X >{\raggedright\arraybackslash}X@{}}
\toprule
\rowcolor{tableHeader}
\textcolor{tableInk}{\bfseries Research direction} & \textcolor{tableInk}{\bfseries Open question} & \textcolor{tableInk}{\bfseries Evidence of progress} \\
\midrule
\rowcolor{tableHeader}
\multicolumn{3}{@{}l@{}}{\textcolor{tableInk}{\textbf{\ref{frontier-work-systems} Future applications}: what can people delegate?}} \\
\MethodName{Personal work agents (\S\ref{personal-working-agents})} & How can an agent advance unfinished goals between requests? & Timely progress on agreed goals with less repeated specification and intervention. \\
\midrule
\MethodName{Human digital twins (\S\ref{digital-workforce-twins})} & How can an agent reflect a particular person's judgment in new situations? & Agreement with that person's decisions under comparable information, assessed separately from task quality. \\
\midrule
\MethodName{Organizational work (\S\ref{human-agent-organization})} & How can roles, decision rights, and dependencies be coordinated toward a joint delivery? & Completed handoffs and end-to-end deliveries with less rework caused by coordination failures. \\
\midrule
\rowcolor{tableHeader}
\multicolumn{3}{@{}l@{}}{\textcolor{tableInk}{\textbf{\ref{frontier-technical} Technical frontiers}: what makes that delegation possible?}} \\
\MethodName{Environment (\S\ref{frontier-environment})} & Which work settings and personal differences must practice represent? & Transfer across work settings and correspondence to individual judgments in held-out situations. \\
\midrule
\MethodName{Persistence (\S\ref{reliability})} & How should goals, personal judgments, commitments, and feedback guide continuing work? & Timely action, retained obligations, and updates consistent with the current basis of delegation. \\
\midrule
\MethodName{Verification (\S\ref{frontier-verification})} & What evidence establishes work quality, personal fidelity, and return on delegation separately? & Separate measures of deliverable quality, person-specific decision agreement, and benefit relative to cost. \\
\midrule
\rowcolor{tableHeader}
\multicolumn{3}{@{}l@{}}{\textcolor{tableInk}{\textbf{\ref{frontier-loops} Interacting loops}: how can work continue while capability improves?}} \\
\MethodName{Work execution (\S\ref{work-execution-loop})} & How can responsibility survive interruptions and changes of executor? & Resumed work that preserves confirmed effects, pending dependencies, and agreements. \\
\midrule
\MethodName{Capability improvement (\S\ref{capability-improvement-loop})} & Which experience should change memory, skills, harness, or policy? & Gains on independent tasks and work settings from targeted updates. \\
\midrule
\MethodName{Coupling the loops (\S\ref{continual-adaptation})} & Which update scope and timing yield gains after rework and coordination costs? & More accepted and used work per total effort, including repair of affected ongoing work. \\
\midrule
\rowcolor{tableHeader}
\multicolumn{3}{@{}l@{}}{\textcolor{tableInk}{\textbf{\ref{frontier-value} Research outlook}: where does the system expand useful work?}} \\
\MethodName{Value under openness and persistence (\S\ref{frontier-value})} & Under which discovery and continuity demands does delegation pay off? & A wider range of work completed and used with sustainable demands on people and resources. \\
\bottomrule
\end{tabularx}
\end{table}

\subsection{Future Applications: From Assistance to Delegated Work}\label{frontier-work-systems}

The three applications concern different parts of the same reporting process: the manager's continuing goals, the analyst's professional judgment, and the shared delivery. Each places different demands on what the agent must remember, decide, and verify.

\subsubsection{Personal Work Agents}\label{personal-working-agents}

A personal work agent helps a user keep an unfinished goal moving across requests. For the manager, this means retaining the purpose of the meeting, gathering materials as they become available, and following up on missing inputs before the deadline. When new information arrives, the agent must decide whether to advance an agreed task, ask the user for a decision, or wait~\cite{humansautomation}. Proactive-agent research studies anticipation and intervention timing \cite{proactiveagent,proactivitygym}; VibeLifeBench and PAST-Bench examine continuing assistance and gains from retained experience \cite{vibelife,pastbench}.

Such service begins with helping the user specify the work. A manager may know which decision the report should support without knowing which records or analyses it requires. The agent can clarify the decision, identify missing inputs, and turn the agreed goal into tasks for specialist executors. Magentic-One's progress tracking and AutoGen's configurable agent conversations provide mechanisms for coordinating those tasks \cite{magenticone,autogen}. Retaining the open goals, completed work, and events that warrant further action lets the agent resume across reporting cycles with less repeated explanation from the user.

\begin{samepage}
Claude Duet, introduced as an experimental release in October 2026, illustrates this direction in a personal service: it checks connected applications, prepares briefings, and continues delegated tasks while the user is away, notifying the user when work is complete or needs a decision~\cite{claudeduet}. This makes the interval between requests part of the work: the agent must choose what to advance and when the user's attention is needed.

\end{samepage}

\subsubsection{Human Digital Twins for Professional Work}\label{digital-workforce-twins}

The manager may need a recommendation that reflects how a particular analyst weighs new evidence. A human digital twin aims to represent that person's judgment across new situations. Two analysts can draw different conclusions from the same records because they use different assumptions or accept different tradeoffs. Professional competence concerns the quality of the work; personal fidelity concerns whose judgment the system represents.

Existing studies address parts of this correspondence. LLM-based twin architectures support context-aware knowledge exchange, agents grounded in interviews and questionnaires model individual behavior, and TwinVoice evaluates fidelity to a person's communication and persona \cite{humandigitaltwin,selfreportagents,twinvoice}. GRACE extends representation into clinical communication through clinician-informed dialogue scripts and retrieved medical knowledge \cite{providertwin}. Extending this correspondence to professional action requires learning how a particular person weighs evidence, handles uncertainty, and chooses among acceptable courses of action.

Such a twin should be evaluated on held-out cases in which the represented person makes or assesses a decision from comparable information. A personalized professional agent that knows the person's preferences provides a comparison for testing whether the twin also captures characteristic judgments.

The twin's recommendations also enter work done by other participants. Authenticated delegation provides mechanisms for scoped authority~\cite{authdelegation}, while handoff studies show how constraints can be lost when context is transferred~\cite{handoffcollapse}. Passing the analysis to the manager's agent therefore requires retaining both its assumptions and the authority under which it was produced.

\subsubsection{Organizational Work Systems}\label{human-agent-organization}

Organizational work joins contributions into a delivery that other people can use. The analyst's forecast must reflect assumptions agreed with other teams, reach the manager's report, and arrive before the meeting. MetaGPT provides an example of coordination through structured roles, procedures, and intermediate artifacts in software development~\cite{metagpt}. TheAgentCompany places agents in shared applications with simulated colleagues~\cite{agentcompany}. Failure analyses and controlled scaling studies show how communication structure, verification, and task dependencies affect multi-agent execution \cite{mast,scalingagents}.

The central challenge is coordinating role responsibilities, decision rights, and interdependent contributions toward a joint delivery. A revised sales estimate may require an analyst to update the forecast, another team to confirm an assumption, and a manager to authorize release. The system must determine who is responsible for each unresolved dependency, which decisions can proceed in parallel, and how revisions affect work already passed to another participant. Human--agent teamwork research and WORKBank's occupational preferences inform how people choose to share those responsibilities~\cite{teamplayer,workbank}. Progress requires dependable handoffs and conflict resolution that advance the overall delivery while limiting avoidable rework.

\FloatBarrier
\subsection{Technical Frontiers for Sustained Delegation}\label{frontier-technical}

The three applications place different demands on sustained delegation: personal agents must advance goals between requests, digital twins must remain faithful to a particular person's judgment, and organizational systems must coordinate work across roles. \textbf{Environment} research asks how to represent situations that exercise these abilities. \textbf{Persistence} asks how goals, personal judgments, commitments, and dependencies remain available and applicable over time. \textbf{Verification} asks which evidence establishes progress on the delegated responsibility, correspondence to the represented person, and the quality and value of the resulting work.

\subsubsection{Environment: Learning across Realistic Work Worlds}\label{frontier-environment}\label{env-synthesis}\label{env-fidelity}\label{generalization}\label{execution-infrastructure}

Environment research must decide what work to represent, which features make the experience realistic, how to run it, and where additional practice is most useful.

\textbf{Occupational coverage and transfer.} A teacher, analyst, and procurement specialist may all produce presentations while relying on different knowledge and standards. O*NET organizes occupational requirements; GDPval and the Remote Labor Index connect professional tasks to deliverables; WORKBank adds preferences about delegation \cite{onetcontentmodel,gdpval,rli,workbank}. Together with analyses of occupational exposure~\cite{gptsaregpts}, these resources can guide coverage across domains, occupations, and workflows. Testing held-out combinations would reveal what transfers: reconciling inconsistent records may reuse a common procedure, while deciding why a discrepancy matters may require knowledge of the new profession. Those gaps then guide the construction of further experience.

For digital twins, environments should also vary the represented person within the same occupation and task. The same reporting problem may lead two analysts to different choices because they use different assumptions or commitments. Holding the available evidence and authorization constant where feasible would isolate differences in judgment; varying authorization separately would test whether actions remain within the delegated role.

\textbf{Decision-relevant realism.} Within each kind of work, an environment should preserve the relationships that affect decisions. Workspace-Bench makes file dependencies explicit; business and workspace synthesis methods construct connected entities and evidence relations \cite{workspacebench,agentmercury,graphforge}. Generating workspaces through successive episodes of activity could add the histories behind these relationships. Earlier reports would establish conventions, revisions would create competing versions, and parallel workflows would leave plausible but irrelevant materials. Shared-state settings such as $\tau^2$-bench and Collaborative Gym add participants who change the environment \cite{tau2bench,cogym}, while simulation-to-reality studies examine transfer to real interactions \cite{sim2realgap}. Varying source ambiguity, dependency depth, and incomplete updates would then exercise the decisions needed to clarify and repair work.

\textbf{Execution infrastructure.} These environments need to support repeated practice and continuation from interrupted work~\cite{computerrl}. BrowserGym provides common web observations and actions, while Agent Lightning connects harness execution to policy training \cite{browsergym,agentlightningv1,agentlightning}. DSec separates stateful rollouts from preemptible GPU training \cite{deepseekdsec}, and AIOpsLab composes services, workloads, faults, and telemetry for operational tasks \cite{aiopslab}. Building on these mechanisms, a work environment could package the application versions, source materials, business state, and events needed to reproduce an episode, together with checks for its outcome. Saved states and records of completed effects would let training and evaluation resume the same unfinished work.

\textbf{Practice allocation.} Once environments are available, the remaining question is which experiences deserve the execution budget. Repository environments provide executable practice \cite{swegym,r2egym}; Skill2Env targets capability gaps, and AgentGym-RL progressively expands interaction horizons \cite{skill2env,agentgymrl}. For professional work, an expensive episode involving clarification and coordination may teach more than many easily scored actions. Comparing learning gains and transfer across these choices would turn the construction methods in Section~\ref{experience} into a curriculum guided by the work agents still struggle to perform.

\subsubsection{Persistence: Memory, Initiative, and Adaptation}\label{reliability}\label{delayed-effects}\label{persistent-work-state}\label{continuous-work}

Realistic environments expose changes that unfold between interactions. Handling them requires maintaining an open responsibility: work that remains due even when no agent is currently running. The reporting deadline, pending source confirmation, and authority to release the result must survive the pause. This creates three connected needs: remembering what still applies, deciding when to act, and learning from repeated work.

\textbf{Memory grounded in current work.} The agent needs to retrieve relevant decisions and procedures and determine whether they still apply. Personalized workflow prediction studies which procedures fit a user's context~\cite{wfwa0519}. Memory management and temporal representations support retention and revision \cite{memgpt,zep,chronomem,mem0,generativeagents}; A-MEM links and revises notes, while Memory-R1 learns memory operations and selection from downstream rewards \cite{amem,memoryr1}. StreamMemBench exposes gaps between storing evidence, using it, and reusing feedback \cite{streammembench}. In the report, a convention may expire when the audience changes while the reconciliation procedure remains useful.

Memory should also distinguish a user's goals and preferences from a represented person's judgments, existing commitments, and authorization. Linking each item to the relevant person, role, source, and period of applicability would help the agent distinguish a preference about presentation from a commitment governing the analysis or the authority needed to release it.

\textbf{Initiative guided by responsibilities.} Retained obligations give the agent a reason to act between requests. Proactive Agent learns from activity traces and judgments of proposed assistance; Proactivity-Gym studies intervention timing and trust \cite{proactiveagent,proactivitygym}. VibeLifeBench examines needs that become visible through inspecting external state \cite{vibelife}, and temporal and asynchronous environments provide settings for choosing when to observe, ask, act, or wait \cite{timearena,asynclm,asynchow}. A revised sales record may require several reports to be updated, while an approaching deadline may warrant asking for a missing input. The agent also needs to know which decisions it has been delegated. Collecting a missing input, recommending a conclusion, approving its release, and accepting a new commitment involve different decision rights. It can advance tasks covered by the current delegation and seek a decision when a new choice or commitment lies outside that scope.

\textbf{Learning from repeated work.} Repeated reporting cycles supply experience for improving source selection, analysis, and explanations. Reflection, experience abstraction, skill libraries, and workflow memory retain reusable lessons \cite{reflexion,expel,voyager,awm}; SkillRL connects skill evolution to policy learning, and PAST-Bench measures gains from retained experience \cite{skillrl,pastbench}.

Feedback can revise the current report, the procedure used for later reports, or the representation of the analyst. A corrected source value repairs the current result; repeated reconciliation errors motivate a reusable check; feedback from the analyst reveals how that person weighs evidence. Retaining which part of the work each correction concerns gives the agent a basis for later adaptation. Work quality and personal fidelity require separate assessment. Section~\ref{continual-adaptation} examines how procedural updates affect work already in progress.

\subsubsection{Verification: Work Quality, Personal Fidelity, and Return on Delegation}\label{frontier-verification}\label{oracle-free-verification}\label{adaptive-verification}\label{continuous-verification}\label{downstream-value}

Continuing service raises three evaluation questions: does the delivery meet its requirements, does a digital twin reflect the represented person's judgment, and does delegation yield a useful return? These questions require separate evidence about work quality, personal fidelity, and benefit relative to human and computational effort. Verification must follow each as the work and its use evolve.

\textbf{Task and deliverable acceptance.} A report may contain correct numbers yet fail to inform the decision it is meant to support. Professional acceptance therefore combines direct checks with judgments about purpose and audience. PresentBench uses slide-specific checklists, PRBench elicits expert criteria, and GDPval uses expert comparisons of occupational deliverables \cite{presentbench,prbench,gdpval}. Agent-as-a-Judge provides mechanisms for inspecting files and execution evidence against requirements \cite{agentjudge}. The research task is to connect these checks to the brief: determine which criteria transfer across occupations, which require expert input, and which requirements need clarification. Variation across web benchmarks and evidence of judge bias make consistency across these settings an important evaluation target \cite{agentrewardbench,mtbench,selfpreference,greatmodels}.

\textbf{Personal decision fidelity.} A twin's fidelity is assessed against decisions elicited from the represented person on held-out cases under comparable information. For the same report, two analysts may recommend different forecasts because they weigh uncertainty differently. Both recommendations may meet professional standards, yet only one may reflect the represented person's judgment. Conversely, a faithful prediction can reproduce that person's mistake. Evaluation should therefore report decision agreement separately from deliverable quality, record uncertainty or changes in the person's position, and check execution against the agreed authority. This distinguishes whose judgment the twin represents from the quality of the resulting work.

\textbf{Return on delegation.} Measuring the effort of delegation requires tracing the whole delivery process. Agentic ROI considers quality relative to a human baseline, time saved after human interaction, and monetary cost \cite{agenticroi}. The relevant effort includes specification, supervision, correction, and integration into later work. Computational cost also depends on the harness: HarnessTax reports similar success rates at substantially different API costs for the same model across harnesses on SWE-bench Lite and Terminal-Bench 2.0 \cite{harnesstax}. Recording these costs by stage would reveal where changes to the model or harness could reduce the effort of obtaining an accepted result.

The benefit side requires following what happens to that result. For the report, useful evidence includes whether it supported discussion at the meeting and reduced the work needed to reach a decision. Such benefits depend on the people and process using the artifact. A natural experiment on Change.org found changes in petition language without improved platform-level outcomes after AI writing assistance became available \cite{petitionoutcomes}. Comparisons with alternative ways of completing the same work are therefore needed to estimate the benefit attributable to delegation. These observations complement acceptance checks and supply the evidence used in Section~\ref{frontier-value}.

\textbf{Evidence over time.} New source data may invalidate an accepted analysis, while later use may reveal a missing requirement. StateGuard represents constraints, versioned variables, and conclusions in a validity graph \cite{stateguard}. Linking acceptance checks to the source, artifact, and requirement versions they concern would help identify which results need rechecking. Linking later feedback to the delivery and its use would show which procedures need improvement. These two paths lead directly to the interaction between ongoing execution and learning.

\subsection{Two Interacting Loops: Delivering Work and Improving Capability}\label{frontier-loops}

\begin{figure}[!htb]
\centering
\includegraphics[width=\linewidth]{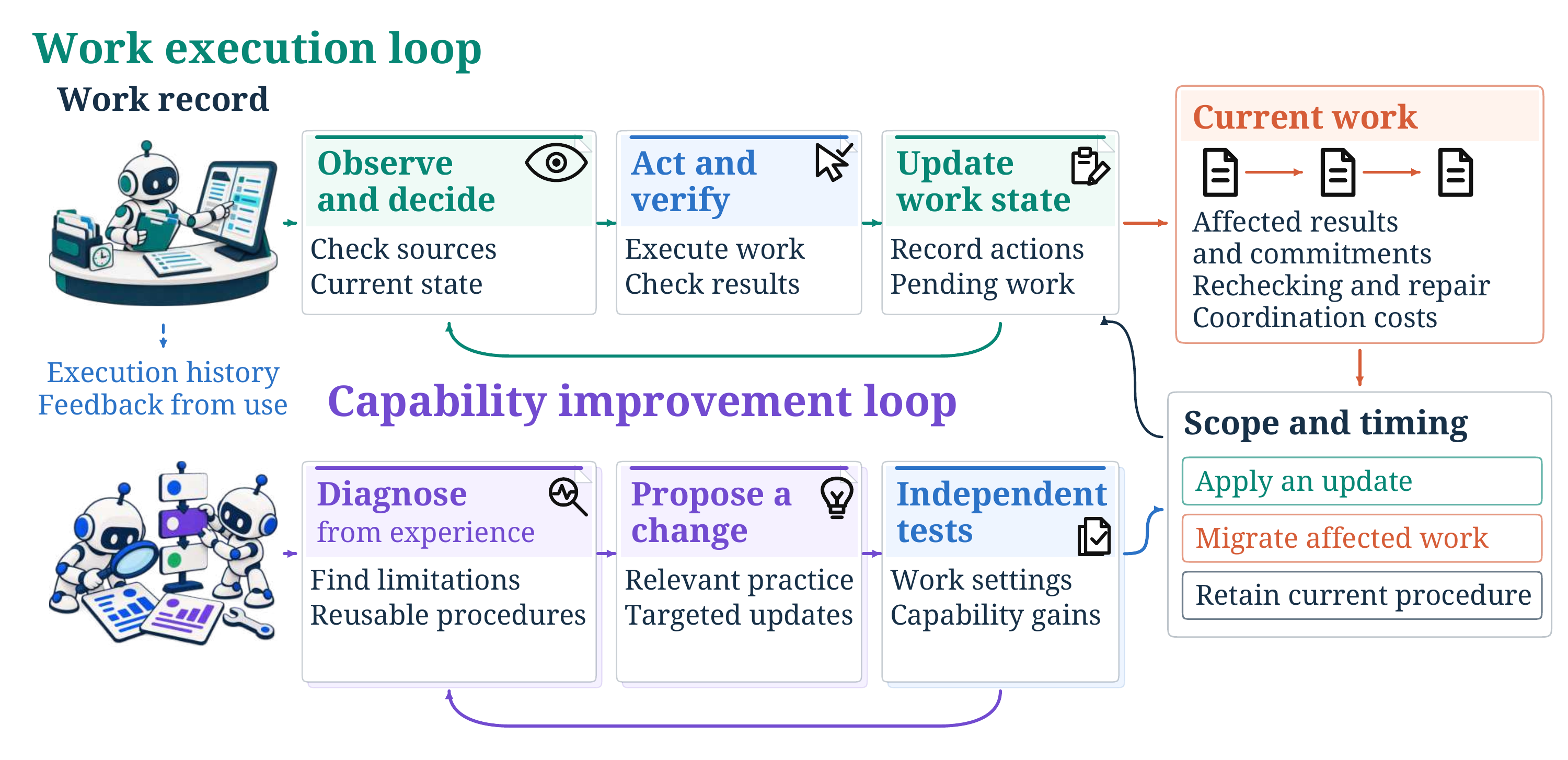}
\caption{Work execution and capability improvement during continuing service. A persistent work record retains agreed requirements, source versions, unfinished work, and completed actions. Execution history and feedback from downstream use inform targeted updates. Independent tests assess capability gains, while checks on current work identify affected results and commitments. This evidence guides whether to apply an update, migrate affected work, or retain the current procedure. The benefit of an update depends on its gains relative to the resulting rework and coordination costs.}
\label{fig:frontier-two-loops}
\end{figure}

Execution and learning share the same workspace. A new analysis procedure may improve future reports while changing an estimate that another team has already used. Introducing it requires tracing how its effects propagate through intermediate results, downstream decisions, and ongoing commitments. The \textit{work execution loop} advances current responsibilities and records what happened. The \textit{capability improvement loop} learns from that experience and changes how later work is done. Their connection supplies the evidence for choosing the scope and timing of updates while work remains active. Figure~\ref{fig:frontier-two-loops} shows how updates return to ongoing execution.

\subsubsection{The Work Execution Loop}\label{work-execution-loop}\label{work-object}

The execution loop needs a record that lasts across agent runs. A \textit{persistent work record} can hold the report's agreed requirements, source versions, pending approvals, and completed actions; case handling provides a precedent for such enduring records \cite{casehandling}. When a revised sales estimate arrives, the agent identifies the affected analysis, updates it, checks the result, and records what remains due. Approval closes that delivery while leaving the next reporting cycle active.

Interruptions make the relationship between this record and the workspace especially important. An external action may complete before the agent retains its confirmation \cite{nonatomic,helland}. On resuming, the agent needs to inspect the actual effect before deciding to retry or repair, using the execution-state distinctions in Section~\ref{execution}. Provenance and compensating operations provide mechanisms for reconstructing and correcting these effects \cite{provenance,sagas,sagallm,atomix}. The resulting history supports both the next participant taking over the work and the improvement loop diagnosing how it was performed.

\subsubsection{The Capability Improvement Loop}\label{capability-improvement-loop}

The improvement loop uses execution history and feedback from downstream use to identify reusable procedures and changes that could improve later work. If reports repeatedly miss a deadline because the agent discovers inconsistent inputs too late, an earlier source check may be more useful than a stronger drafting model. Other failures may call for professional training, a repaired tool interface, or better timing. The methods in Section~\ref{optimization} provide the update mechanisms; reflection, workflow memory, skill evolution, and harness-aware training offer ways to retain and learn from the experience \cite{reflexion,awm,skillrl,agentlightningv1}.

The loop then constructs relevant practice, proposes a change, and tests it in independent work settings. For the earlier source check, evaluation would examine whether it catches discrepancies and reduces later rework across different reports. Independent output checks against shared task requirements make gains comparable across changes to both the agent and its grader~\cite{rewardoveropt,aeslides,whoverifies}. Environment construction supplies practice that exercises the missing ability; verification tests whether the proposed update improves the resulting work. The remaining question is how to apply a successful update to work already in progress.

\subsubsection{Coupling Learning to Open Responsibilities}\label{continual-adaptation}\label{verifier-robustness}

Continual learning studies how updates preserve performance on earlier tasks~\cite{gem}. During service, other outputs or participants may already depend on results produced by the method being updated. The research problem is to trace the update's effects through these dependencies. In the reporting example, records should link source versions and analytical assumptions to calculations, conclusions, approvals, and downstream decisions. A new forecasting method may change a budget estimate already used by another team; an additional source check may leave that estimate intact.

For a digital twin, these dependencies also include the represented person's assumptions and subsequent choices. Evaluation therefore follows both task quality and personal fidelity, using the separate assessments in Section~\ref{frontier-verification}.

\Needspace{3\baselineskip}
These links define the scope of an update: which results to recompute, which checks to repeat, and which handoffs to reopen. Studies should compare changes to a single check, a reusable procedure, and the full model--harness configuration, measuring how far their effects propagate. This connects capability gains to the repair and coordination needed to put them into use. Audits of self-evolving financial agents further motivate examining how learning changes behavior and its effects~\cite{finevolveaudit}.

This makes \textit{net benefit during service} a concrete research target. Studies should vary update scope and timing while tasks remain unfinished, measuring gains in accepted and used work alongside training, inference, rechecking, repair, and human coordination costs over the same period. Comparisons with retaining the current procedure or updating only between deliveries would show when those gains repay the cost of change. The goal is to identify the work conditions and update granularity under which continual improvement pays off, including settings where a stable task template remains the more productive choice.

\FloatBarrier
\subsection{Research Outlook: Useful Work under Openness and Persistence}\label{frontier-value}\label{accepted-work}

\begin{figure}[!htb]
\centering
\includegraphics[width=\linewidth]{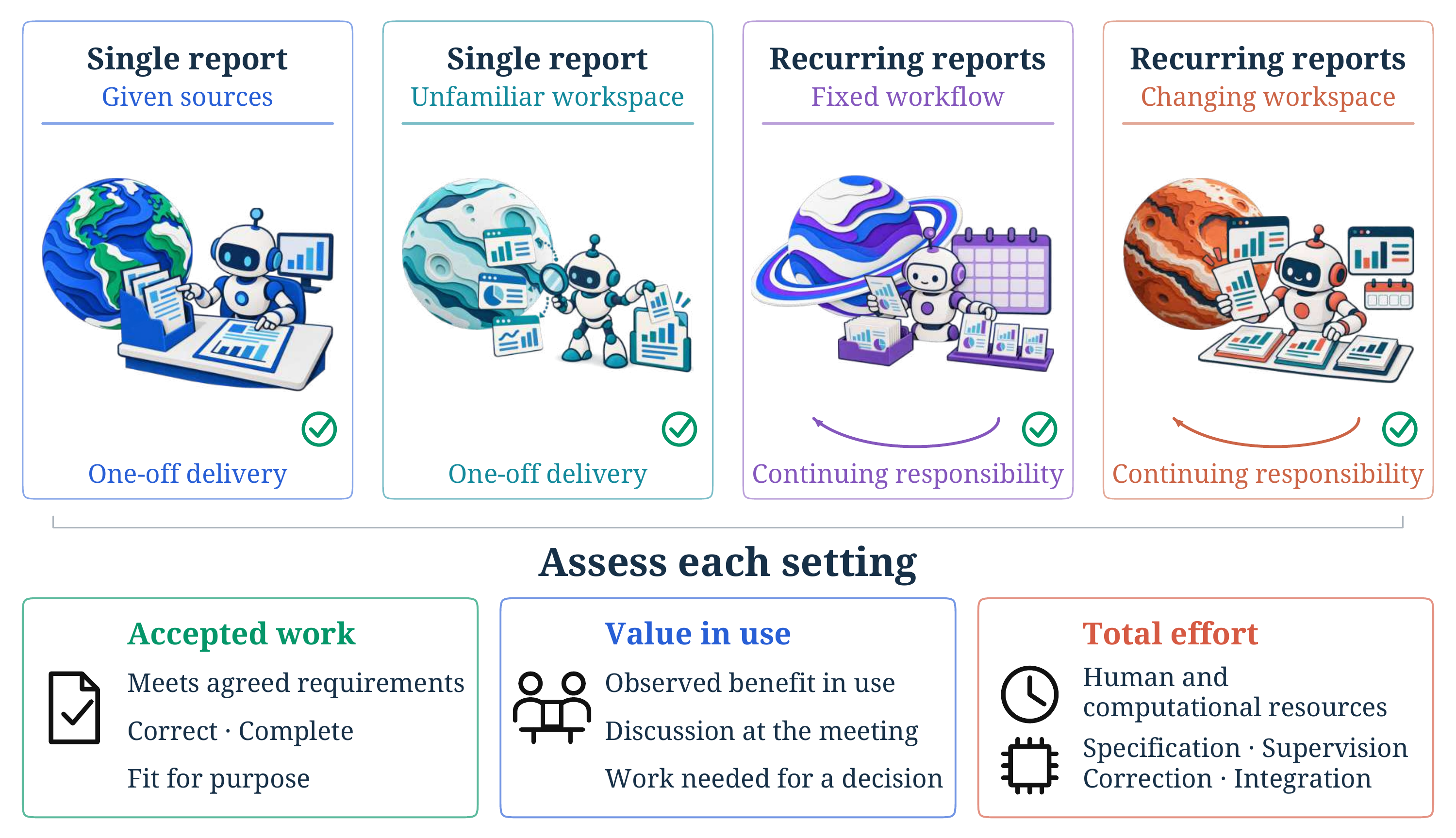}
\caption{Assessing beneficial delegation across work conditions. The four scenarios illustrate variation in work openness and persistence: single reports from given sources or in an unfamiliar workspace, and recurring reports within a fixed workflow or a changing workspace. Across all four settings, accepted work, value in use, and total human and computational effort provide complementary evidence for assessing delegation.}
\label{fig:frontier-beneficial-delegation}
\end{figure}

The systems above should expand the range of work that people can usefully delegate. Openness and persistence, introduced in Section~\ref{foundation}, provide two independent dimensions along which to test that expansion. Preparing a single report in an unfamiliar organization stresses discovery of sources, relationships, and authority. Preparing recurring reports in a fixed environment stresses continuity across revisions and interruptions. Combining the two asks whether an agent can learn its way around changing workspaces while keeping existing work moving. Figure~\ref{fig:frontier-beneficial-delegation} contrasts these settings and the evidence needed to assess them.

These conditions create different opportunities and costs. Persistent service can reuse context and procedures across deliveries, but it also requires continued observation, maintenance, and recovery. Open workspaces offer access to tasks whose inputs were never fully specified, while adding discovery and clarification effort. Studies can vary these demands separately and together, using the evidence in Section~\ref{frontier-verification} and the measures in Section~\ref{metrics} over the full period of delegation. The resulting comparisons would show where agents reduce the effort of existing work and where they make previously inaccessible work feasible.

The division of work between people and agents is part of this comparison. In a product-innovation experiment, individuals with AI assistance matched human teams without AI and produced proposals spanning technical and commercial perspectives \cite{cyberneticteammate}. In a study of experienced developers, the tested AI tools increased completion time \cite{metrdevrct}. Alongside experiments on professional writing \cite{noyzhang}, these findings motivate testing delegation across tasks and working arrangements. Changes to processes and skills can determine whether local gains improve overall productivity \cite{jcurve,itproductivityparadox}. For the reporting process, faster analysis matters when assessment and integration can turn it into a timely decision.

This returns the agenda to the two loops. As execution improves, assessment, integration, or human decisions may become the limiting stage~\cite{ironies}. The execution loop must organize work around that stage, and the improvement loop must address the capability or coordination that limits the next delivery. Progress is a wider range of professional work completed, accepted, and used with sustainable demands on people and resources.
\FloatBarrier

\section{Conclusion}\label{conclusion}

Reliable digital work depends on keeping task requirements, actions, and outcomes connected as work unfolds. The system view developed here makes these connections visible across tasks, domains, and interfaces. As work expands from returning an answer or making a local change to coordinating workflows and related workspace objects, agents must track more dependencies and use the results of one step to guide the next. Openness makes discovery part of this process; persistence carries responsibility across stages and events. The five core capabilities provide a way to understand where the process succeeds and where it breaks down.

Work execution supplies observations, procedures, corrections, and feedback that can improve subsequent work. Training can develop reusable knowledge and action patterns, while harness changes can improve access to evidence, execution control, and retained state. Model--harness optimization coordinates these updates as the model and the work change. Evaluation checks whether the resulting system produces answers, artifacts, state changes, and workflow results that meet their requirements. Distinguishing the inspected result, the assessed properties, the judging method, and the reported metric helps identify which part of the work an improvement actually advances.

Sustained delegation brings execution and learning together. An agent must respond to new information and improve its methods while completing work under existing agreements. Gains on new tasks must therefore be considered alongside the repair of dependent work and the effort of introducing updates. Personal work agents, human digital twins, and organizational work systems bring different tests of this capability: progress on continuing goals, fidelity to a particular person's judgment, and coordination toward a joint delivery. Advances in realistic environments, persistent coordination, and verification can widen the conditions under which these systems are useful. Their practical value depends on the work people accept and use, the benefits it brings, and the human and computational effort required to obtain it.

\clearpage
\bibliographystyle{unsrtnat}
\bibliography{main}
\end{document}